\RequirePackage[l2tabu, orthodox]{nag}
\documentclass[12pt,phd,a4paper,oneside]{ucl_thesis}

\usepackage{blindtext}

\usepackage{emptypage}

\usepackage{graphicx}

\usepackage{float}

\usepackage{amsmath}

\usepackage{gensymb}
\usepackage{textcomp}

\usepackage{setspace}
\usepackage{multirow}

\usepackage{bibentry} 

\usepackage[format=hang,font=small,labelfont=bf]{caption}

\usepackage{etoolbox}

\usepackage[printonlyused]{acronym}
\usepackage{comment}
\usepackage{bm}

\usepackage{afterpage}

\usepackage{booktabs}

\usepackage{xargs}                      
\usepackage[pdftex,dvipsnames]{xcolor}  

\usepackage[colorinlistoftodos,prependcaption,textsize=medium]{todonotes}
\newcommandx{\unsure}[2][1=]{\todo[linecolor=red,backgroundcolor=red!25,bordercolor=red,#1]{#2}}
\newcommandx{\change}[2][1=]{\todo[linecolor=blue,backgroundcolor=blue!25,bordercolor=blue,#1]{#2}}
\newcommandx{\info}[2][1=]{\todo[linecolor=OliveGreen,backgroundcolor=OliveGreen!25,bordercolor=OliveGreen,#1]{#2}}
\newcommandx{\improvement}[2][1=]{\todo[linecolor=Plum,backgroundcolor=Plum!25,bordercolor=Plum,#1]{#2}}
\newcommandx{\thiswillnotshow}[2][1=]{\todo[disable,#1]{#2}}

\usepackage{soul} 

\usepackage[list=true]{subcaption}

\usepackage{graphicx}
\usepackage{tabularx}

\usepackage{algorithm}
\usepackage{algpseudocode}

\newcolumntype{x}[1]{%
>{\centering\hspace{0pt}}p{#1}}%

\newcolumntype{y}[1]{%
>{\raggedleft\hspace{0pt}}p{#1}}%

\newcolumntype{z}[1]{%
>{\raggedright\hspace{0pt}}p{#1}}%

\usepackage{xspace} 

\newcommand*{\eqa}[2]{
\newdimen\textlength
\newdimen\spacelength
\settowidth{\textlength}{$#1$}
\spacelength = 0.90\columnwidth

\ifthenelse{\number\spacelength < \number\textlength}{
\begin{equation}\label{#2}
 \scalebox{\xintRound{2}{\spacelength/\textlength}}{$#1$}
\end{equation}
}
{
\begin{equation}\label{#2}
  #1
\end{equation}
}
}%

\usepackage{bibentry}
\makeatletter\let\saved@bibitem\@bibitem\makeatother
\usepackage[pdftex,hidelinks]{hyperref}
\makeatletter\let\@bibitem\saved@bibitem\makeatother
\makeatletter
\AtBeginDocument{
    \hypersetup{
        pdfsubject={Thesis Ph.D.},
        pdfkeywords={Thesis Keywords},
        pdfauthor={Jingcai Guo},
        pdftitle={Title},
    }
}
\makeatother

\usepackage[numbers,sort&compress]{natbib}
\usepackage{cleveref}
\usepackage{amsmath,epsfig}
\usepackage{threeparttable}
\usepackage{multirow}
\usepackage{booktabs}
\usepackage{amssymb}
\usepackage{graphicx}
\usepackage{epstopdf}
\usepackage{algorithm}
\usepackage{setspace}
\usepackage{caption}

\usepackage{rotating}

\begin{document}



\frontmatter
\makeatletter
\renewcommand {\@degree@string} {Doctor of Philosophy}
\makeatother
\addcontentsline{toc}{chapter}{Title Page}


\title{Learning Robust Visual-semantic Mapping for Zero-shot Learning}
\author{Jingcai Guo}
\department{Department of Computing}
\university{The Hong Kong Polytechnic University}	


\maketitle

\begin{abstract} 
\addcontentsline{toc}{chapter}{Abstract}
Zero-shot learning (ZSL) aims at recognizing unseen class examples (e.g., images) with knowledge transferred from seen classes. This is typically achieved by exploiting a semantic feature space shared by both seen and unseen classes, e.g., attributes or word vectors, as the bridge. In ZSL, the common practice is to train a mapping function between the visual and semantic feature spaces with labeled seen class examples. When inferring, given unseen class examples, the learned mapping function is reused to them and recognizes the class labels on some metrics among their semantic relations. However, the visual and semantic feature spaces are generally independent and exist in entirely different manifolds. Under such a paradigm, the ZSL models may easily suffer from the domain shift problem when constructing and reusing the mapping function, which becomes the major challenge in ZSL. In this thesis, we explore effective ways to mitigate the domain shift problem and learn a robust mapping function between the visual and semantic feature spaces. We focus on fully empowering the semantic feature space, which is one of the key building blocks of ZSL.

First, we consider to adaptively adjust to rectify the semantic feature space for ZSL. Conventional ZSL models generally regard the semantic feature space unchangeable during training. However, it can be observed that when mapping visual features to semantic feature space, the obtained semantic features are usually overly concentrated. This deficiency affects models' ability to adapt and generalize to more unseen classes. As we know, the process of human beings understanding things is constantly improving. Similarly, we argue the semantic feature space also needs to be dynamically adjusted to accommodate more robust learning of mapping function. Specifically, the adjustment is conducted on both the class prototypes and global distribution during training. Moreover, we also propose to combine the adjustment with a cycle mapping to formulate the training to a more efficient framework that can not only rectify the semantic feature space but also speed up the training process. 

Second, we consider to align the manifold structures between the visual and semantic feature spaces by expanding the semantic features. Compared to our first method which directly adjusts to rectify the semantic feature space, the expansion process is more conservative and soft. Specifically, we build upon an autoencoder-based model to generate several auxiliary semantic features combining with the previous ones, to expand the space. Additionally, the expansion is jointly guided by an embedded manifold extracted from the visual feature space, which retains its geometrical and structural information. By aligning the two feature spaces, the trained mapping function is more robust and well-matched that significantly mitigates the domain shift problem. 

Last, we consider to take a further step to explore and empower the correlation between the visual and semantic feature spaces in a more fine-grained perspective. Unlike most existing and our previous works, we decompose an image example into several parts and use an example-level graph-based model to measure and utilize certain relations among these parts. Taking advantage of recently developed graph neural networks, we further formulate the ZSL to a graph-to-semantic mapping problem, which can better exploit the visual and semantic correlation and the local substructure information in example. 

In summary, this thesis targets fully empowering the semantic feature space and design effective solutions to mitigate the domain shift problem and hence obtain a more robust visual-semantic mapping function for ZSL. Extensive experiments on various datasets demonstrate the effectiveness of our proposed methods.
\end{abstract}

\setcounter{tocdepth}{2} 

\renewcommand{\contentsname}{Table of Contents}
\addcontentsline{toc}{chapter}{Table of Contents}

\tableofcontents
\listoffigures
\addcontentsline{toc}{chapter}{List of Figures}
\listoftables
\addcontentsline{toc}{chapter}{List of Tables}
\newpage


\mainmatter
 
\setstretch{1.75}
\chapter{Introduction}
\label{ch:intro}
This thesis explores effective ways towards fully empowering the semantic feature space to mitigate the domain shift problem, and hence obtain a more robust visual-semantic mapping function for zero-shot learning. This is an important research problem with a wide range of applications in machine learning and multimedia areas. In this chapter, we first give a brief overview of the research problems in Section \ref{sec:intro:overview}. Then we highlight the main contributions of this thesis in Section \ref{sec:intro:contributions}. Finally, Section \ref{sec:intro:organization} outlines the thesis organization.

\section{Overview}
\label{sec:intro:overview}
\begin{figure}[t]
    \centerline{\includegraphics[width=0.99\textwidth]{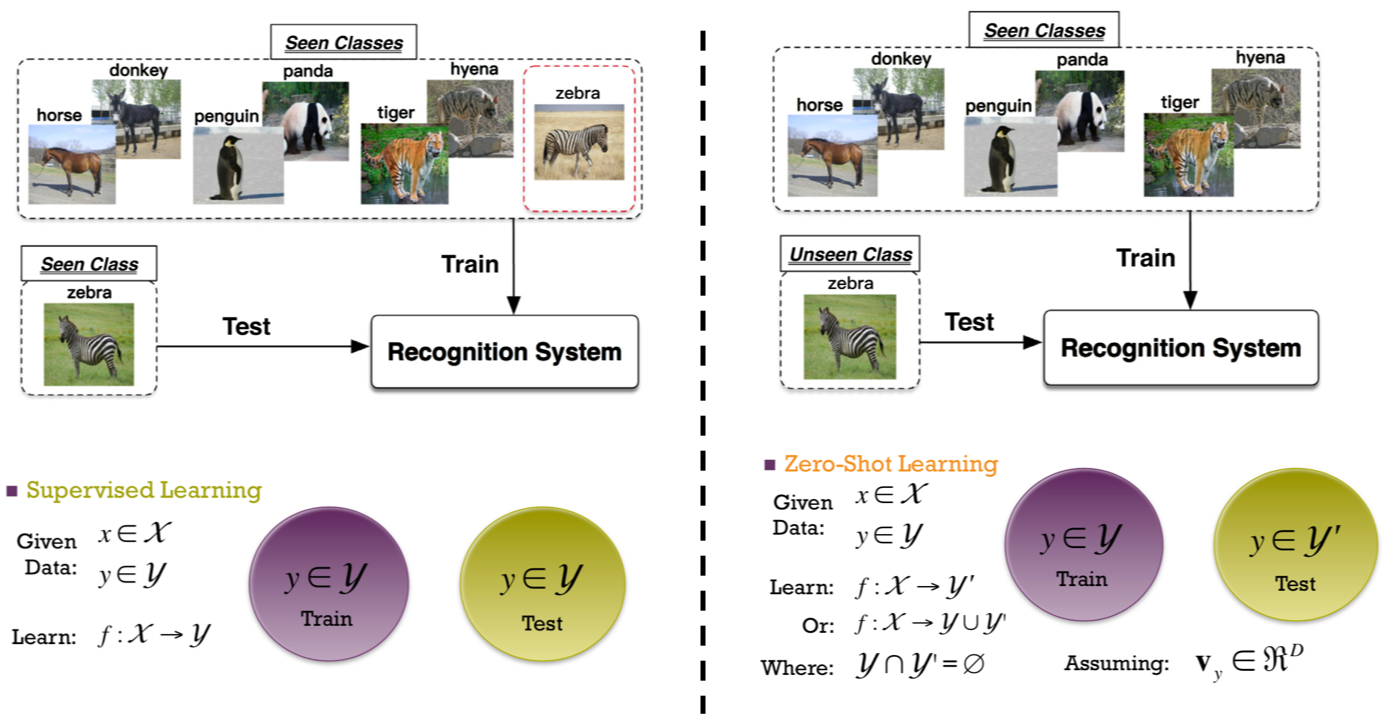}}
    \caption{Illustration of zero-shot learning setting: compared with the conventional supervised learning, the training and testing class-sets of zero-shot learning are disjoint from each other.}
    \label{zsl_setting}
\end{figure}

In the past few years, with the increasing development of deep learning techniques, many machine learning tasks and techniques have been proposed and consistently achieved state-of-the-art performance. Among them, most of the tasks can be grouped into supervised learning problems, such as image processing \cite{krizhevsky2012imagenet,ren2016faster,zhao2017diversified,zhang2018multilabel,guo2020dual}, face verification \cite{hu2014discriminative,ding2015robust,taigman2014deepface,chopra2005learning,kumar2011describable}, multimedia retrieval \cite{rasiwasia2010new,kang2015learning}, medical imaging \cite{shen2010active,wernick2010machine,greenspan2016guest,oakden2020hidden}, time-series forecasting \cite{ding2015deep,luo2018beyond,ma2019position} and so on. These tasks usually require a large amount of labeled examples to train the model and then to make inferences on testing examples. 
Generally, in most cases, the larger the quantity of data, the better the model performance. Taking ImageNet \cite{deng2009imagenet} as an example, which consists of 21,841 classes and 14,197,122 images in total, its emergence has brought unprecedented opportunities to the machine learning and multimedia areas. Many tasks trained on large datasets, e.g., ImageNet, and related tasks, e.g., pre-trained models on ImageNet, continue to make progress in contrast to many other areas. These tasks have achieved superior performance and even surpassed humans in some scenarios \cite{he2015delving}. 
Despite the great power of supervised learning, it relies too much on a large amount of labeled data examples, which makes it difficult for models to be generalized to unfamiliar or even unseen classes. \textbf{\textit{Transfer Learning}} \cite{pan2009survey,torrey2010transfer,long2017deep,jia2018transfer} has partially solved this problem. It pre-trains a model on a source domain dataset of a similar task and then transfers the whole or part of the trained model to the target domain task and fine-tunes with target data. For example, it is easier for a learner who has already learned English to learn French because many internal similarities and overlaps exist between these two languages. Human beings have an excellent ability to generalize learned knowledge to explore the unknown. It has been proven that humans can recognize over 30,000 object classes and many more subclasses \cite{biederman1987recognition}, e.g., breeds of birds and combinations of attributes and objects. Moreover, human beings are also very good at recognizing object classes without previously seeing them before. For example, if a learner who has never seen a panda is taught that the panda is a bear-like animal that has black and white fur, then he or she will be able to easily recognize a panda when seeing a real example. 
In machine learning, this process is considered as the problem of \textbf{\textit{Zero-shot Learning}} \cite{lampert2009learning}. The settings of zero-shot learning can be regarded as an extreme case of transfer learning: the model is trained to imitate human ability in recognizing examples of unseen classes that are not shown during training stage \cite{frome2013devise,lampert2014attribute,shigeto2015ridge,wang2015zero,bucher2016improving,zhang2016zero,changpinyo2016synthesized,kodirov2017semantic,zhu2018generative,guo2019ams}. 
In conventional supervised learning, the training and testing examples belong to the same class-set, which means that the learned model has already seen some examples of all the classes it encounters during testing. In contrast, the zero-shot learning only trains the model on seen class examples, and the learned model is expected to infer novel unseen class examples. Thus, as shown in Figure \ref{zsl_setting}, the essential difference between the zero-shot learning and conventional supervised learning is that the training and testing class-sets of zero-shot learning are disjoint from each other. As a result, the zero-shot learning can be regarded as a complement to the conventional supervised learning, and gaps the scenarios where collecting and labeling a large amount of examples for all classes is impossible in real-world applications. As such, the zero-shot learning has received increasing attention in recent years \cite{frome2013devise,lampert2014attribute,shigeto2015ridge,wang2015zero,bucher2016improving,zhang2016zero,changpinyo2016synthesized,kodirov2017semantic,zhu2018generative,guo2019ams}.

\begin{figure}[t]
    \centerline{\includegraphics[width=0.87\textwidth]{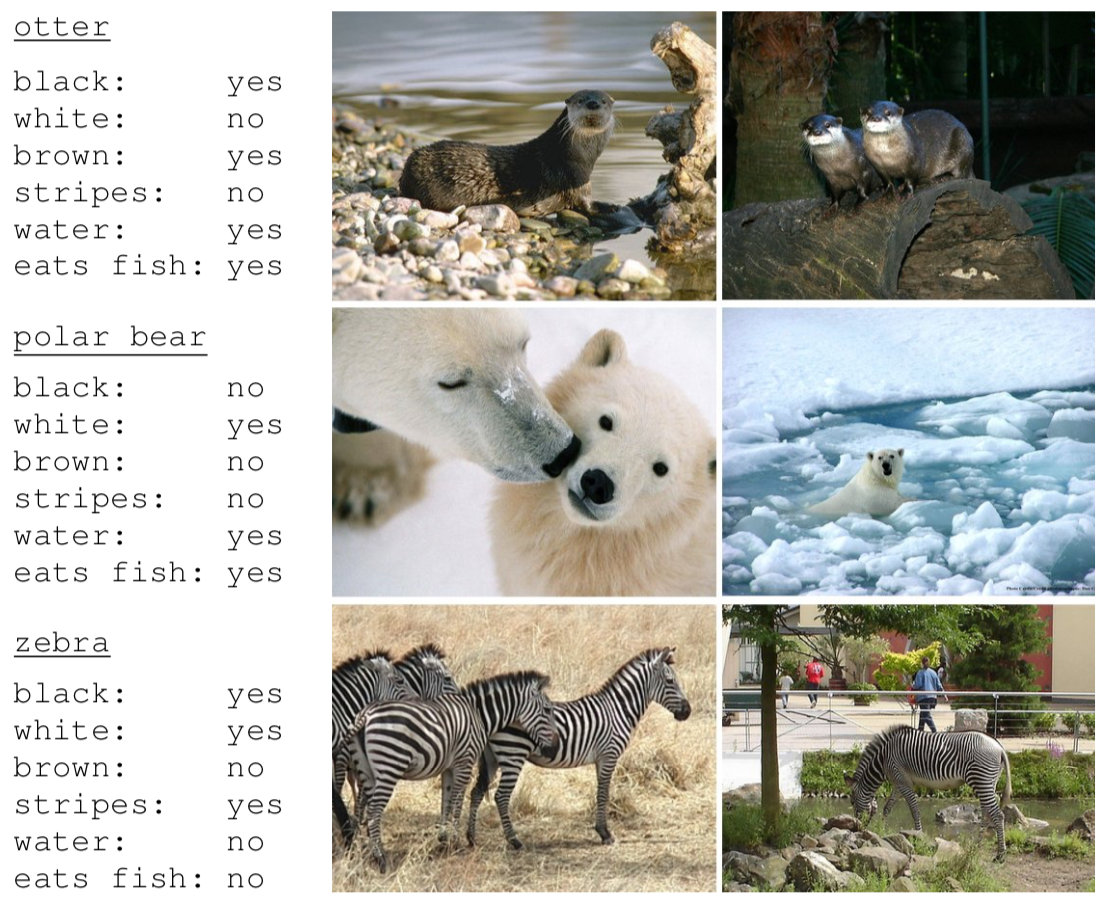}}
    \caption{A brief demonstration of semantic attributes.}
    \label{attribute}
\end{figure}

To recognize zero-shot classes (i.e., novel class examples not shown during training) in the target domain, one has to utilize the knowledge learned from source domain. Unfortunately, in the zero-shot setting, since there are no examples available during training phase on the target domain, it may be difficult for existing methods to do the domain adaptation. Thus, The key idea to achieve the zero-shot recognition is to discover and exploit the knowledge of how an unseen class can be related to seen classes. In zero-shot learning, this is typically achieved by taking utilization of labeled seen class examples and certain \textbf{\textit{Common Knowledge}} that can be shared and transferred between seen and unseen classes. This common knowledge is per-class, semantic and high-level features for both the seen and unseen classes \cite{lampert2009learning}, which enables easy and fast implementation and inference to zero-shot learning. Among them, the \textit{Semantic Attributes} and \textit{Semantic Word Vectors} become the most popular ones in recent years. 
Taking the semantic attributes as an example. The attributes are meaningful high-level descriptions of examples, such as their shapes, colors, components, texture, etc. Intuitively, the cat is more closely related to the tiger than to the snake. In the semantic feature space, this intuition also holds: the similar classes should have similar patterns. To this end, in zero-shot learning, each class can be represented by a semantic feature vector. This particular pattern is called semantic prototype and each class is endowed with a unique prototype. Figure \ref{attribute} demonstrates a simple case of semantic attributes (binary attributes). For example, the animal that has black and brown fur, lives in/near the water, and eats fish can be recognized as ``otter''. 
Thus, to handle the zero-shot recognition, one can construct a link from the original feature space, e.g., visual feature space of image examples, to the semantic feature space, to establish the cross-modal retrieval.

Under the most commonly adopted \textit{visual-semantic mapping} paradigm, the zero-shot learning task can be formalized as follows. Given a set of labeled seen class examples $\mathcal{D}=\left \{ x_{i}, y_{i} \right \}_{i=1}^{N}$, where $x_{i}$ is seen class example, i.e., image, with class label $y_{i}$ belonging to $m$ seen classes $C=\left \{ c_{1}, c_{2}, \cdots , c_{m}\right \}$. 
The goal is to construct a model for a set of unseen classes ${C}' = \left \{ {c}'_{1}, {c}'_{2}, \cdots , {c}'_{v}\right \}$ ($C \bigcap {C}' = \phi$) from which no example is available during training.
In the inference phase, given a testing unseen class example ${x}'$, the model is expected to predict its class label $c({x}')\in{C}'$. To this end, some common knowledge, e.g., the semantic features, denoted as $s = \left ( a_{1}, a_{2}, \cdots , a_{n} \right )\in \mathbb{R}^{n}$, are needed to bridge the gaps between the seen and unseen classes. Therefore, the labeled seen class examples $\mathcal{D}$ can be further specified as $\mathcal{D}=\left \{ x_{i}, y_{i}, s_{i} \right \}_{i=1}^{N}$. Each seen class $c_{i}$ is endowed with a semantic prototype $p_{c_{i}}\in \mathbb{R}^{n}$, and similarly, each unseen class ${c_{i}}'$ is also endowed with a semantic prototype $p_{{c_{i}}'}\in \mathbb{R}^{n}$. Thus, for each seen class example we have its semantic features $s_{i} \in P=\left \{p_{c_{1}}, p_{c_{2}}, \cdots , p_{c_{m}}  \right \}$, while for testing unseen class example ${x}'$, we need to predict its semantic features ${s}' \in \mathbb{R}^{n}$ and set the class label by searching the most closely related semantic prototype within ${P}'=\left \{p_{{c_{1}}'}, p_{{c_{2}}'}, \cdots , p_{{c_{v}}'}  \right \}$. In summary, given $\mathcal{D}$, the training can be described as:
\begin{equation}
	\mathop{\arg\min}_{W} \ \frac{1}{N}\cdot \sum_{i=1}^{N} \mathcal{L} \left ( f\left ( \phi \left ( x_{i} \right ); W \right ), s_{i} \right ) + \varphi \left ( W \right ),
\label{intro:loss}
\end{equation}
where $\mathcal{L}\left ( \cdot  \right )$ being the loss function and $\varphi \left ( \cdot \right )$ being the regularization term (if needed). The $f\left ( \cdot; W  \right )$ is a mapping function with parameter $W$ maps from the visual feature space to semantic feature space. The $\phi \left ( \cdot  \right )$ is a feature extractor, e.g., a pre-trained CNNs, to obtain the visual features of $x_{i}$. For the inference, given a testing example, e.g., $x_{test}$, the recognition can be described as:
\begin{small}
\begin{align}
&c (x_{test}) = \mathop{\arg\max}_{j} \ Sim \left ( f\left ( \phi \left ( x_{test} \right ); W \right ), {{P}'}^{(j)} \right ), 
\label{intro:reco_czsl}\\ 
&c (x_{test}) = \mathop{\arg\max}_{j} \ Sim \left ( f\left ( \phi \left ( x_{test} \right ); W \right ), \left \{ {{P}' \cup P} \right \}^{(j)} \right ), 
\label{intro:reco_gzsl}
\end{align}
\end{small}
where $Sim \left (\cdot \right )$ is a similarity metric and $c (x_{test})$ searches the most closely related prototype and set the class corresponding with this prototype to $x_{test}$. Specifically, Eq. \ref{intro:reco_czsl} is used for conventional zero-shot learning which the similarity search is only on unseen classes, and Eq. \ref{intro:reco_gzsl} is used for generalized zero-shot learning which the search can also generalize to novel examples from seen classes. 

In view of the above formalization that the zero-shot learning is indeed a cross-modal mapping problem, i.e., training and reusing the mapping function between the visual and semantic feature spaces, and in view of the absence of unseen classes during training, the \textbf{\textit{Domain Shift}} problem \cite{fu2015transductive} may easily occur. 
The domain shift problem refers to the phenomenon that when mapping an unseen class example from the visual to semantic feature spaces, the obtained semantic features may easily shift away from its real prototype. The domain shift problem is essentially caused by the nature of zero-shot learning: the training (seen) and testing (unseen) classes are mutually disjoint from each other. Thus, the mapping function learned from source domain (seen classes) may not well-adopted to the target domain (unseen classes). Moreover, since the visual and semantic feature spaces are generally independent and may exist in entirely different manifolds, the trained visual-semantic mapping function can hardly reach a stable convergence. Therefore, when the learned mapping function is transferred and reused by unseen classes, it may further exacerbate the shift phenomenon. 

Being an open issue in zero-shot learning, the domain shift problem has been hindering the performance of zero-shot learning models for a long time. 
Moreover, based on our observation, some key building blocks of zero-shot learning, e.g., the semantic feature space itself and its correlation with surrounding elements such as the visual feature space, does not seem to receive comparable attention. 
Therefore, in this thesis, we aim at exploring effective ways to fully empower the semantic feature space towards better performance on mitigation of the domain shift problem, which is rarely studied by previous work. The research of this thesis comprises three parts, and basically follows a bottom-up progressivism. In the first part, we consider to adaptively adjust the semantic feature space to enhance the model's ability to adapt and generalize to unseen classes. In the second part, we consider to expand the semantic features and to conduct an alignment of manifold structures between the visual and semantic feature spaces. This approach is more conservative compared to the first method for not directly adjusting the previous semantic features. In the third part, we take a further step by considering the correlation between the visual and semantic feature spaces in a more fine-grained perspective. This approach makes use of the graph techniques to better exploit the visual-semantic correlation and the local substructure information in examples.

\section{Thesis Contributions}
\label{sec:intro:contributions}
We briefly summarize our contributions below.

\begin{itemize}
 \item [1. ] \textbf{Rectification on Class Prototype and Global Distribution}.\\
To deal with the domain shift and hubness problems, we propose a novel zero-shot recognition framework to adaptively adjust to rectify the semantic feature space. This adjustment is based on both the prototypes and the global distribution of data during each loop of training phase. To the best of our knowledge, our work is the first to consider such adaptive adjustment of semantic feature space in zero-shot learning. Moreover, we also propose to combine the adjustment jointly with a cycle mapping, which first maps the examples from visual to semantic feature space and then vice versa, guaranteeing the mapping function obtains more robust results to further mitigate the domain shift problem. We formulate our solution to a more efficient framework which can significantly boosts the training. Experimental results on several benchmark datasets demonstrate the significant performance improvement of our method compared with other existing methods on both recognition accuracy and training efficiency.

\item [2. ] \textbf{Manifold Structure Alignment by Semantic Feature Expansion}.\\
To address the domain shift problem, we propose a novel model to implicitly align the manifold structures between the visual and semantic feature spaces by expanding the semantic feature space. Specifically, we build upon an autoencoder-based model that takes the visual features as input to generate/expand several auxiliary semantic features for each prototype. We then combine these auxiliary and predefined semantic features to discover better adaptability for the semantic feature space. This adaptability is mainly achieved by aligning the manifold structure of the combined semantic feature space, to an embedded manifold structure extracted from the original visual feature space of data. To simultaneously supervise the expansion and alignment phases, we propose to combine the reconstruction term with an alignment term within the autoencoder-based model. Our model is the first attempt to align both feature spaces by expanding semantic features and derives two benefits: first, we expand some auxiliary features that enhance the semantic feature space; second and more importantly, we implicitly align the manifold structures between the visual and semantic feature spaces, which resulting in a more robust well-matched visual-semantic mapping function and can better mitigates the domain shift problem. Extensive experiments show remarkable performance improvement and verifies the effectiveness of our method.

 \item [3. ] \textbf{Zero-shot Learning as a Graph Recognition}.\\
To take a further step on mitigating the domain shift problem, our interest is to focus on the fine-grained perspective. We propose a fine-grained zero-shot learning framework based on the example-level graph. Specifically, we decompose an image example into several parts and use a graph-based model to measure and utilize certain relations between these parts. Taking advantage of recently developed graph neural networks (GNNs), we formulate the zero-shot learning to a graph-to-semantic mapping problem which converts the zero-shot recognition to a graph recognition task. Our method can better exploit part-semantic correlation and local substructure information in examples, and makes the obtained visual-semantic mapping function more robust and accurate. Experimental results demonstrate that the proposed method can mitigate the domain shift problem and achieve competitive performance against other representative methods. 
\end{itemize}

\section{Thesis Organization}
\label{sec:intro:organization}
The rest of this thesis consists of five chapters and organized as follows.
\begin{itemize}
\item \textbf{\Cref{ch:2}}\\
This chapter reviews the background knowledge and some related works of this thesis. First, we briefly introduce the semantic feature space which is critical for zero-shot learning in \cref{sec:review:semantic}, including the categories of semantics, pros and cons. Next, we explain and compare the learning frameworks and datasets of the zero-shot learning in \cref{sec:review:zsl}. Last, we introduce the domain shift problem which is the open issue and main challenge in zero-shot learning. 

\item \textbf{\Cref{ch:3}}\\
This chapter presents our study on the adaptive adjustment and rectification of semantic feature space for zero-shot learning, which can helps to mitigate the domain shift and hubness problems. In particular, \cref{c3:Related_Work} introduces some related work. In \cref{c3:Methodology}, we explain our proposed method regarding the adjustment on class prototypes and global distribution, and the formulation to a more efficient unified training framework. The experiments including the results and analysis are addressed in \cref{c3:Experiments}.

\item \textbf{\Cref{ch:4}}\\
This chapter presents our study on the alignment of manifold structures between the visual and semantic feature spaces, which is typically achieved by expanding semantic features. In particular, we introduce some related work in \cref{c4:Related_Work}. Then \cref{c4:Methodology} explains our proposed method including the semantic feature expansion, manifold extraction and alignment, etc. The experimental results and analysis are addressed in \cref{c4:Experiments}.

\item \textbf{\Cref{ch:5}}\\
This chapter presents our study on the conversion of zero-shot learning to the graph recognition task, which is a fine-grained zero-shot learning framework based on the example-level graph. In particular, \cref{c5:Related_Work} introduces some related work. In \cref{c5:Methodology}, we explain our proposed method including the parts decomposition, example graph construction and example graph recognition. In \cref{c5:Experiments}, we introduce the experimental results and analysis.

\item \textbf{\Cref{ch:6}}\\
This chapter summarizes this thesis and provides some potential future research directions.
	
\end{itemize}

\chapter{Background Review}
\label{ch:2}
This chapter reviews some background knowledge and related work. We first introduce the semantic feature space which is the key building block of zero-shot learning in Section \ref{sec:review:semantic}. Then we introduce the learning framework, models and datasets of zero-shot learning in Section \ref{sec:review:zsl}. Last, we explain the domain shift problem, which is the main challenge in Section \ref{sec:review:domain}.

\section{Semantic Feature Space}
\label{sec:review:semantic}
The semantic feature space is critical for zero-shot learning. Since the unseen class examples are not available during training, it may be difficult for existing methods to do the domain adaptation to transfer learned knowledge from seen to unseen classes. Thus, some common knowledge is required to discover and model how an unseen class can be related to seen classes. In zero-shot learning, this common knowledge is considered as the \textit{semantic feature space} which is shared by both seen and unseen classes \cite{lampert2009learning}. The corresponding semantics (a.k.a., semantic features) are generally per-class high-level features describing a particular class. In zero-shot learning, each class is endowed with a unique semantic feature representation which is called semantic prototype.

\begin{figure}[t]
    \centerline{\includegraphics[width=0.99\textwidth]{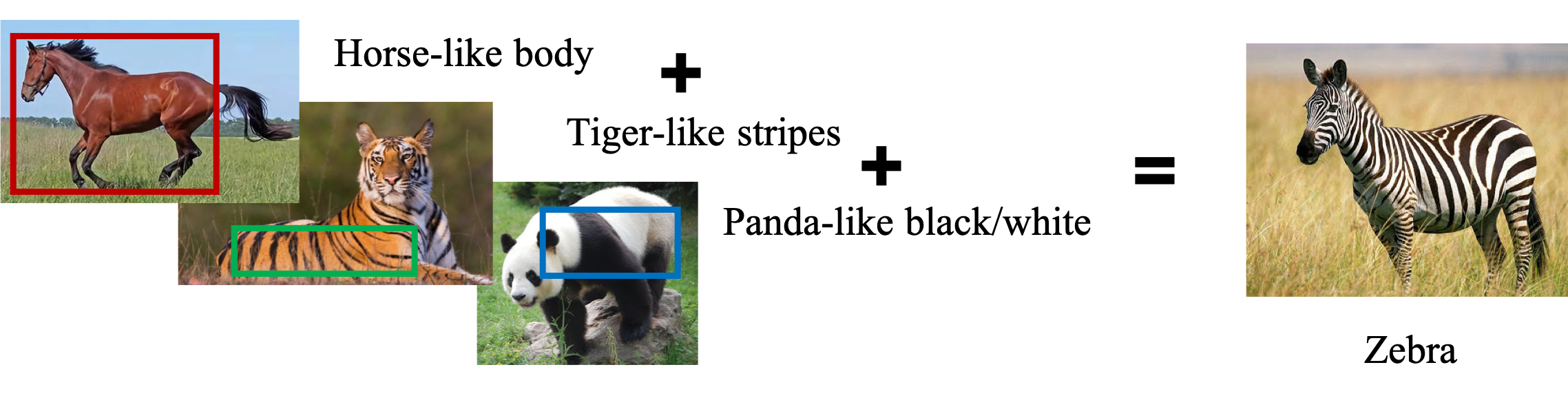}}
    \caption{How semantic feature space is shared between the zebra and \{horse, tiger, panda\}.}
    \label{sfs}
\end{figure}

The semantic features work as a bridge between seen and unseen classes. By sharing a common feature space, each class can be easily implemented and connected to other classes if exist. For example, as shown in Figure \ref{sfs}, the ``zebra'' can be recognized as an animal with a ``horse''-like body, ``tiger''-like stripes, and ``panda''-like black and white color appearance. Because the semantic features are global high-level descriptions which are semantically meaningful to each class, thus, intuitively, for any descriptions the cat is more closely related to the tiger than to the snake. This intuition should be also held in the semantic feature space, which means that under certain metrics, e.g., cosine similarity, the cat prototype should be more similar to the tiger prototype than to the snake prototype, which resulting similar classes reside nearby in the semantic feature space.

\subsection{Categories of Semantics}
The semantic features can be roughly grouped into two mainstreams based on either supervised or unsupervised settings. For the supervised setting, the semantic attributes are the most popular semantics and are adopted in various zero-shot learning models \cite{farhadi2009describing,lampert2009learning,patterson2014sun,wah2011caltech}. The attributes are usually generated/defined and verified by human experts, which is some kind of laborious task. For the unsupervised setting, the semantic word vectors are usually learned from some large-scale unannotated linguistic knowledge corpus, e.g., from Wikipedia, news, etc. Such kind of word embeddings are widely used in natural language processing problems and can be efficiently extended to zero-shot learning. Among them, word2vec \cite{mikolov2013efficient,mikolov2013distributed}, FastText \cite{joulin2017bag} and GloVe \cite{pennington2014glove} vectors are most frequently used \cite{xian2016latent,zhang2016fast,bansal2018zero,guo2020novel}. Now, we elaborate on different categories of semantic features as follows.

\textbf{Attribute:} In the attribute space, a list of human understandable characteristics describing various properties of the classes are defined as attributes. The attributes are meaningful high-level common descriptions of classes, where each attribute is usually a word or a phrase corresponding to one particular property of these classes. For example, in an animal image recognition problem, the attributes can be defined as various properties of animals such as their body colors (e.g., ``black'', ``white'', ``brown'', etc.), their habitats (e.g., ``water'', ``forest'', ``desert'', etc.), their components (e.g., ``stripes'', ``spots'', ``lumps'', etc.) and so on. These attributes are then used to form the semantic feature space which is shared and can transfer learned knowledge between the seen and unseen classes. For the semantic prototype of each class, the values of each dimension are determined by whether the class has a corresponding attribute. Taking the most simple animal recognition problem as the example, suppose there are 6 attributes, i.e., ``black'', ``white'', ``brown'', ``stripes'', ``water'', and ``eats fish'', describing each class. In this attribute space, the ``otter'' can be recognized as one kind of animal who has black and brown fur, lives in/near the water, and eats fish. Thus, the ``otter'' prototype can be presented by a semantic attribute vector as $\left [ 1, 0, 1, 0, 1, 1 \right ]$, where each dimensional binary value ``0/1'' indicates whether the corresponding attribute exists or not. The \textit{binary attribute space} is the most simple case of semantic feature space. In general, the attribute can also be real numbers indicating the degree or confidence of a class having a corresponding attribute, which is the \textit{continuous attribute space}. Moreover, there also exist the \textit{relative attribute space} which can measure the relative degree or confidence of having an attribute among different classes. Given an input example image, the traditional recognition models directly classify the label or name of the example, without paying any attention to various attributes. To solve this problem, several investigations have been made to predict the class attribute before classifying the labels or names \cite{farhadi2009describing,lampert2009learning,wah2011caltech,patterson2014sun}. By using the attributes, we can develop more advanced models to describe, compare, and classify examples easily in a human understandable format. Moreover, the attributes can also help to classify novel or unseen class by the shared attribute space where the certain combinations of known attributes may become an unseen class \cite{lampert2014attribute,xian2018zero}.

\begin{figure}[t]
    \centerline{\includegraphics[width=0.8\textwidth]{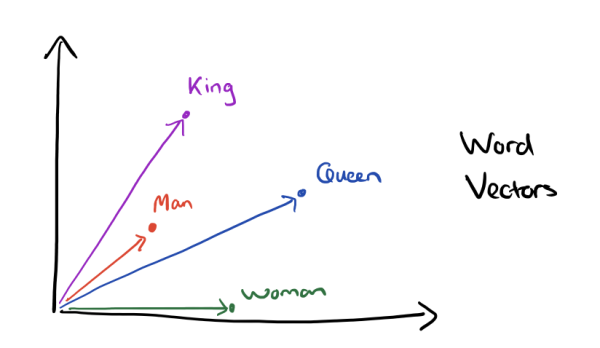}}
    \caption{Algebraic operations within word vectors (e.g., word2vec) can also demonstrate linguistic regularities.}
    \label{word2vec}
\end{figure}

\textbf{Word2vec:} The word2vec is a continuous-valued high-dimensional vector representation of a linguistic word of a vocabulary \cite{mikolov2013efficient, mikolov2013distributed}. 
They are learned from billions of words. Moreover, the vocabulary collated in this manner contains millions of words itself. 
Word vector is first investigated in the area of natural language processing (NLP). Two basic architectures named continuous bag-of-words (CBOW) and continuous skip-gram are proposed to generate word2vec vectors \cite{mikolov2013efficient}.
Both architectures contain a two-layer neural network with non-linear activation. Within an input sentence, CBOW defines a word as the current word. Considering the words adjacent to our word of interest in a sentence as input to the network, the method aims to predict the current word. 
In contrast, the skip-gram method aims at predicting the previous and future word based on the input of the current word. In this way, the word2vec vectors are learned by considering the similarity of the word with the context of the descriptions. 
Therefore, algebraic operations within word2vec vectors can also demonstrate linguistic regularities (as demonstrated in Figure \ref{word2vec}). For example, if we subtract the vector ``Man'' from the vector ``king'', and add the vector "woman", the resulting vector should be close to the vector ``queen'': 
\begin{equation}
	V_{King} - V_{Man} + V_{Woman} \approx V_{Queen}.
\end{equation}
By using the word2vec as the semantic feature space, the above advantages can be easily extended to zero-shot learning by mapping examples from the visual to the word2vec embedding space, and be recognized by searching the mostly closely related prototypical vector.

\textbf{FastText:} The FastText \cite{joulin2017bag} is an extended version of Word2Vec. It is an open-sourced library from Facebook containing pre-trained models of word vectors of 294 languages. The key difference from word2vec is that FastText breaks a word into several character n-grams or sub-words. So that the vector for a word is made of the sum of this character n-grams. For example, the word vector ``apple'' can be the sum of the vectors of the n-grams: ``ap'', ``app'', ``appl'', ``apple'', ``apple'', ``ppl'', ``pple'', ``pple'', ``ple'', ``ple'' and ``le''.
In contrast to the word2vec which learns vectors only for complete words found in the training corpus. FastText aims to learn vectors for the n-grams that are found within each word, as well as each complete word. 
At each training step in FastText, the mean of the target word vector and its component n-gram vectors are used for training. The adjustment that is calculated from the error is then used uniformly to update each of the vectors that were combined to form the target. The above process brings about more calculation resource while the resulting vectors have been shown to be more accurate than word2vec by different metrics.

\textbf{GloVe:} The GloVe \cite{pennington2014glove} is an another important type of word representation. It is generated using a log-bilinear model trained on the word to word co-occurrence statistics of a given corpus. The training tries to match the dot product of any given pair of the vocabulary word vectors to the logarithm co-occurrence probability of that pair. GloVe vectors are especially useful for the word analogy task.

\subsection{Semantic Pros and Cons}

\begin{figure}[t]
\centering
\begin{minipage}[t]{0.49\textwidth}
\centering
\includegraphics[width=6.7cm]{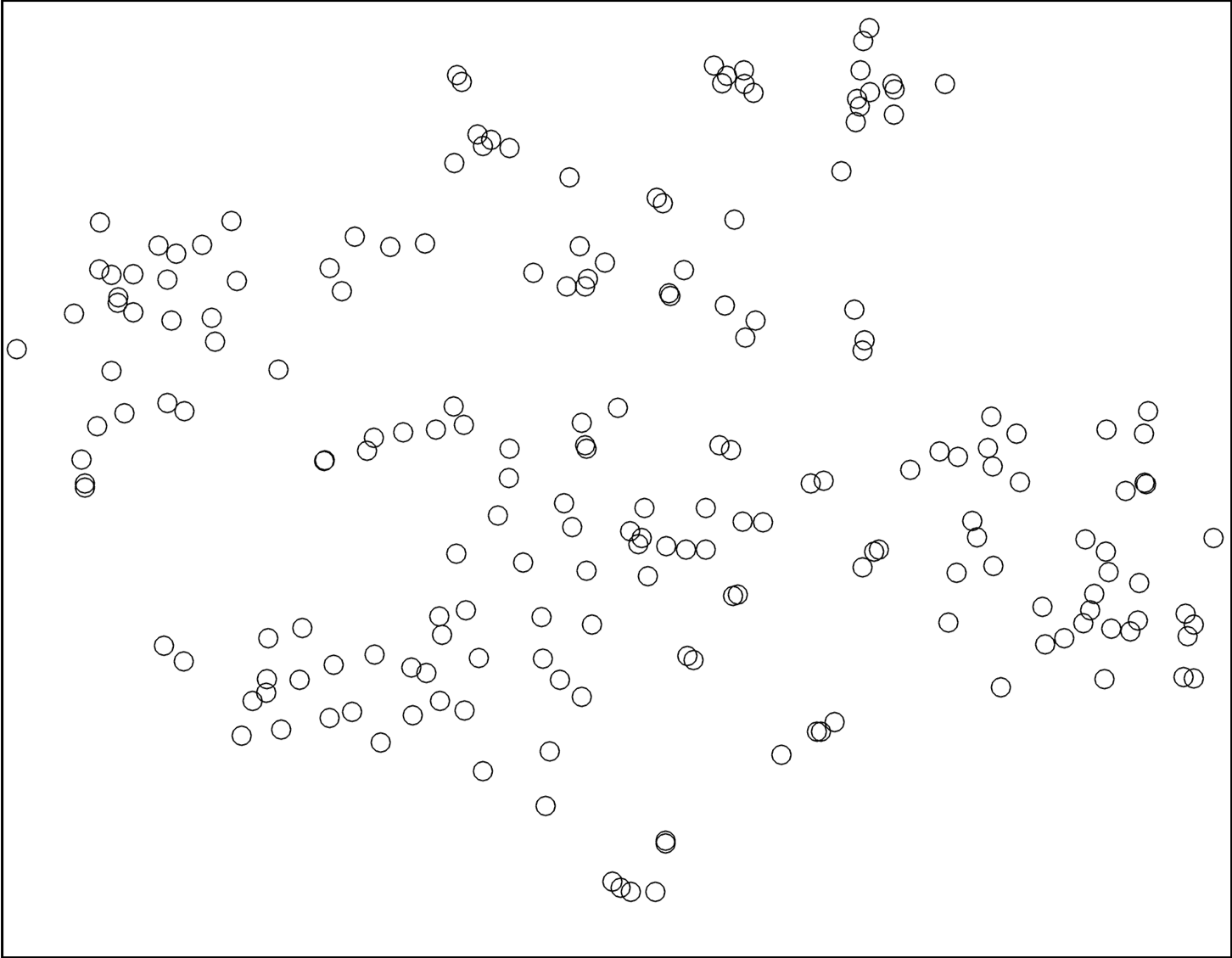}
\caption{2D tSNE plot of semantic \\attributes.}
\label{tsne_cub}
\end{minipage}
\begin{minipage}[t]{0.49\textwidth}
\centering
\includegraphics[width=6.7cm]{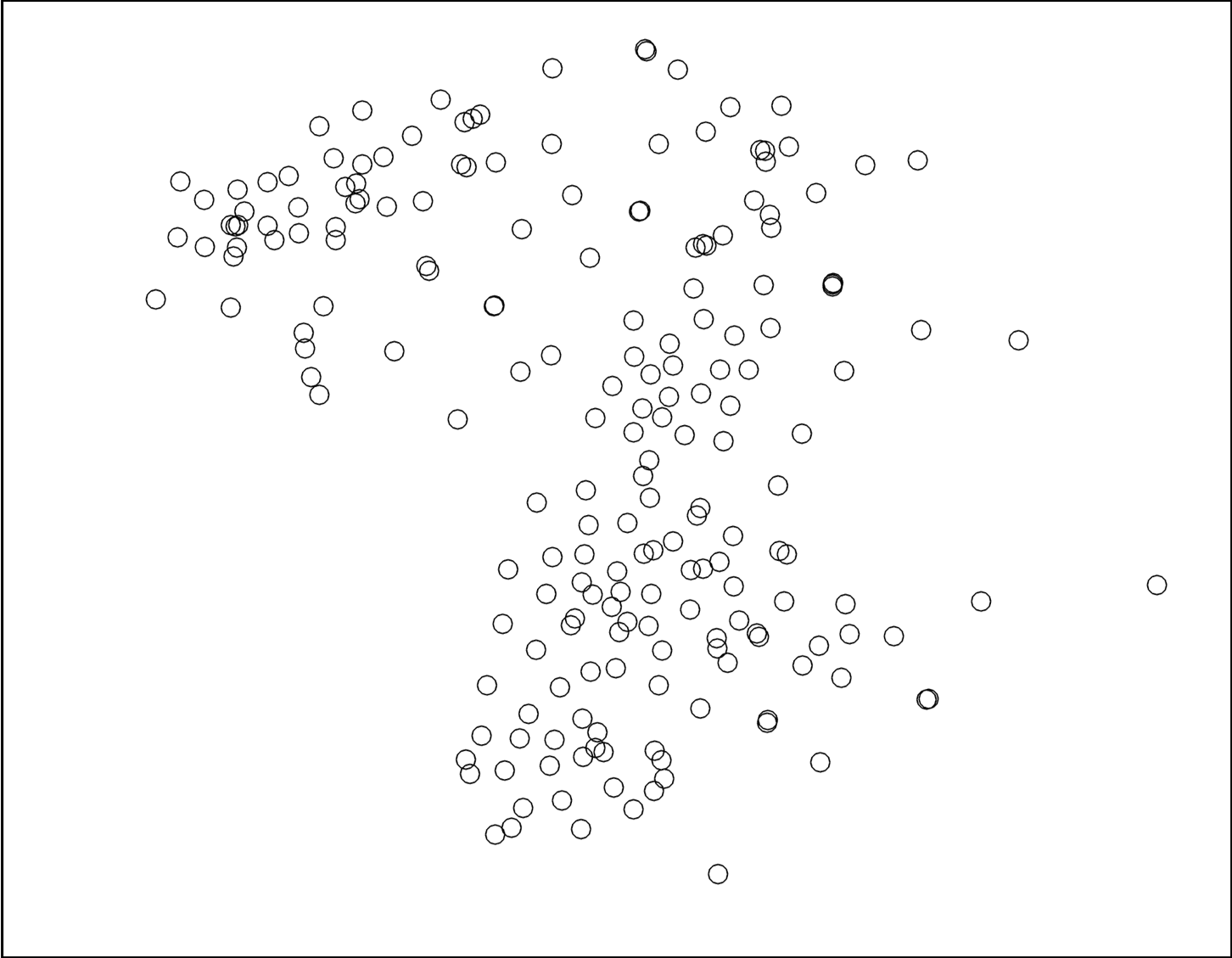}
\caption{2D tSNE plot of semantic word vectors (e.g., word2vec).}
\label{tsne_imgnet}
\end{minipage}
\end{figure}

To generate the semantic features of any class, we can either use the supervised or unsupervised settings. 
The supervised features are the attributes that are usually generated/defined and verified manually by human experts \cite{wah2011caltech,lampert2014attribute}. Thus, the attributes are more capable of describing each class with less noise and obtains a more accurate semantic prototype for object classes. Being widely used in various zero-shot learning methods, however, since the attributes require considerable human efforts to acquire annotations, the process is usually costly. As a workaround, the unsupervised features, i.e., word vectors, can be generated automatically from a large corpus of unannotated text, e.g., Wikipedia, news, etc., or the hierarchical relationship of classes in WordNet corpus \cite{miller1995wordnet}. Some widely used examples of such semantic features are word2vec \cite{mikolov2013efficient,mikolov2013distributed}, FastText \cite{joulin2017bag} and GloVe \cite{pennington2014glove}. Since these vectors are generated in an unsupervised manner, there may exist more noisy components compared to the attributes. Despite the defect, the word vectors are more flexible and can provide more scalability compared to the manually acquired attributes.
We visualize the semantic attributes and semantic word vectors based on CUB-200-2011 dataset \cite{wah2011caltech} in Figure \ref{tsne_cub} and Figure \ref{tsne_imgnet}, respectively. It can be observed that these semantic prototypes are clustered better in attribute space than word vector space. In this thesis, our works are based on both attributes and word vectors to implement our methods. Specifically, we mainly use the attributes for small and medium scale experiments, and use the word vectors for large scale experiment.
 
\section{Zero-shot Learning} 
\label{sec:review:zsl}
\begin{figure}[t]
    \centerline{\includegraphics[width=0.99\textwidth]{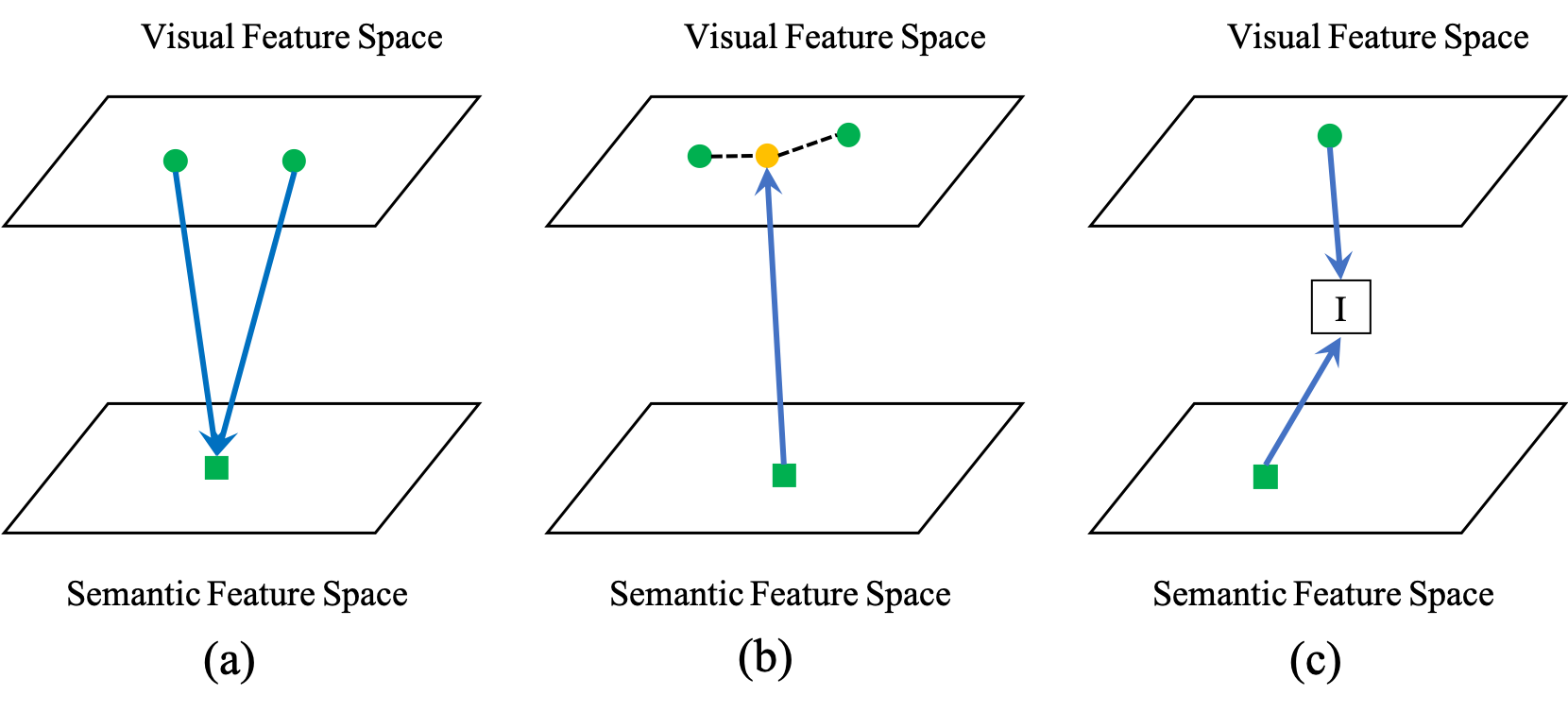}}
    \caption{Three learning frameworks of zero-shot learning: (a) Forward Mapping maps from visual to semantic feature spaces, (b) Reverse Mapping maps from semantic to visual feature spaces, and (c) Intermediate Mapping maps from both feature spaces to an intermediate one.}
    \label{mappings}
\end{figure}

As stated, one can construct a mapping function between the visual and semantic feature spaces to establish the cross-modal retrieval for zero-shot leaning task. The key is to map examples and classes into the same latent space and to perform the similarity search (e.g., nearest neighbor search). Based on the mapping directions, the main learning frameworks of zero-shot learning can be grouped into three ways: Forward Mapping, Reverse Mapping, and Intermediate Mapping (Figure \ref{mappings}). We now elaborate different learning frameworks including the above three typical ones and others, and introduce the benchmark datasets of zero-shot learning.

\subsection{Learning Framework}
\textbf{Forward Mapping:} The forward mapping is the most widely used mapping in zero-shot learning. As shown in Figure \ref{mappings}.(a), it refers to finding a mapping function that maps from the visual feature space to the semantic feature space. 
SOC \cite{palatucci2009zero} maps the visual features to the semantic feature space and then searches the nearest class embedding vector. 
\textit{Akata et al.} \cite{akata2013label} proposed to view attribute-based image classification as a label-embedding problem, and introduce a function which measures the compatibility between an image and a label embedding. 
SJE \cite{akata2015evaluation} optimizes the structural SVM loss to learn the bilinear compatibility, 
while ESZSL \cite{romera2015embarrassingly} utilizes the square loss to learn the bilinear compatibility and adds a regularization term to the objective with respect to the Frobenius norm. 
ALE \cite{akata2016label} trains a bilinear compatibility function between the semantic attribute space and the visual space by ranking loss. 
Similarly, DeViSE \cite{frome2013devise} also trains a linear mapping function between visual and semantic feature space by an efficient ranking loss formulation. 
\textit{Bucher et al.} \cite{bucher2016improving} embedded the visual features into the attribute space. Recently, 
SAE \cite{kodirov2017semantic} uses a semantic autoencoder to regularize zero-shot recognition. 
\textit{Xian et al.} \cite{xian2016latent} extended the bilinear compatibility model of SJE \cite{akata2015evaluation} to be a piecewise linear model by learning multiple linear mappings with the selection being a latent variable. 
\textit{Socher et al.} \cite{socher2013zero} used a deep learning model that contains two hidden layers to learn a nonlinear mapping from the visual feature space to the semantic word vector space \cite{mikolov2013distributed}.
Unlike other works which build their embedding on top of fixed image features, \textit{Ba et al.} \cite{lei2015predicting} trained a deep CNNs and learning a visual to semantic embedding.
SP-AEN \cite{chen2018zero} introduces an independent visual-to-semantic mapping that disentangles the semantic space into two subspaces and an adversarial-style discriminator between them to prevent the semantic loss and improve the zero-shot recognition.

\textbf{Reverse Mapping:} In contrast to forward mapping, some investigations reported the hubness phenomenon in zero-shot learning that mapping the high-dimensional visual features to the low-dimensional semantic feature space could reduce the variance of features, and the results may become more clustered \cite{dinu2014improving,shigeto2015ridge,zhang2017learning}. Thus, some researchers propose to reversely map from the semantic to visual feature spaces (Figure \ref{mappings}.(b)). 
\textit{Dinu et al.} \cite{dinu2014improving} proposed to take the proximity distribution of potential neighbours across many mapped vectors into account to correct the hubness. 
\textit{Shigeto et al.} \cite{shigeto2015ridge} analyzed the mechanism behind the emergence of hubness and proved that mapping labels into the visual feature space is desirable to suppress the emergence of hubs in the subsequent nearest neighbor search step. 
\textit{Ba et al.} \cite{ba2015predicting} and \textit{Zhang et al.} \cite{zhang2017learning} both proposed to train a deep neural network to map the semantic features to the visual feature space. 
\textit{Changpinyo et al.} \cite{changpinyo2016predicting} proposed a simple model based on a support vector regressor to map the semantic features to the visual feature space and performed nearest-neighbor algorithms. However, since each class prototype or label has various corresponding visual features in different examples, the conventional reverse mapping may still problematic. Thus, some recent works proposed to generate some examples or visual features based on the semantic features of classes, to increase the diversity of examples especially for unseen classes. 
For example, CVAE-ZSL \cite{mishra2018generative} proposes to generate various visual examples from given semantic attribute by conditional variational autoencoders (VAEs), and uses the generated examples for classification of the unseen classes. 
GFZSL \cite{verma2017simple} models each class-conditional distribution as a Gaussian and learns a regression that maps a class embedding into the latent space. 
SE-GZSL \cite{kumar2018generalized} also adopts the VAEs architecture that consisting of a probabilistic encoder and a probabilistic conditional decoder to generate novel visual examples from seen/unseen classes. 
f-CLSWGAN \cite{xian2018feature} proposes a novel generative adversarial network (GAN)-based model to generate CNNs features rather than real examples conditioned on class-level semantic features, which can offer a shortcut from a semantic descriptor of a class to a class-conditional feature distribution.

\textbf{Intermediate Mapping:} The intermediate mapping refers to finding an intermediary feature space that both the visual and semantic features are mapped to \cite{lu2015unsupervised} (Figure \ref{mappings}.(c)). The intermediate mapping can also be considered as the metric learning which learns to compare or evaluate the relatedness of a pair of visual and semantic features. 
SSE \cite{zhang2015zero} utilizes the mixture of seen class parts as the intermediate feature space; then, the examples belonging to the same class should have similar mixture patterns. 
\textit{Zhang et al.} \cite{zhang2016zero} proposed to map the visual features and semantic features to two separate intermediate spaces. 
Additionally, some researchers also proposed several hybrid models to jointly embed several kinds of textual features and visual features to ground attributes \cite{fu2015zero,akata2016multi,long2017zero,changpinyo2016synthesized}. 
\textit{Lu et al.} \cite{lu2015unsupervised} proposed to linearly combine the base classifiers trained in a discriminative learning framework to construct the classifier of unseen classes. 
RelationNet \cite{sung2018learning} proposes to take a pair of visual and semantic features as input and learns a deep metric network which calculates their similarity. 
DCN \cite{liu2018generalized} proposes to map the visual features of image examples and the semantic prototypes of classes into one common embedding space, thus the compatibility of seen classes to both the source and target classes can be maximized.

\textbf{Others:} Expect for the above three typical learning frameworks, there also exist several other methods including either earlier and recently proposed ones. 
Some previous works of ZSL usually take utilization of the attributes in a two-phase model to predict the label of an example belonging to one from unseen classes \cite{kankuekul2012online,norouzi2013zero,jayaraman2014zero,lampert2014attribute,al2016recovering}. 
In the first stage, the model predicts the attribute feature of an example. Then, in the second stage, the class label is predicted by searching the class that attains the most similar set of attributes. 
For example, DAP \cite{lampert2014attribute} proposes to first estimates the posterior of all attributes of the input visual example by learning probabilistic classifiers on attributes, and then computes the classes posteriors and infers the class labels by MAP estimate. 
CAAP \cite{al2016recovering} proposes to first learn a probabilistic classifier for each attribute and use the random forest to estimate the class posteriors. IAP \cite{lampert2014attribute} chooses to first predicts the class posterior of seen classes and then to uses the probability of every class to compute the attribute posteriors of examples. 
Additionally, the two-stage approach can also be extended to the scenario where attributes are not available. For example, based on IAP \cite{lampert2014attribute}, CONSE \cite{norouzi2013zero} chooses to first predicts the posteriors of seen classes, then it maps the visual features into the Word2vec \cite{mikolov2013distributed} feature space. Recently, some graph-based methods have been proposed to handle to zero-shot learning. 
For example, \textit{Wang et al.} \cite{wang2018zero} proposes to use graph neural networks to present and propagate information among each class. DGP \cite{kampffmeyer2019rethinking} further extends it to a dense graph propagation module to jointly consider the distant nodes. So that the model can make use of the hierarchical structural information of graphs. Meanwhile, some recent methods such as ${\rm S^2GA}$ \cite{ji2018stacked}, \textit{Zhu et al.} \cite{zhu2019semantic} and \textit{Chen et al.} \cite{chen2019learning}, which make use of attention mechanism on local regions or learn dictionaries through joint training with examples, attributes and labels are proposed to focus on a more fine-grained perspective of zero-shot learning.

In our thesis, we mainly contribute to the forward mapping, graph and fine-grained based methods. Specifically, the methods proposed in \Cref{ch:3} and \Cref{ch:4} mainly consider the forward mapping in zero-shot learning, while the method proposed in \Cref{ch:5} mainly consider graph and fine-grained based model.

\subsection{Datasets}
There are five widely used benchmark datasets for zero-shot learning including Animals with Attributes\footnote{http://cvml.ist.ac.at/AwA/} (AWA) \cite{lampert2014attribute}, CUB-200-2011 Birds\footnote{http://www.vision.caltech.edu/visipedia/CUB-200-2011.html} (CUB) \cite{wah2011caltech}, aPascal\&Yahoo\footnote{http://vision.cs.uiuc.edu/attributes/} (aPa\&Y) \cite{farhadi2009describing}, SUN Attribute\footnote{http://cs.brown.edu/~gmpatter/sunattributes.html} (SUN) \cite{patterson2014sun}, and ILSVRC2012\footnote{http://image-net.org/challenges/LSVRC/2012/index} / ILSVRC2010\footnote{http://image-net.org/challenges/LSVRC/2010/index} (ImageNet) \cite{russakovsky2015imagenet}. Among them, the first four are small and medium scale datasets, and ImageNet is a large-scale dataset. The basic description of these datasets is listed in Table \ref{datasets}.

\begin{table}[h]
    \begin{center}
        \caption{Description of datasets. The \# denotes the number, SCs/UCs denotes seen and unseen classes, and D-SF denotes the dimension of semantic features.}
        \label{datasets}
        \setlength{\tabcolsep}{6mm}{  
            \begin{tabular}{lcccc}        
                \hline                   
                Dataset         & \# Examples       & \# SCs          & \# UCs            & D-SF \\
                \hline
                AWA \cite{lampert2014attribute}             & 30475             & 40              & 10                & 85    \\ 
                CUB \cite{wah2011caltech}             & 11788             & 150             & 50                & 312   \\ 
                aPa\&Y \cite{farhadi2009describing}          & 15339             & 20              & 12                & 64    \\
                SUN \cite{patterson2014sun}             & 14340             & 645             & 72                & 102    \\
                ImageNet \cite{russakovsky2015imagenet}        & $2.54\times 10^5$ & 1000            & 360               & 1000  \\
                \hline  
        \end{tabular}
        }
     \end{center} 
\end{table}

\textbf{Animals with Attributes:} The AWA \cite{lampert2014attribute} is an animal example collection from public sources such as Flickr. It consists of 30,475 images of 50 animal classes, of which 40 of them are seen classes and the remaining 10 animals are unseen classes. In the dataset, each class is represented by an 85-dimensional numeric attribute feature vector as the prototype. By using the shared attributes, it is possible to transfer information between different classes. For the zero-shot learning setting, the 40 seen classes including 24,295 images are used for training, and the remaining 10 unseen classes associating with 6,180 images are used for testing. 

\textbf{CUB-200-2011 Birds:} The CUB \cite{wah2011caltech} is a bird example collection from Flickr image search and then further filtered by showing each image example to various users of Mechanical Turk \cite{welinder2010multidimensional}. All image examples in CUB are annotated with attribute labels, bounding boxes and part locations. The dataset consists of 11,788 images of 200 bird species, from which 150 of them are seen classes and the remaining 50 species are unseen classes. In the dataset, each class is represented by a 312-dimensional semantic attribute feature vector as the prototype. The attributes are generally visual in nature, with most pertaining to a pattern, shape, or color of a particular part. For the zero-shot learning setting, 8,855 images within 150 seen classes are used for training, and the remaining 2,933 images within 50 unseen classes are used for testing. 

\textbf{aPascal\&Yahoo:} The aPa\&Y \cite{farhadi2009describing} is a natural object example collection consists of 15,339 images. The dataset is formed by two subsets including the Pascal VOC 2008 dataset and the Yahoo dataset. Among them, the Pascal VOC 2008 dataset contains 20 classes of 12,695 natural object image examples such as people, cat, bicycle, bus, sofa, etc. Each class is presented by a 64-dimensional attribute feature vector as the prototype. The Yahoo dataset is a supplementary to the Pascal VOC 2008 dataset which consists of 2,644 natural object image examples from 12 additional classes by the Yahoo image search, such as monkey, bag, carriage, mug, etc. The annotations of the Yahoo dataset follow the same standard as the Pascal VOC 2008 dataset, each class prototype is presented as a 64-dimensional attribute feature vector. For the zero-shot learning setting, the Pascal VOC 2008 dataset is used as seen classes for training, and the Yahoo dataset is used as unseen classes for testing.

\textbf{SUN Attribute:} The SUN \cite{patterson2014sun} is a scene image example collection consists of 717 classes, such as river, ice cave, railroad, forest, etc. In this dataset, each class is presented by a 102-dimensional attribute feature vector as the prototype. These discriminative attributes are obtained by crowd-sourced human studies and cover various properties such as surface, materials, lighting, functions/affordances, and spatial envelope, etc. The SUN dataset contains total 14,340 image examples and has two commonly used standard splits including 707/10 and 645/72. In our zero-shot learning experimental settings, we consider the latter 645/72 split from which 645 classes are used as seen classes for training, and the remaining 72 classes are used as unseen classes for testing.

\textbf{ILSVRC2012 / ILSVRC2010:} The ImageNet \cite{russakovsky2015imagenet} is a large scale benchmark dataset for zero-shot learning which covers a wide range of objects from Flickr and other search engines. The dataset consists of total 1,360 classes from two parts including ILSVRC2012 and ILSVRC2010. For the zero-shot learning setting, 1,000 classes containing $2.0\times 10^5$ image examples from the ILSVRC2012 are used as seen classes for training, and the remaining 360 classes containing $5.4\times 10^4$ image examples from ILSVRC2010 which are not included in ILSVRC2012 are used as the unseen classes for testing. The prototype of each class is presented by a 1,000-dimensional semantic word vector trained by a skip-gram text model on a corpus of 4.6M Wikipedia documents.

\section{Domain Shift Problem}
\label{sec:review:domain}
\begin{figure}[t]
    \centerline{\includegraphics[width=0.87\textwidth]{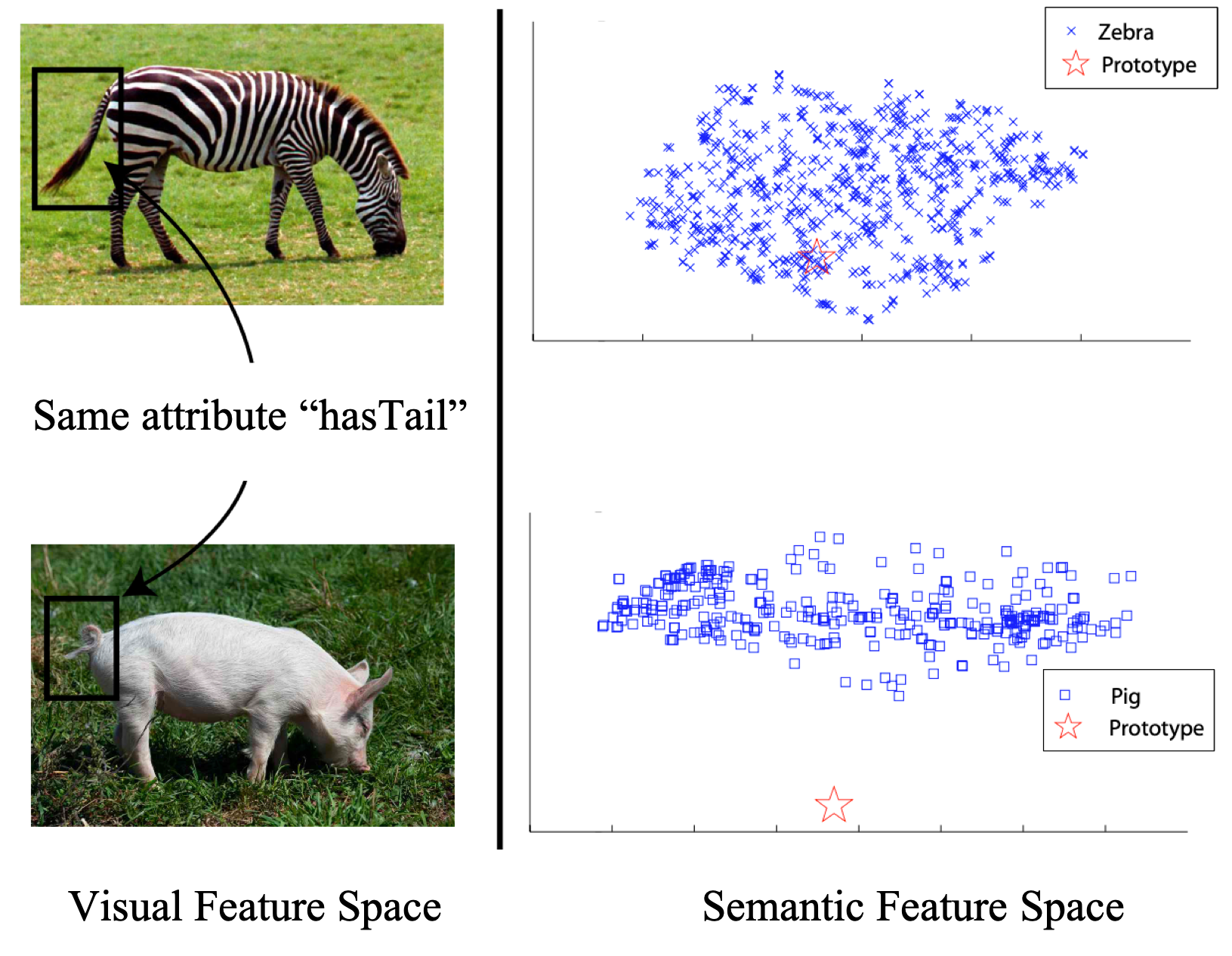}}
    \caption{Illustration of domain shift problem. The semantic prototypes are shown as red stars, and the mapped semantic features of visual examples are shown as blue rectangles.}
    \label{domain_shift}
\end{figure}

The domain shift problem \cite{fu2015transductive} is an open issue in zero-shot learning which refers to the phenomenon that when mapping an unseen class example from the visual to semantic feature spaces, the obtained semantic features may easily shift away from its real prototype. As shown in Figure \ref{domain_shift}, in the semantic feature space, both the ``Zebra'' and ``Pig'' are endowed with the same attribute property ``hasTail''. However, their visual ``Tail'' appearances are considerably different from each other in the visual feature space. Thus, if we transfer the mapping function learned from the ``Zebra'' to infer the ``Pig'' examples, the obtain semantic features of these ``Pigs'' may not reside or even far away from the real ``Pig'' prototype. 
The domain shift problem is essentially caused by the nature of zero-shot learning: the training (seen) and testing (unseen) classes are mutually disjoint from each other. 
Moreover, the visual and semantic feature spaces are generally independent. Specifically, the visual features represented by high-dimensional vectors are usually not semantically meaningful, and the semantic features represented by attributes or word vectors are not visually meaningful as well. Thus, these two feature spaces may exist in entirely different manifolds, and makes it difficult to obtain a well-matched mapping function between them. This visual-semantic gap can further exacerbate the domain shift problem and weakens the performance of zero-shot recognition models.

Recently, several investigations have been made to mitigate the domain shift problem including inductive learning-based methods, which enforce additional constraints from the training data \cite{fu2015zero,changpinyo2016synthesized}, and transductive learning-based methods, which assume that the unseen class examples or their visual features (unlabeled) are also available during training \cite{fu2015transductive,li2017zero,song2018transductive}. 
It should be noted that the model performance of the transductive paradigm is generally better than that of the inductive paradigm, because of the utilization of extra information from unseen classes during training can naturally mitigate the domain shift problem. 
However, the transductive paradigm does not fully comply with the zero-shot setting: no unseen class examples are available during training. 
With the popularity of generative adversarial networks (GANs), some generative-based methods have also been proposed recently. GANZrl \cite{tong2018adversarial} applied GANs to synthesize examples with specified semantics to cover a higher diversity of seen classes. In contrast, GAZSL \cite{zhu2018generative} leverages GANs to imagine unseen classes from text descriptions. Despite the progress made, however, since most existing methods inherently lack enough adaptability to the visual-semantic correlation when constructing the mapping function, the domain shift problem is still an open issue hindering the further development of zero-shot learning. In this thesis, we focus on fully empowering the semantic feature space, one of the key building block, to explore effective ways towards better performance on mitigation of the domain shift problem.

\chapter{Rectification on Class Prototype and Global Distribution}\label{ch:3}

In most recent years, zero-shot learning (ZSL) has gained increasing attention in multimedia and machine learning areas. It aims at recognizing unseen class examples with knowledge transferred from seen classes. This is typically achieved by exploiting a semantic feature space, i.e., semantic attributes or word vectors, as a bridge to transfer knowledge between seen and unseen classes. However, due to the absence of unseen class examples during training, the conventional ZSL models easily suffers from domain shift and hubness problems. In this chapter, we propose a novel ZSL model that can handle these two issues well by adaptively adjusting and rectifying semantic feature space. Specifically, this adjustment is conducted by jointly considering both the class prototypes and the global distribution of data. To the best of our knowledge, our work is the first to consider the adaptive adjustment of semantic feature space in ZSL. Moreover, we also formulate the adjustment process to a more efficient training framework by combining it with a cycle mapping, which significantly boosts the training. By using the proposed method, we could mitigate the domain shift and hubness problems and obtain more generalized results to unseen classes. Experimental results on several widely used benchmark datasets show the remarkable performance improvement of our method compared with other representative methods.
 
\section{Introduction}
Zero-shot learning (ZSL) imitates human ability in recognizing novel unseen classes. ZSL is achieved by exploiting labeled seen class examples and a semantic feature space, e.g., attribute or word vector space, which is shared and can be transferred between seen and unseen classes \cite{lampert2014attribute,rahman2018unified}. In ZSL, the common practice is to map an unseen class example from its original feature space, e.g., visual feature space, to the semantic feature space by a mapping function learned on seen classes. Then with such obtained semantic feature vector, we search its most closely related semantic prototype whose corresponding class is set to this example. Specifically, this relatedness can be measured by a certain similarity metrics, e.g., cosine similarity, between two semantic prototypes. Thus, we can use some simple algorithms such as the K-Nearest-Neighbour (KNN) to search the class prototypes.

However, this mapping function is trained solely on seen classes that concerns the mapping from the visual to semantic feature spaces with only seen class examples. 
Although the semantic feature space is shared by both seen and unseen classes, the training and testing classes are intuitively different. 
Due to the absence of unseen classes during training, ZSL easily suffers from the domain shift problem \cite{fu2015transductive}, which means that the mapping function learned from source domain (seen classes) may not well-adopted to the target domain (unseen classes) when transferring and reusing it to infer unseen class examples.
Moreover, during the KNN search, a small number of semantic prototypes may easily become the most closely related semantic prototypes to most testing unseen class examples and become the hubs \cite{shigeto2015ridge}. The hubness commonly exists in most similarity or distance based algorithms, while the causes are still under investigation \cite{radovanovic2010hubs}. The domain shift and hubness problems hinder the performance of ZSL models and become the open challenges. Several investigations have been made to mitigate the domain shift problem including inductive learning-based methods, which enforce additional constraints from the training data \cite{fu2015zero,changpinyo2016synthesized}, and transductive learning-based methods, which assume that the unseen class examples or their visual features (unlabeled) are also available during training \cite{fu2015transductive,li2017zero,song2018transductive}. 
Despite the efforts made, most existing methods still have some drawbacks that need further investigation. For example, some key building blocks in ZSL, e.g., the semantic feature space itself and the inherent data distribution, do not seem to receive comparable attention. Conventional ZSL methods usually treat the semantic feature space as unchangeable features and keep each class prototype fixed during training. 
However, based on our observation, the mapped unseen class examples are usually quite concentrated in the semantic feature space. Worse still, some class prototypes are also too closely distributed. 
These deficiencies affect the model's ability to adapt and generalize to unseen classes. As we know, the process of human beings understanding things is constantly improving. Similarly, we argue the unchangeable semantic feature space as another inducement for the domain shift and hubness problems. 

\begin{figure}[t]
\centerline{\includegraphics[width=0.97\textwidth]{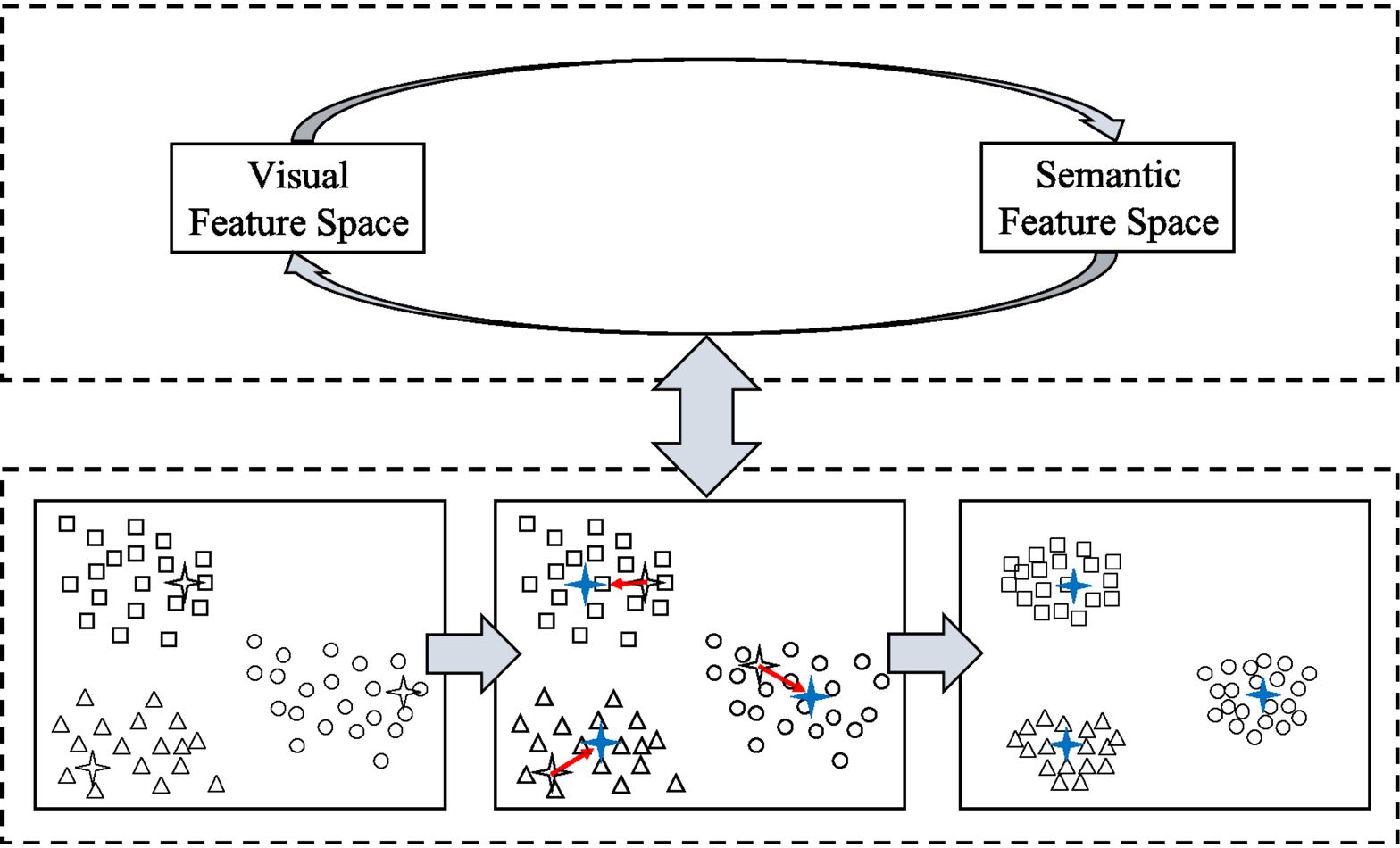}}
	\caption{The unified framework of our proposed method.}
	\label{AdaZSL}
\end{figure}

To address these problems, we propose a novel ZSL model combined with a cycle mapping to adaptively adjust the semantic feature space (Figure \ref{AdaZSL}). 
Specially, this adjustment is conducted on both the semantic prototypes and the global distribution of data, focus on decreasing the intra-class variance and enlarging the inter-class diversity. This adaptively adjustment has advantages in two aspects: first, in the statistical point of view, the semantic features become more discriminative and powerful. Second, in the geometrical point of view, the obtained semantic features could be more spatially separated and has more diversity. Thus, the semantic feature space can be better shared and transferred between seen and unseen classes. 
Moreover, we further combine the above adjustment with a cycle mapping, e.g., an encoder-decoder structure, to formulate our solution to a more efficient training framework. Our model first maps examples from visual to semantic feature spaces, and then from semantic to visual feature spaces vice versa. This cycle mapping makes the visual-semantic mapping function more faithful and robust, which can not only obtain the semantic features, but can also retain and embed more information from the visual features. Last, we construct the whole training process to a unified framework and formulate it to a generalized Lyapunov equation that significantly boosts the training efficiency. 
Experimental results on several benchmark datasets demonstrate the effectiveness of our method. 
Our contributions can be summarized as follows:
\begin{itemize}
\item The first model proposed to adaptively adjust the semantic feature space for zero-shot learning.
\item We combine the adjustment with a cycle mapping to further obtain the robust mapping function and boosts the training efficiency.
\item Our method can better handle the domain shift and hubness problems.
\end{itemize}

\section{Related Work}
\label{c3:Related_Work}

\subsection{Existing Works}
There are several recently proposed works have partially addressed the domain shift problem. TMV-HLP \cite{fu2015transductive} proposes the transductive multi-view ZSL which assumes that the unseen class examples (unlabeled) are also available during training. DeViSE \cite{frome2013devise} trains a linear mapping function between visual and semantic feature spaces by an effective ranking loss formulation. ESZSL \cite{romera2015embarrassingly} applie the square loss to learn the bilinear compatibility and adds regularization to the objective with respect to Frobenius norm. SSE \cite{zhang2015zero} proposes to use the mixture of seen class parts as the intermediate feature space. AMP \cite{bucher2016improving} embeds the visual features into the attribute space. RRZSL \cite{shigeto2015ridge} argues that using semantic space as shared latent space may reduce the variance of features and causes the domain shift, and thus proposes to map the semantic features into visual feature space for zero-shot recognition. SynC$^{struct}$ \cite{changpinyo2016synthesized} and CLN+KRR \cite{long2017zero} propose to jointly embed several kinds of textual features and visual features to ground attributes. Similarly, JLSE \cite{zhang2016zero} proposes a a joint discriminative learning framework based on dictionary learning to jointly learn model parameters for both domains. SAE \cite{kodirov2017semantic} proposes to use a linear semantic autoencoder to regularize the zero-shot learning and makes the model generalize better to unseen classes. MFMR \cite{xu2017matrix} utilizes the sophisticated technique of matrix tri-factorization with manifold regularizers to enhance the mapping function between the visual and semantic feature spaces. Differently, RELATION NET \cite{sung2018learning} learns a metric network that takes paired visual and semantic features as inputs and calculates their similarities. \emph{Chen et al.} \cite{chen2019learning} proposed learn dictionaries through joint training with examples, attributes and labels to achieve the zero-shot recognition. With the popularity of generative adversarial networks (GANs), CAPD-ZSL \cite{verma2017simple} proposes a simple generative framework for learning to predict previously unseen classes, based on estimating class-attribute-gated class-conditional distributions. GANZrl \cite{tong2018adversarial} proposes to apply GANs to synthesize examples with specified semantics to cover a higher diversity of seen classes. Instead, GAZSL \cite{zhu2018generative} applies GANs to imagine unseen classes from text descriptions. Recently, SGAL \cite{yu2019zero} uses the variational autoencoder with class-specific multi-modal prior to learn the conditional distribution of seen and unseen classes.

Despite the progress made, most of these methods ignore one of the key building blocks in ZSL, i.e., the semantic feature space, which hinders further mitigation of the domain shift problem. Worse still, since the causes for hubness problem are still under investigation in other areas, most of these ZSL methods can hardly handle the hubness problem.

\subsection{Autoencoder}
The basic autoencoder was first introduced in 1986 and recently became popular and widely used in various applications in machine learning and data mining areas. The autoencoder is an encoder-decoder structural networks which can automatically learn a latent feature representation of data. Firstly, the encoder is given an input data example $x$, then the encoder maps the input example to a latent feature space in which we obtain the latent feature representation $h$ of $x$ as:
\begin{equation}
h = W_{e}\cdot x+b_{e},
\end{equation}
where $W_{e}$ is the weight of encoder and $b_{e}$ is a bias. Then the latent feature representation $h$ is mapped back to its original feature space by the decoder, and be further reconstructed as ${x}'$:
\begin{equation}
{x}'=W_{d}\cdot l+b_{d},
\end{equation}
where $W_{d}$ is the weight of decoder and $b_{d}$ is a bias. The error between the original input data example $x$ and the reconstructed ${x}'$ can be calculated as:
\begin{equation}
\begin{aligned}
\mathcal{L}\left(x, {x}'\right) &= \left \| x-{x}' \right \|_{2}\\
&=\left \| x-\left[W_{d}\cdot \left(W_{e}\cdot x+b_{e}\right)+b_{d}\right] \right \|_{2}.
\end{aligned}
\end{equation}

The autoencoder forces the latent feature representation to retain the most powerful information of input data. The optimization target of the autoencoder is to minimize the error or loss function with respect to the input and output to achieve a better reconstruction ability and at the same time obtain a powerful latent feature representation of data. In recent years, some variants extended from the vanilla autoencoder have been proposed. These variants usually associate with some regularizers to achieve different objectives. Some representative methods include Denoising AE \cite{vincent2010stacked}, Sparse AE \cite{xu2016stacked}, Graph AE \cite{yu2013embedding}, Winner-take-all AE \cite{makhzani2014winner}, Similarity-aware AE \cite{chu2017stacked}, etc. In our method, we apply the autoencoder to form the cycle mapping and combine with the adaptive adjustment of semantic feature space to construct our unified framework.

\section{Methodology}
\label{c3:Methodology}
In this section, we elaborate on the design of our proposed method. We first introduce the cycle mapping which is mainly implemented by an autoencoder structural network. Then we introduce the adaptive adjustment of the semantic feature space regarding the class prototypes and the global data distribution. Last, we construct a unified framework of the whole training process and further formulate it to a more efficient solution.

\subsection{Cycle Mapping}
\begin{figure}[t]
\centerline{\includegraphics[width=0.99\textwidth]{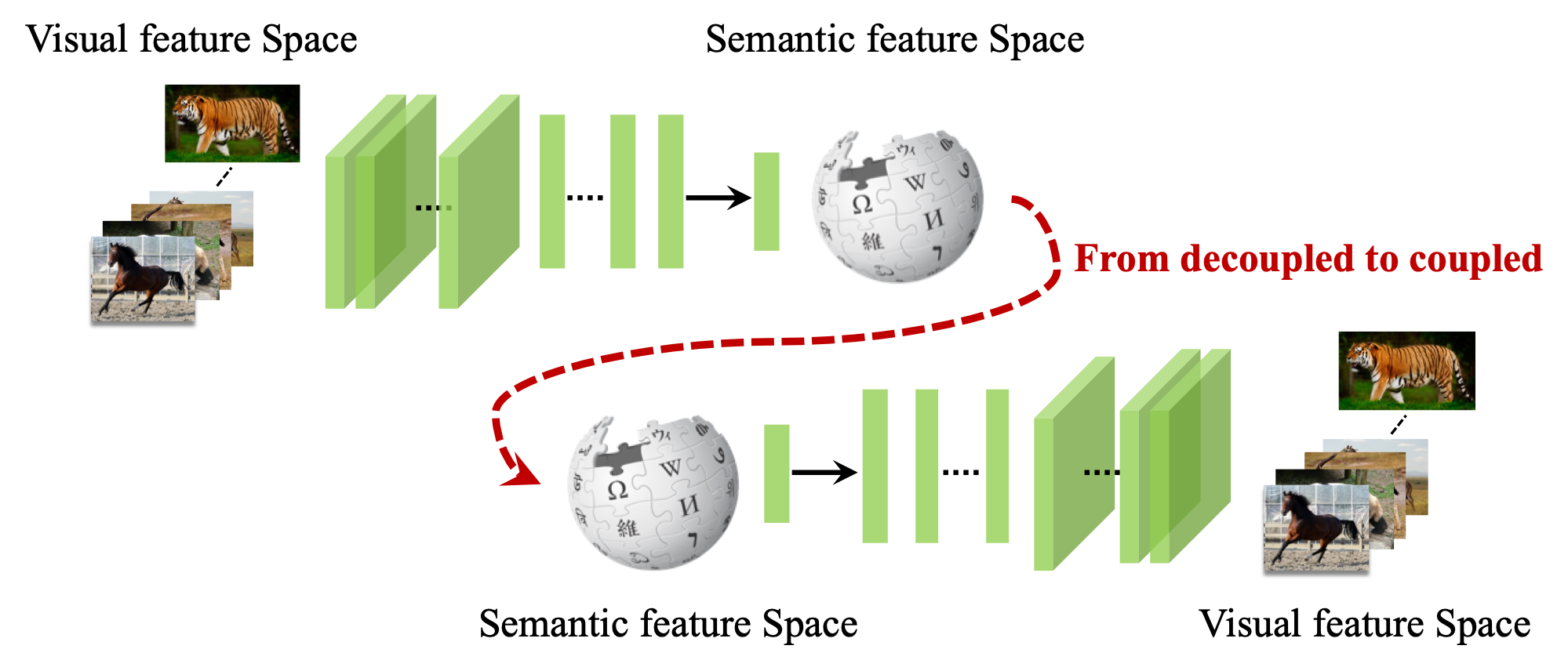}}
	\caption{The upper part denotes the visual to semantic mapping and the lower part denotes the semantic to visual mapping. In our method, we propose to couple these two mappings, namely cycle mapping, for joint training.}
	\label{decoupled_coupled}
\end{figure}

There usually consists of two phases of ZSL: mapping and searching. Taking the visual to semantic mapping as an example, the model first maps the visual features of an unseen class example to the semantic feature space. Then with the obtained semantic feature vector, the model searches the most closely related semantic prototype and sets the class corresponding with this prototype to the testing example. The recognition can be described as:
\begin{equation}
c (x_{test}) = \mathop{\arg\max}_{j} \ Sim \left ( f_{v\rightarrow s}\left ( \phi \left ( x_{test} \right ); W_{v} \right ), p^{(u_{j})} \right ), 
\end{equation}  
where $Sim(\cdot,\cdot )$ is a similarity metric that predicts the class $c (x_{test})$ of the testing example $x_{test}$. $p^{(u)}$ being the unseen class prototypes, and $f_{v\rightarrow s}(\cdot;W_{v})$ being the mapping function with the trainable weight $W_{v}$ which can map the visual features $\phi \left ( x_{test} \right )$ to the semantic feature space. $\phi \left (\cdot \right )$ is a CNNs feature extractor which is usually trained by large scale dataset. Specifically, the mapping function $f_{v\rightarrow s}(\cdot;W_{v})$, i.e., as demonstrated in the upper part of Figure \ref{decoupled_coupled}, is trained on labeled seen class examples as:
\begin{equation}
\mathop{\arg\min}_{W_{v}} \ \frac{1}{M}\cdot \sum_{i=1}^{M} \left \| f_{v\rightarrow s}\left ( \phi \left ( x^{(s)}_{i} \right ); W_{v} \right ) - p^{(s_{i})} \right \|_{2},
\end{equation}
where $M$ is the number of total seen class examples, $x^{(s)}_{i}$ is the $i$-th seen class example, and $p^{(s_{i})}$ is the corresponding semantic prototype. 
Similarly, for the semantic to visual mapping, i.e., as demonstrated in the lower part of Figure \ref{decoupled_coupled}, we can also reversely map an example from its semantic features (e.g., corresponding semantic prototype) to the visual feature space as:
\begin{equation}
\mathop{\arg\min}_{W_{s}} \ \frac{1}{M}\cdot \sum_{i=1}^{M} \left \| f_{s\rightarrow v}\left ( p^{(s_{i})}; W_{s} \right ) - \phi \left ( x^{(s)}_{i} \right ) \right \|_{2}.
\end{equation}
This reverse mapping is indeed to find a template visual feature representation for each class that minimizes the variance among examples within this class.

However, it should be noted that whether it is visual to semantic mapping, i.e., $f_{v\rightarrow s}\left (\cdot \right )$, or semantic to visual mapping, i.e., $f_{s\rightarrow v}\left (\cdot \right )$, they are all independent training processes. Taking $f_{v\rightarrow s}\left (\cdot \right )$ as an example, we denote the whole mapped space as $d$-dimensional feature space as $\mathbb{R}^{d}$. In such a space, only a compact sub-space, i.e., we denote as $S^{d} \subset \mathbb{R}^{d}$, can really represent the semantic feature space of data. In practice, it is not difficult to train such a mapping function approaching $\mathbb{R}^{d}$, and due to the supervision, i.e., semantic class prototypes, the obtained semantic feature space of data are certainly can be used for recognition, regardless of whether identical to $S^{d}$ or not. In our method, we propose to couple these two mappings, namely cycle mapping, for joint training (Figure \ref{decoupled_coupled}). Specifically, the cycle mapping can be implemented by an encoder-decoder structure as:
\begin{equation}
	f_{v \rightleftharpoons s}\left ( \phi \left ( x^{(s)} \right ); W_{v}, W_{s} \right ) = f_{s\rightarrow v}\left ( f_{v\rightarrow s}\left ( \phi \left ( x^{(s)} \right ); W_{v} \right ); W_{s} \right ).
\end{equation}
Thus, given seen class examples, the training can be described as:
\begin{equation}
\begin{aligned}
\mathop{\arg\min}_{W_{v}, W_{s}} \ &\left \| x^{(s)} - f_{v \rightleftharpoons s}\left ( \phi \left ( x^{(s)} \right ); W_{v}, W_{s} \right ) \right \|_{2},\\
\mbox{s.t.}\quad &f_{v\rightarrow s}\left ( \phi \left (x^{(s)} \right ); W_{v}  \right ) = p^{(s_{i})},
\end{aligned}
\label{c3_f1}
\end{equation}
where the cycle mapping $f_{v \rightleftharpoons s}\left ( \cdot; W_{v}, W_{s} \right )$ first maps the visual features of training examples to semantic feature space with $W_{v}$, and then reconstructs them by reversely mapping them back to visual feature space with $W_{s}$. The constraint $f_{v\rightarrow s}\left ( \phi \left (x^{(s)} \right ); W_{v}  \right ) = p_i$ is applied to learn the exact mapping function between the visual and semantic feature spaces. 
By using the cycle mapping structure, the obtained semantic feature space can not only correctly describe the semantic features of examples, but also retain as much information as it could from the visual feature space by the reconstruction process. In our method, one may also understand that the decoder part indeed provides an additional regularization in an end-to-end manner and forces the mapped results must be able to reconstruct the original inputs, which can make the mapping more accurate.

\subsection{Adaptive Adjustment}
Following the cycle mapping, we propose to adaptively adjust the semantic feature space. Specifically, during each training epoch, we jointly adjust the class prototypes and the global data distribution. 
For the adjustment of class prototypes, we mainly consider the current centroid and overall distribution of examples from each class in the semantic feature space. This prototype adjustment is partly inspired by PSO \cite{awange2018particle, eberhart1995new}, which simulates a kind of behavior performed by a group of animals such as wild gooses for their adaptation of changing the flight positions. 
For the adjustment of global data distribution, we propose a regularization term to decrease the intra-class variance of examples from each class, and enlarge the the inter-class diversity at the same time. 
These adjustments are jointly conducted during the training process and involved with the cycle mapping, to form a unified framework, which will be explained in Section \ref{Unified_Framework}.

\subsubsection{Seen Class Prototype}
\begin{figure}[t]
	\centerline{\includegraphics[width=0.99\textwidth]{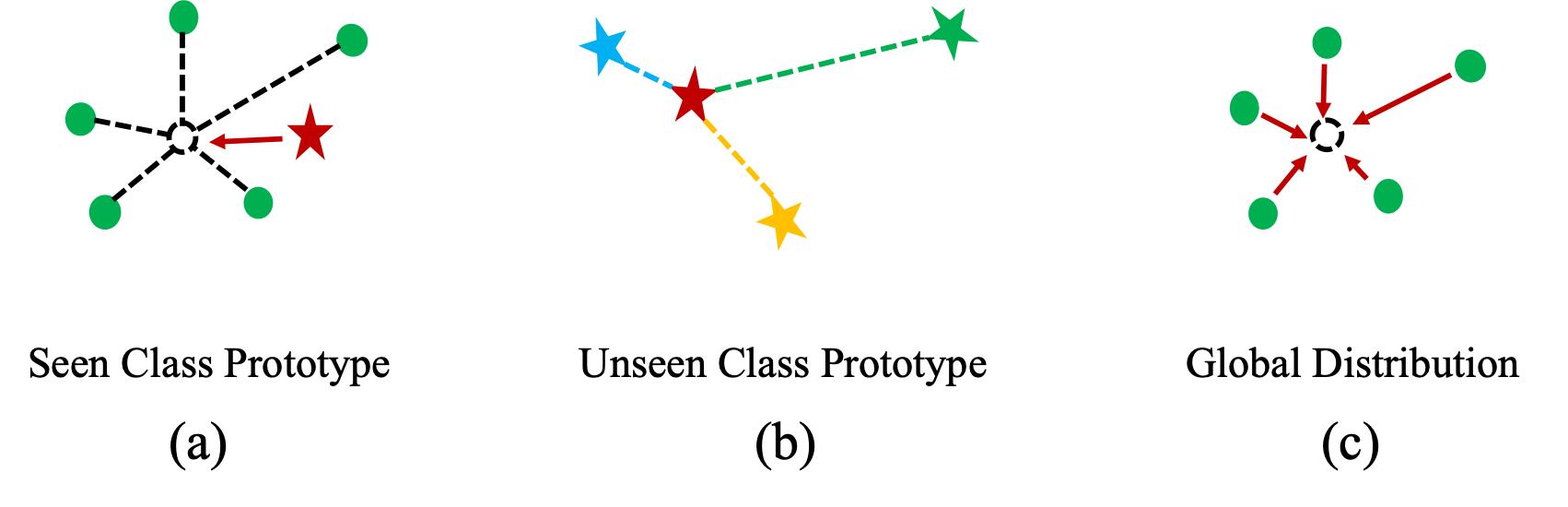}}
	\caption{Adaptive adjustment of semantic feature space.}
	\label{adjustment}
\end{figure}

To adjust seen class prototypes, we focus on the current centroid and overall distribution of examples from each class in the semantic feature space as:
\begin{equation}
\begin{aligned}
{p^{(s_{i})}}' = \lambda_{1} p^{(s_{i})} + \gamma_{1} \frac{1}{z}\sum_{j=1}^{z}f_{v\rightarrow s}\left ( \phi \left (x^{(s_{i})}_{j}\right ); W_{v} \right ),
\end{aligned}
\label{c3_f2}
\end{equation}
where ${p^{(s_{i})}}'$ and $p^{(s_{i})}$ are the updated and current prototype of the $i$-th seen class, respectively. $x^{(s_{i})}_{j}$ is the example and $z$ is the total number of examples belonging to this class. $f_{v\rightarrow s}\left ( \phi \left (x^{(s_{i})}_{j}\right ); W_{v} \right )$ is the visual to semantic mapping that calculates the semantic features of $x^{(s_{i})}_{j}$. $\lambda_{1}$ and $\gamma_{1}$ are two hyper-parameters used to control the balance of these two terms. We can consider the above adjustment in the following reason. First, for each class, the class prototype is usually not the centroid of examples in the semantic feature space, which is mainly caused by the mismatch between the diversity of various examples and the identity of unique semantic attributes or word vectors. However, this unchangeable semantic feature space hinders the improvement of models to adapt to more unseen classes. Thus, it is necessary to do adaptation, e.g., adjust the semantic prototypes. Second, since the semantic attributes or word vectors are obtained from expertise or learned knowledge which cannot be adjusted drastically. Therefore, during each training epoch, the prototype is forced to move a small step (controlled by $\gamma_{1}$) towards the centroid $\frac{1}{z}\sum_{j=1}^{z}f_{v\rightarrow s}\left ( \phi \left (x^{(s_{i})}_{j}\right ); W_{v} \right )$ of examples in the semantic feature space (Figure \ref{adjustment}.(a)).

\subsubsection{Unseen Class Prototype}
Compared to the adjustment of seen class prototypes whose visual examples are available during training, we cannot adjust that straightforward for unseen class prototypes. 
However, since the seen and unseen classes share the common semantic feature space, e.g., attribute space or word vector space, which can be transferred among classes, we thus can have another strategy to deal with the unseen class prototypes by associating with seen class prototypes. More specifically, for each unseen class prototype, we focus on its nearest seen class prototype neighbors and adjust the current unseen class prototype partially based on these neighbors (Figure \ref{adjustment}.(b)). The adjustment can be described as:
\begin{equation}
	{p^{(u_{i})}}' = \lambda_{2} p^{(u_{i})} + \gamma_{2} \sum_{j=1}^{k} \frac{\Omega (p^{(u_{i})}, p^{(s_{j})})}{{\sum }\Omega}\cdot p^{(s_{j})},
	\label{c3_f3}
\end{equation}
where ${p^{(u_{i})}}'$ and $p^{(u_{i})}$ are the updated and current prototype of the $i$-th unseen class, respectively. $p^{(s_{j})}$ ($j\in \left [ 1, k \right ]$) are the $k$ nearest seen class prototype neighbors of $p^{(u_{i})}$. The $\Omega(\cdot , \cdot )$ is a similarity metric, e.g., cosine similarity, between two vectors. $\lambda_{2}$ and $\gamma_{2}$ are two hyper-parameters used to control the balance of these two terms. In the above adjustment, we calculate the similarity between each pair of the unseen class prototype and its selected $k$ nearest seen class prototype neighbors to obtain an updating score for each neighbor as $\frac{\Omega (p^{(u_{i})}, p^{(s_{j})})}{{\sum }\Omega}$, which can represent the relative importance of each neighbor contributing to the adjustment of an unseen class prototype.

\subsubsection{Global Distribution}
The adjustment of class prototypes can rectify the mismatch between the diversity of various examples and the identity of unique semantic attributes or word vectors, which can help to mitigate the domain shift problem. However, the global domain examples are usually not evenly distributed. For example, some intra-class examples are too sparse from each other, and some inter-class examples may have some overlaps in the semantic feature space. This property exacerbates the hubness problem and can also hinder the further mitigation of the domain shift problem. To address these issues, we propose a regularization term to consider both the identity within a class and the diversity among different classes for the adjustment of global data distribution. The regularization term can be described as:
\begin{equation}
	\mathop{\arg\min}_{W_{v}} \ \sum_{i=1}^{n}\sum_{j=1}^{d}\left \| f_{v\rightarrow s}\left ( \phi \left (x^{(s_{i})}_{j}\right ); W_{v}\right ) - O_{i} \right \|_{2},
	\label{c3_f4}
\end{equation}
where $n$ is the number of seen classes and $d$ is the number of examples belonging to $i$-th seen class $s_{i}$. $O_{i}$ is the semantic centroid of the $i$-th seen class, which can be calculated by averaging the mapped semantic feature vectors of examples belonging to $s_{i}$ as $\frac{1}{d} \underset{j} {\sum } f_{v\rightarrow s}(\phi \left (x^{(s_{i})}_{j} \right ); W_{v})$. By using this regularization, we can force intra-class examples to be more concentrated, and at the same time mitigate the overlaps of inter-class examples in the semantic feature space (Figure \ref{adjustment}.(c)).
 
\subsection{Unified Framework}
\label{Unified_Framework}
In this section, we formulate our proposed method to a unified framework and introduce our algorithm in detail. 
Specifically, our method contains two components for the zero-shot recognition including the cycle mapping process which mainly act as the base trainer, and the adaptation process which is responsible for adjusting the semantic feature space. Our method can be optimized alternately with Eqs. \ref{c3_f1}$\sim$\ref{c3_f4}. We first optimize Eq. \ref{c3_f1} to obtain the initial weight of mapping function. Then Eq. \ref{c3_f2} and Eq. \ref{c3_f3} are applied to perform the adaptive adjustment for the semantic feature space by considering the class prototypes. Last, Eq. \ref{c3_f1} and Eq. \ref{c3_f4} are jointly optimized to obtain an updated weight of mapping function and adaptively adjust the global distribution at the same time. These steps are performed iteratively to reach an optimum.

In general, our objective to be minimized in the combination of Eq. \ref{c3_f1} and Eq. \ref{c3_f4} can be described as:
\begin{equation}
\begin{aligned}
& J = \sum_{i=1}^{m} \left \| x^{(s)}_{i} - f_{v \rightleftharpoons s}\left ( \phi \left ( x^{(s)}_{i} \right ); W_{v},W_{s}  \right ) \right \|_{2} + \alpha\sum_{i=1}^{n}\left \| f_{v\rightarrow s}\left ( \phi \left ( x^{(s)}_{i} \right ); W_{v}  \right ) - O_{{y^{(s)}_{i}}} \right \|_{2}, \\
& \mbox{s.t.}\quad f_{v\rightarrow s}\left ( \phi \left ( x^{(s)}_{i} \right ); W_{v}  \right ) = p^{(s_{i})},
\end{aligned}
\end{equation}
where the first term is the cycle mapping, which is implemented by an autoencoder structural network. $m$ is the number training examples and $n$ is the number of seen classes. The current semantic centroid of seen class $y^{(s)}_{i}$ is denoted as $O_{{y^{(s)}_{i}}}$. A hyper-parameter $\alpha$ is used to balance these two terms. For simplicity, we rewrite the objective to matrix form as:
\begin{equation}
\begin{aligned}
& J = \left \| \mathbf{X} - \mathbf{W_{s}}\mathbf{W_{v}}\mathbf{X}\right \|_2 + \alpha\left \| \mathbf{W_{v}}\mathbf{X} - \mathbf{O} \right \|_2,   \\
& \mbox{s.t.}\quad \mathbf{W_{v}}\mathbf{X} = \mathbf{P},
\end{aligned}
\label{thislabel1}
\end{equation}
where $\mathbf{X}$ is the matrix of visual features whose elements are obtained by $\phi \left ( x^{(s)} \right )$. $\mathbf{W_{v}}$ and $\mathbf{W_{s}}$ are the weight of the visual-semantic mapping $f_{v\rightarrow s}(\cdot)$ and semantic-visual mapping $f_{s\rightarrow v}(\cdot)$, respectively. $\mathbf{P}$ and $\mathbf{O}$ are class prototypes and semantic centroids. Both of them are duplicated and re-organized to the shape of $\mathbf{W_{v}}\mathbf{X}$, so that each semantic feature row in matrix $\mathbf{W_{v}}\mathbf{X}$ has a corresponding semantic prototype and centroid row in the matrix of $\mathbf{P}$ and $\mathbf{O}$, respectively. The objective of Eq. \ref{thislabel1} contains two parameter matrixes $\mathbf{W_{v}}$ and $\mathbf{W_{s}}$, which are usually somehow redundant and considerably increases the training cost. To further optimize the model, we apply the tied weights \cite{boureau2008sparse} to half the parameters to be optimized in our objective. The tied weights are proven to be more efficient and can obtain similar results for the encoder-decoder structural network. Hence, $\mathbf{W_{v}}$ and $\mathbf{W_{s}}$ can be simplified to tied weights as $\mathbf{W}$ and $\mathbf{W}^\top$. In this step, we can also substitute $\mathbf{W}\mathbf{X}$ with $\mathbf{P}$ for the cycle mapping term, our objective can be rewritten as:
\begin{equation}
\begin{aligned}
& J =  \left \| \mathbf{X} - \mathbf{W}^\top \mathbf{P}\right \|_2 + \alpha \left \| \mathbf{W}\mathbf{X} - \mathbf{O} \right \|_2,   \\
& \mbox{s.t.}\quad \mathbf{W}\mathbf{X} = \mathbf{P}.
\end{aligned}
\label{c3_j1}
\end{equation}
To solve this problem, we consider two operations. First, we consider to relax the hard constraint $\mathbf{W}\mathbf{X} = \mathbf{P}$ in Eq. \ref{c3_j1} to make our solution more efficient:
\begin{equation}
J =  \left \| \mathbf{X} - \mathbf{W}^\top \mathbf{P}\right \|_2 + \alpha \left \| \mathbf{W}\mathbf{X} - \mathbf{O} \right \|_2 + \beta \left \| \mathbf{W}\mathbf{X} - \mathbf{P} \right \|_2,
\end{equation}
where $\beta$ is also hyper-parameter controls the balance. By further considering the trace properties of matrix, i.e., $\mathrm{Tr}(\mathbf{X}) = \mathrm{Tr}(\mathbf{X}^\top)$ and $\mathrm{Tr}(\mathbf{W}^\top \mathbf{P}) = \mathrm{Tr}(\mathbf{P}^\top \mathbf{W})$, we can rewritten our objective as:
\begin{equation}
J =  \left \| \mathbf{X}^\top - \mathbf{P}^\top \mathbf{W}\right \|_2 + \alpha \left \| \mathbf{W}\mathbf{X} - \mathbf{O} \right \|_2 + \beta \left \| \mathbf{W}\mathbf{X} - \mathbf{P} \right \|_2,
\label{c3_j2}
\end{equation}
To solve it, we take a derivative of Eq. \ref{c3_j2} with respect to $\mathbf{W}$. Here, for the convenience of calculation, we divide the objective by 2 which does not affect the solution of the parameters:
\begin{equation}
\begin{aligned}
\frac{\partial J}{\partial \mathbf{W}} &= \frac{\partial \left ( \frac{1}{2} \left \| \mathbf{X}^\top - \mathbf{P}^\top \mathbf{W}\right \|_2 + \frac{\alpha}{2} \left \| \mathbf{W}\mathbf{X} - \mathbf{O} \right \|_2 + \frac{\beta}{2} \left \| \mathbf{W}\mathbf{X} - \mathbf{P} \right \|_2 \right )}{\partial \mathbf{W}} \\
&= (\mathbf{X}^\top - \mathbf{P}^\top \mathbf{W})\cdot \frac{\partial (\mathbf{X}^\top - \mathbf{P}^\top \mathbf{W})}{\partial \mathbf{W}} \\
&\quad+ \alpha(\mathbf{W}\mathbf{X} - \mathbf{O})\cdot \frac{\partial (\mathbf{W}\mathbf{X} - \mathbf{O})}{\partial \mathbf{W}} \\
&\quad+ \beta (\mathbf{W}\mathbf{X} - \mathbf{P})\cdot \frac{\partial (\mathbf{W}\mathbf{X} - \mathbf{P})}{\partial \mathbf{W}} \\
&= -\mathbf{P} (\mathbf{X}^\top - \mathbf{P}^\top \mathbf{W}) + \alpha(\mathbf{W}\mathbf{X} - \mathbf{O})\mathbf{X}^{\top} + \beta (\mathbf{W}\mathbf{X} - \mathbf{P})\mathbf{X}^{\top}.
\end{aligned}
\end{equation}
We set the result to zero, i.e., $\frac{\partial J}{\partial \mathbf{W}} = \mathbf{0}$ and obtain the following form:
\begin{equation}
\begin{aligned}
\mathbf{P}\mathbf{P}^\top \mathbf{W} + \left (\alpha + \beta \right )\mathbf{W}\mathbf{X}\mathbf{X}^\top - \left (\left (1+\beta \right )\mathbf{P} + \alpha \mathbf{O} \right )\mathbf{X}^\top = \mathbf{0}.
\end{aligned}
\end{equation}
We denote $\mathbf{L} = \mathbf{P}\mathbf{P}^\top$, $\mathbf{R} = \left (\alpha + \beta \right )\mathbf{X}\mathbf{X}^\top$, and $\mathbf{M} = - \left (\left (1+\beta \right )\mathbf{P} + \alpha \mathbf{O} \right ) \mathbf{X}^\top$. Then the above equation can be rewritten as:
\begin{equation}
\mathbf{L}\mathbf{W} + \mathbf{W}\mathbf{R} + \mathbf{M} = \mathbf{0},
\label{gle} 	
\end{equation}
which is the standard form of generalized Lyapunov equation and can be solved efficiently \cite{pomet1992adaptive,bartels1972solution}. The unified framework of our method is illustrated in Figure \ref{AdaZSL}. The training process is summarized in Algorithm \ref{algorithm_traning}. As to the inference process, since our method adopts the autoencoder structure, so we can either use the visual-semantic mapping or semantic-visual mapping to recognize unseen class examples. The inference processes are summarized in Algorithm \ref{algorithm_testing_vs} and Algorithm \ref{algorithm_testing_sv}, respectively.

\begin{algorithm}[h]
	\caption{Training process of our method}
	\label{algorithm_traning}
	\begin{algorithmic}[1]
		\renewcommand{\algorithmicrequire}{\textbf{Input:}}
		\renewcommand{\algorithmicensure}{\textbf{Output:}}
		\Require Seen class examples $X^{(s)} = \left \{ x^{(s)}_{1},x^{(s)}_{2},\cdots , x^{(s)}_{m} \right \}$, seen class prototypes $P^{(s)} = \left \{ p^{(s_{1})},p^{(s_{2})},\cdots, p^{(s_{n})} \right \}$, and unseen class prototypes $P^{(u)} = \left \{ p^{(u_{1})},p^{(u_{2})},\cdots, p^{(u_{d})} \right \}$. 
		\Ensure  Mapping weight $W(W_{v}, W_{s})$, updated ${P^{(s)}}' = \left \{ {p^{(s_{1})}}',{p^{(s_{2})}}',\cdots, {p^{(s_{n})}}' \right \}$, and updated ${P^{(u)}}' = \left \{ {p^{(u_{1})}}',{p^{(u_{2})}}',\cdots, {p^{(u_{d})}}' \right \}$
		\State Initialization: optimize Eq. \ref{c3_f1} to obtain initial mapping weight $W(W_{v}, W_{s})$.
		
		\State \textbf{repeat 3-5:}
		\State \quad Adjust seen class prototypes with Eq. \ref{c3_f2};
		\State \quad Adjust unseen class prototypes with Eq. \ref{c3_f3};		
		\State \quad Update the model, optimize Eq. \ref{gle} to obtain updated mapping weight $W(W_{v}, W_{s})$;
		\State \textbf{until}: find an optimum.\\
		\Return $W(W_{v}, W_{s})$,  ${P^{(s)}}'$, ${P^{(u)}}'$.
	\end{algorithmic} 
\end{algorithm}


\begin{algorithm}[h]
	\caption{Inference for ZSL on visual-semantic mapping $f_{v\rightarrow s}(\cdot)$}
	\label{algorithm_testing_vs}
	\begin{algorithmic}[1]
		\renewcommand{\algorithmicrequire}{\textbf{Input:}}
		\renewcommand{\algorithmicensure}{\textbf{Output:}}
		\Require Mapping weight $W_{v}$, unseen class examples $X^{(u)} = \left \{ x^{(u)}_{1},x^{(u)}_{2},\cdots, x^{(u)}_{k} \right \}$, unseen class prototypes ${P^{(u)}}' = \left \{ {p^{(u_{1})}}',{p^{(u_{2})}}',\cdots, {p^{(u_{d})}}' \right \}$, and unseen class labels set $C^{(u)} = \left \{c^{{(u_{1})}},c^{{(u_{2})}},\cdots, c^{{(u_{d})}} \right \}$.
		\Ensure Predicted class labels of $X^{(u)}$: $C^{p} = \left \{ c^{p}_1,c^{p}_2,\cdots , c^{p}_k \right \}$
		
		\State \textbf{Inference:}
		\State \quad Map $X^{(u)}$ to semantic feature space with $W_{v}$, obtain semantic features $S^{(u)}$;
		\State \quad Search the nearest neighbor(s) of $S^{(u)}$, targeting on ${P^{(u)}}'$;	    
		\State \quad Set class labels $C^{p}$ to $X^{(u)}$ from $C^{(u)}$ with nearest neighbor(s) results; \\
		\Return $C^{p}$
	\end{algorithmic} 
\end{algorithm}

\begin{algorithm}[t]
	\caption{Inference for ZSL on semantic-visual mapping $f_{s\rightarrow v}(\cdot)$}
	\label{algorithm_testing_sv}
	\begin{algorithmic}[1]
		\renewcommand{\algorithmicrequire}{\textbf{Input:}}
		\renewcommand{\algorithmicensure}{\textbf{Output:}}
		\Require Mapping weight $W_{s}$, unseen class examples $X^{(u)} = \left \{ x^{(u)}_{1},x^{(u)}_{2},\cdots, x^{(u)}_{k} \right \}$, unseen class prototypes ${P^{(u)}}' = \left \{ {p^{(u_{1})}}',{p^{(u_{2})}}',\cdots, {p^{(u_{d})}}' \right \}$, and unseen class labels set $C^{(u)} = \left \{c^{{(u_{1})}},c^{{(u_{2})}},\cdots, c^{{(u_{d})}} \right \}$.
		\Ensure Predicted class labels of $X^{(u)}$: $C^{p} = \left \{ c^{p}_1,c^{p}_2,\cdots , c^{p}_k \right \}$
		
		\State \textbf{Inference:}
		\State \quad Map ${P^{(u)}}'$ to visual feature space with $W_{s}$, obtain predicted template visual features for each class $X^{(t)}$;
		\State \quad Search the nearest neighbor(s) of $X^{(u)}$, targeting on $X^{(t)}$;	    
		\State \quad Set class labels $C^{p}$ to $X^{(u)}$ from $C^{(u)}$ with nearest neighbor(s) results; \\
		\Return $C^{p}$
	\end{algorithmic} 
\end{algorithm}

\clearpage
\section{Experiments}
\label{c3:Experiments}

In this section, we demonstrate the experiments of our proposed method. We first briefly introduce the evaluated datasets and metrics. Then, we introduce some experimental settings related to zero-shot leaning. Last, we introduce the results compared with some existing representative methods and some analysis of our method.

\subsection{Datasets and Metrics}

\subsubsection{Datasets}
Our model is evaluated on four widely used benchmark datasets of zero-shot learning including Animals with Attributes\footnote{http://cvml.ist.ac.at/AwA/} (AWA) \cite{lampert2014attribute}, CUB-200-2011 Birds\footnote{http://www.vision.caltech.edu/visipedia/CUB-200-2011.html} (CUB) \cite{wah2011caltech}, aPascal\&Yahoo\footnote{http://vision.cs.uiuc.edu/attributes/} (aPa\&Y) \cite{farhadi2009describing} and ILSVRC2012\footnote{http://image-net.org/challenges/LSVRC/2012/index} / ILSVRC2010\footnote{http://image-net.org/challenges/LSVRC/2010/index} (ImageNet) \cite{russakovsky2015imagenet}. 
AWA dataset consists of 30,475 images of 50 animal classes (40/10 seen and unseen classes) with six pre-extracted features for each image. The animal classes are aligned with Osherson's classical class/attribute matrix \cite{osherson1991default,kemp2006learning} providing 85-dimensional attribute features for each class. 24,295 images within 40 seen classes are used for training, and the remaining 6,180 images within 10 unseen classes are used for testing. 
CUB dataset consists of 11,788 images of 200 bird species, each of them roughly covers 30 training images and 30 testing images. Among them, 150 are seen classes and the remaining 50 are unseen classes. In this dataset, 8,855 images within 150 seen classes are used for training, and the remaining 2,933 images within 50 unseen classes are used for testing. Each image example is endowed with 312-dimensional semantic features. 
aPa\&Y dataset consists of 15,339 images and each of them is endowed with 64-dimensional attribute features. Among them, 12,695 images within 20 classes are used as the seen classes, and the remaining 2,644 images within 12 classes are used as the unseen classes.
ImageNet dataset consists of 1,000 seen classes from ILSVRC2012 and 360 unseen classes from ILSVRC2010. In the dataset, $2.0\times 10^5$ images within 1000 seen classes are used for training, and the remaining $5.4\times 10^4$ images within 360 unseen classes are used for testing. Each image example has 1,000-dimensional semantic features. 
Among them, AWA, CUB and aPa\&Y are small and medium datasets, and ImageNet is a large scale dataset.

\subsubsection{Metrics}
As the common practice in zero-shot learning, we use hit@k accuracy \cite{frome2013devise, norouzi2013zero} to evaluate the model performance. Hit@k accuracy is a widely used metric in zero-shot learning which refers to predict the top-k possible labels of the testing example, the model classifies the example correctly if and only if the ground truth is within these top-k labels. The hit@k accuracy can be described as:
\begin{equation}
	ACC_{hit@k} = \frac{\sum_{i=1}^{n} \mathbf{1}\left [ y_{i} \in \left \{ \tau^{j}\left ( x_{i} \right )  \right \}_{j=1}^{k} \right ]}{n},
\end{equation}
where $\mathbf{1}\left [ \cdot \right ]$ is an indicator function takes the value ``1'' if the argument is true, and ``0'' otherwise. $\tau(\cdot)$ is the operation which determines the class label of example. Similar with most methods, we choose hit@1 for AWA, CUB and aPa\&Y, which is the normal accuracy, and choose hit@5 for ImageNet to fit a larger scale. 
In our model, the cosine similarity is adopted for the nearest neighbor search. The hyper-parameters $\lambda_{1}$/$\gamma_{1}$ are set to 0.75/0.25, and $\lambda_{2}$/$\gamma_{2}$ are set to 0.8/0.2, respectively by grid-search \cite{hsu2003practical}.

\subsection{Experimental Setups}

\subsubsection{Feature Description}
As to the visual feature space, we choose to use the GoogleNet features \cite{szegedy2015going} consistent with most existing methods. All image examples are extracted by a trained GoogleNet. Hence, each image example is presented by a 1024-dimensional vector as the visual features. 
As to the semantic feature space, we use the semantic attributes for AWA, CUB and aPa\&Y, and use the semantic word vector for ImageNet. Because we do not directly adopt the deep convolutional neural networks in our model as a building block, so our proposed method is a non-deep model. And by using the efficient solver, our training process is more efficient despite the alternate optimization process.

\subsubsection{Non-transductive Learning}
Several zero-shot learning methods adopt the transductive setting to their models \cite{fu2015transductive, wang2017zero}, which refers to assume that the unseen examples (unlabeled) are also available during the training process. While in our proposed method, we strictly comply with the zero-shot setting that the training of our mapping function relies solely on seen class examples. 

\subsection{Results and Analysis}
This section demonstrates the experimental results in detail. Our model is compared with several competitors including DeViSE \cite{frome2013devise}, DAP \cite{lampert2014attribute}, MTMDL \cite{yang2014unified}, ESZSL \cite{romera2015embarrassingly}, SSE \cite{zhang2015zero}, RRZSL \cite{shigeto2015ridge}, \textit{Ba et al.} \cite{ba2015predicting}, AMP \cite{bucher2016improving}, JLSE \cite{zhang2016zero}, SynC$^{struct}$ \cite{changpinyo2016synthesized}, MLZSC \cite{bucher2016improving}, SS-voc \cite{fu2016semi}, SAE \cite{kodirov2017semantic}, CVAE-ZSL \cite{mishra2017generative}, CLN+KRR \cite{long2017zero}, MFMR \cite{xu2017matrix}, RELATION NET \cite{sung2018learning}, CAPD-ZSL \cite{rahman2018unified}, Chen \emph{et al.} \cite{chen2019learning}, and SGAL \cite{yu2019zero}. The selection standard for these competitors is based on following criteria: 1) all of these competitors are published in the most recent years; 2) they cover a wide range of models; 3) all of these competitors are under the same settings, i.e., datasets, evaluation criteria, etc.; and 4) they clearly represent the state-of-the-art. 

\subsubsection{Results on AWA}
The comparison results on AWA dataset is shown in Table \ref{c3_results_awa}. We compare our model with 16 representative methods on hit@1 accuracy. It can be observed that our model outperforms all competitors with great advantages in both visual-semantic and semantic-visual mappings as 88.8\% and 88.6\%, respectively. The average accuracy of our model reaches 88.7\% which produces the state-of-the-art performance.  

\begin{table}[h]
	\begin{center}
		\caption{Comparison results on AWA}
		\setlength{\tabcolsep}{5mm}{
		\label{c3_results_awa}  
			\begin{tabular}{|lc|c|}       
				\hline                   
				Method & Semantic Space &Hit@1 Accuracy(\%) \\
				\hline
				DeViSE \cite{frome2013devise} &A/W  & 56.7/50.4\\
				DAP \cite{lampert2014attribute} &A   & 60.1\\
				MTMDL \cite{yang2014unified} &A/W  & 63.7/55.3\\
				ESZSL \cite{romera2015embarrassingly} &A   & 75.3\\
				SSE \cite{zhang2015zero} &A   & 76.3\\
				SJE \cite{akata2015evaluation} &A+W  & 73.9\\
				RRZSL \cite{shigeto2015ridge} &A   & 80.4\\
				\textit{Ba et al.} \cite{ba2015predicting} &A/W  & 69.3/58.7\\
				AMP \cite{bucher2016improving} &A+W  & 66.0\\
				JLSE \cite{zhang2016zero} &A   & 80.5\\
				CVAE-ZSL\cite{mishra2017generative} &A   &71.4\\
				SynCstruct \cite{changpinyo2016synthesized} &A   & 72.9\\
				MLZSC \cite{bucher2016improving} &A   & 77.3\\
				SS-voc \cite{fu2016semi} &A/W  & 78.3/68.9\\
				SAE \cite{kodirov2017semantic} &A   & 84.4\\
				CAPD-ZSL \cite{verma2017simple} &A  &80.8 \\
				CLN+KRR \cite{long2017zero} &A  &81\\
				RELATION NET \cite{sung2018learning} &A  &84.5 \\
				Chen \emph{et al.} \cite{chen2019learning}  &A  &82.7 \\
				SGAL \cite{yu2019zero} &A &84.1 \\
				\textbf{Ours ($f_{v\rightarrow s}(\cdot)$)}&A  & \textbf{88.8}\\
				\textbf{Ours ($f_{s\rightarrow v}(\cdot)$)}&A   & \textbf{88.6}\\
				\textbf{Ours (average)}&A   &\textbf{88.7}\\
				\hline  
		\end{tabular}
		}
	\end{center}
	\footnotesize{A denotes the attribute and W denotes word vector. A/W denotes the model considers attribute and word vector as the semantic feature space, respectively. A+W denotes the model considers the combination or fusion of them as the semantic feature space.}  
\end{table}

\subsubsection{Results on CUB}
The comparison results on AWA dataset is shown in Table \ref{c3_results_cub}. We compare our model with 14 representative methods on hit@1 accuracy. 
We can observe that our model also obtains state-of-the-art performance with great advantages among these competitors in both visual-semantic and semantic-visual mappings. The average accuracy of our model reaches 64.3\%, and the results of visual-semantic mapping is 0.8\% higher than semantic-visual mapping.

\begin{table}[h]
	\begin{center}
		\caption{Comparison results on CUB}
		\setlength{\tabcolsep}{5mm}{
		\label{c3_results_cub}  
			\begin{tabular}{|lc|c|}        
				\hline                   
				Method & Semantic Space &Hit@1 Accuracy(\%) \\
				\hline
				DeViSE \cite{frome2013devise} &A/W  & 33.5\\
				MTMDL \cite{yang2014unified} &A/W  & 32.3\\
				SSE \cite{zhang2015zero} &A   & 30.4\\
				SJE \cite{akata2015evaluation} &A+W  & 50.1\\
				\textit{Ba et al.} \cite{ba2015predicting} &A/W  & 34.0\\
				ESZSL \cite{romera2015embarrassingly} &A   & 48.7\\
				RRZSL \cite{shigeto2015ridge} &A   & 52.4\\
				JLSE \cite{zhang2016zero} &A   & 41.8\\
				SynCstruct \cite{changpinyo2016synthesized} &A   & 54.4\\
				MLZSC \cite{bucher2016improving} &A   & 43.3\\
				DS-SJE \cite{reed2016learning} &A    &50.4\\
				CVAE-ZSL\cite{mishra2017generative} &A   &52.1\\
				CLN+KRR \cite{long2017zero} &A  &58.6\\
				SAE \cite{kodirov2017semantic} &A   & 61.2\\
				CAPD-ZSL \cite{verma2017simple} &A  &56.5 \\
				RELATION NET \cite{sung2018learning} &A  &62.0 \\
				Chen \emph{et al.} \cite{chen2019learning}  &A  &58.5 \\
				SGAL \cite{yu2019zero} &A &62.5 \\
				\textbf{Ours ($f_{v\rightarrow s}(\cdot)$)}&A   & \textbf{64.7}\\
				\textbf{Ours ($f_{s\rightarrow v}(\cdot)$)}&A   & \textbf{63.9}\\
				\textbf{Ours (average)}&A   &\textbf{64.3}\\
				\hline  
		\end{tabular}
		}
	\end{center}
\footnotesize{A denotes the attribute and W denotes word vector. A/W denotes the model considers attribute and word vector as the semantic feature space, respectively. A+W denotes the model considers the combination or fusion of them as the semantic feature space.}    
\end{table}

\subsubsection{Results on aPa\&Y}
The comparison results on aPa\&Y dataset is shown in Table \ref{c3_results_apay}. We compare our proposed model with 7 representative methods. It can be observed from the results that, although the advantages of our model are not significantly greater than other competitors, we can also obtain the best results among them. The average accuracy of our model reaches 56.4\%. Different from AWA and CUB, the results of semantic-visual mapping is slightly higher, i.e., 0.3\%, than visual-semantic mapping.

\begin{table}[h]
	\begin{center}
		\caption{Comparison results on aPa\&Y (A denotes attribute space)}
		\setlength{\tabcolsep}{5mm}{
		\label{c3_results_apay}   
			\begin{tabular}{|lc|c|}        
				\hline                   
				Method & Semantic Space &Hit@1 Accuracy(\%) \\
				\hline
				ESZSL \cite{romera2015embarrassingly} &A   & 24.3\\
				DAP \cite{lampert2014attribute} &A   & 38.2\\
				SSE \cite{zhang2015zero} &A   & 46.2\\
				RRZSL \cite{shigeto2015ridge} &A   & 48.8\\
				JLSE \cite{zhang2016zero} &A   & 50.4\\
				MLZSC \cite{bucher2016improving} &A   & 53.2\\
				SAE \cite{kodirov2017semantic} &A   & 55.1\\
				\textbf{Ours ($f_{v\rightarrow s}(\cdot)$)}&A   &\textbf{56.2}\\
				\textbf{Ours ($f_{s\rightarrow v}(\cdot)$)}&A   &\textbf{56.5}\\
				\textbf{Ours (average)}&A   &\textbf{56.4}\\
				\hline  
		\end{tabular}}
	\end{center}     
\end{table}

\subsubsection{Results on ImageNet}
The comparison results on ImageNet dataset is shown in Table \ref{c3_results_imagenet}. Our model is compared with 7 representative methods on hit@5 accuracy. We can observe from the results that our model can obtain competitive results against these competitors. The average accuracy of our model reaches 27.3\%. Similar with aPa\&Y, the results of semantic-visual mapping is also slightly higher, i.e., 0.3\%, than visual-semantic mapping.

\begin{table}[h]
	\begin{center}
		\caption{Comparison results on ImageNet (A denotes attribute space and W denotes word vector space)}
		\setlength{\tabcolsep}{5mm}{
		\label{c3_results_imagenet}  
			\begin{tabular}{|lc|c|}        
				\hline                   
				Method & Semantic Space &Hit@5 Accuracy(\%) \\
				\hline
				DeViSE \cite{frome2013devise} &W   & 12.8\\
				AMP \cite{bucher2016improving} &W   & 13.1\\
				ConSE \cite{norouzi2013zero}  &W   & 15.5\\
				SS-voc \cite{fu2016semi} &A   & 16.8\\
				SAE \cite{kodirov2017semantic} &W   & 26.8\\
				CVAE-ZSL\cite{mishra2017generative} &W   &24.7\\
				\textbf{Ours($f_{v\rightarrow s}(\cdot)$)}&W   &\textbf{27.1}\\
				\textbf{Ours($f_{s\rightarrow v}(\cdot)$)}&W   &\textbf{27.4}\\
				\textbf{Ours(average)}&W   &\textbf{27.3}\\
				\hline  
		\end{tabular}}
	\end{center}  
\end{table}

\subsection{Further Analysis}
\begin{figure}[htbp]
	\centerline{\includegraphics[width=0.8\textwidth]{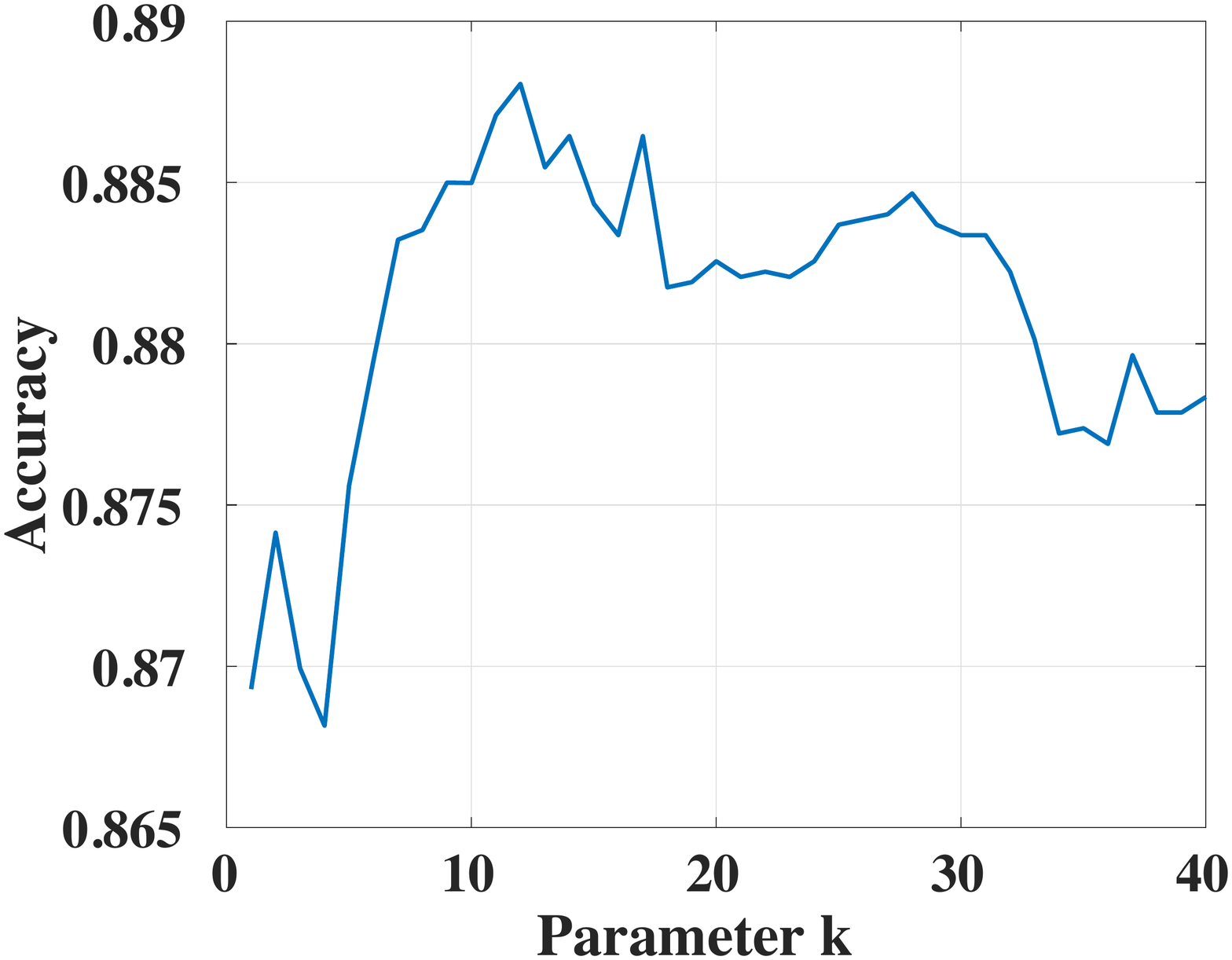}}
	\caption{K search for AWA}
	\label{k_awa}
\end{figure}

\begin{figure}[htbp]
	\centerline{\includegraphics[width=0.8\textwidth]{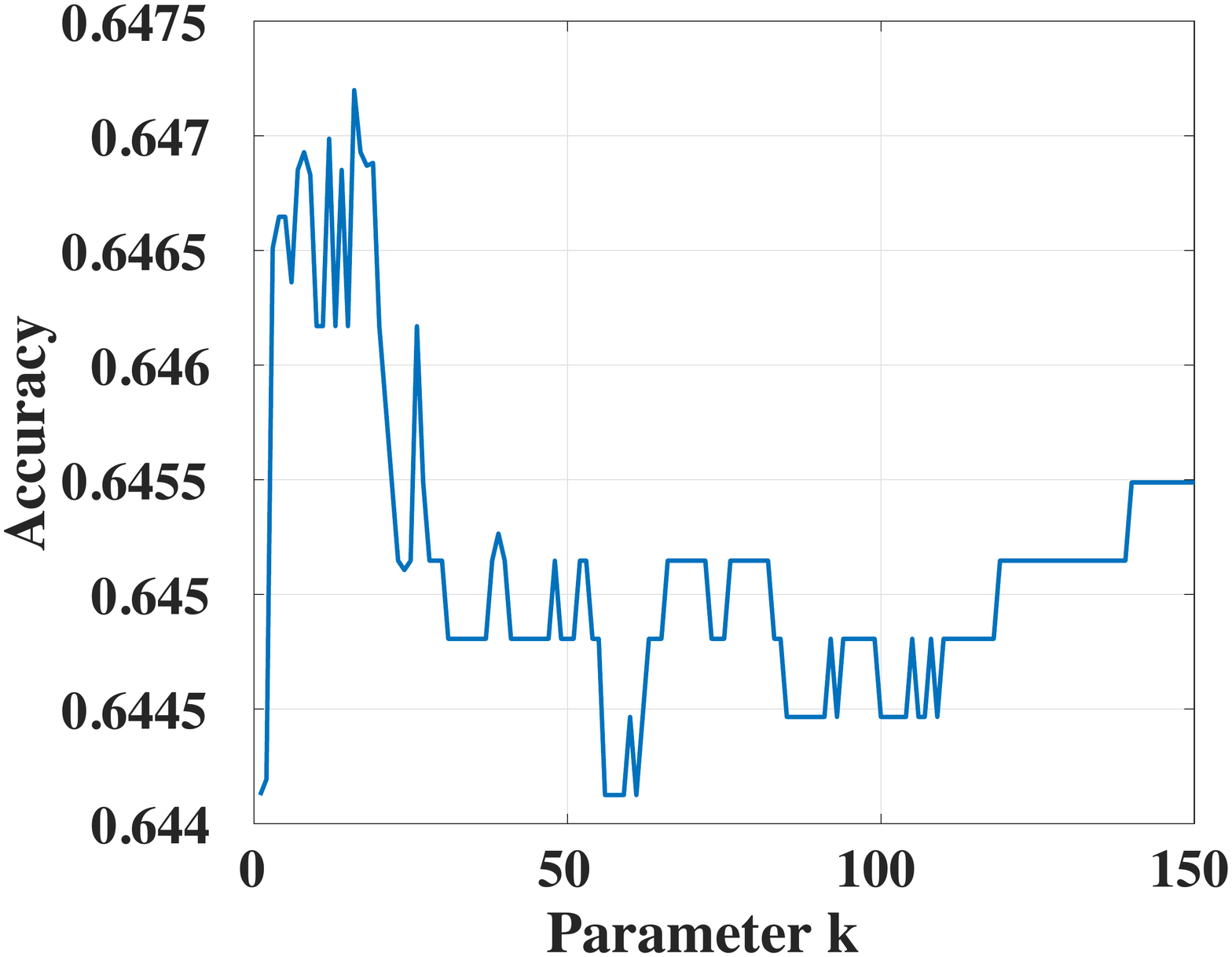}}
	\caption{K search for CUB}
	\label{k_cub}
\end{figure}

\begin{figure}[htbp]
	\centerline{\includegraphics[width=0.8\textwidth]{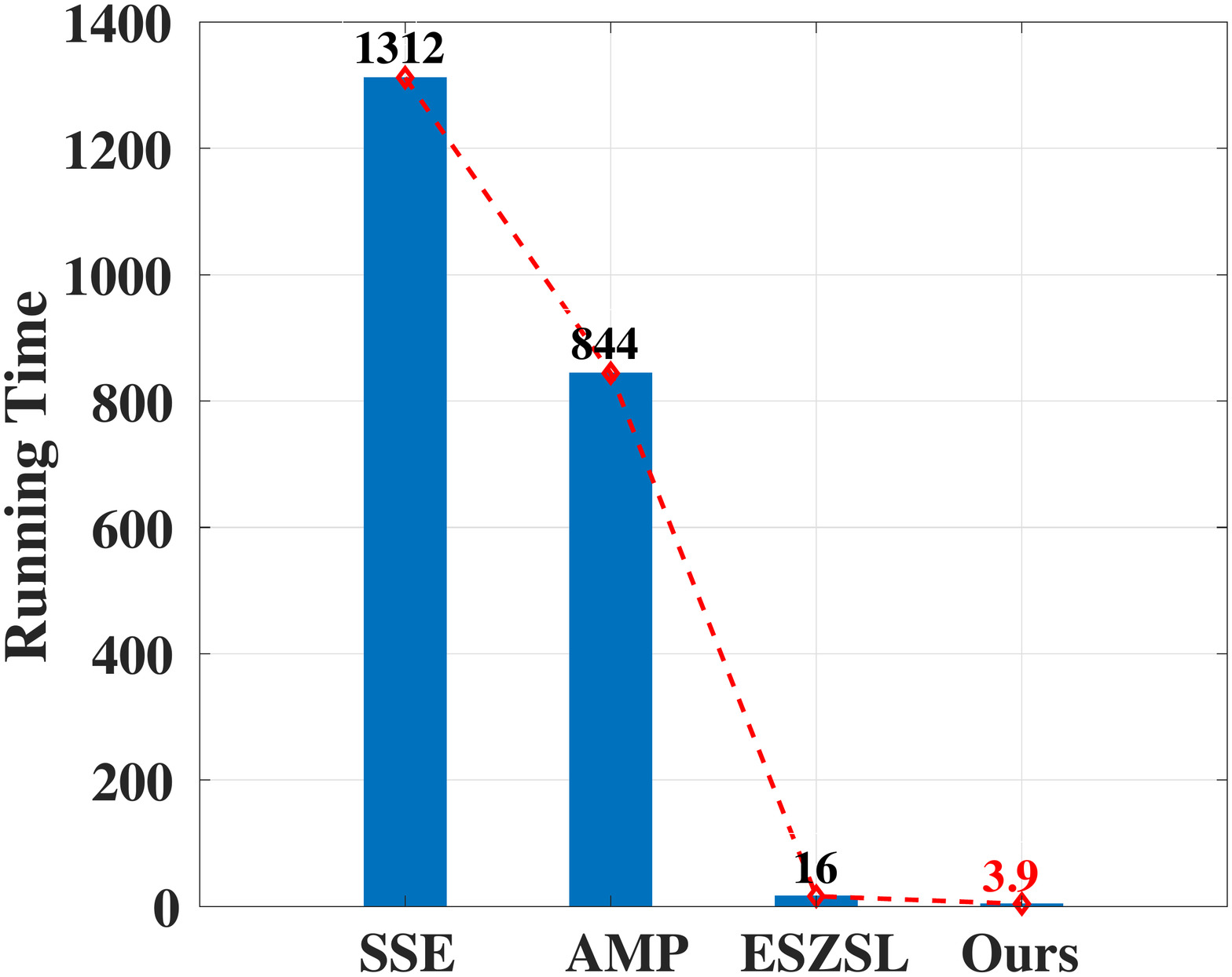}}
	\caption{Training efficiency}
	\label{time}
\end{figure}

 We introduce some further analysis regarding the experimental results. 
 First, we explore the impact of parameter $k$ on the model recognition accuracy, which refers to the k nearest seen class prototype neighbors in the adjustment for the unseen class prototypes (Eq. \ref{c3_f3}). 
 To deal with it, we choose to evaluate on AWA dataset for a smaller k-search, and on CUB for a larger k-search. The results are demonstrated in Figure \ref{k_awa} and \ref{k_cub}, respectively.
We can observe that for AWA dataset, the preferred range is $k\in [7,17]$ and the optimal $k$ is around 12. 
For CUB, the preferred range is $k\in [4,19]$ and the optimal $k$ is around 16. 
Second, we evaluate the running time of our proposed method. Because the inference process of zero-shot learning is usually not costly compared to the training process. Thus, we choose to consider the AWA training time of our proposed method against three representative methods including SSE \cite{zhang2015zero}, AMP \cite{bucher2016improving} and ESZSL \cite{romera2015embarrassingly}, and the results are shown in Figure \ref{time}. From the results, we can observe that our method has the fastest training speed among these competitors, which the speed is 336$\times$, 216$\times$ and 4.1 $\times$ faster than SSE, AMP and ESZSL, respectively.

\section{Remarks}
In this chapter, we proposed a novel model based on a unified learning framework for zero-shot learning. Our model can adaptively adjust to rectify the semantic feature space by considering both the class prototype and the global distribution of data. Moreover, we further formulate our model combined with a cycle mapping process to a much more efficient framework that significantly boosts the training process. By using the proposed method, we could mitigate the domain shift and hubness problems and make the model better adapted to recognize unseen classes. Experimental results on several widely used benchmark datasets verified the effectiveness of our model.

\chapter{Manifold Structure Alignment by Semantic Feature Expansion}\label{ch:4}

To mitigate the domain shift problem, our previous work in \Cref{ch:2} proposes to adaptively adjust to rectify the semantic feature space regarding the class prototypes and global distribution. 
The adjustment improves the zero-shot learning (ZSL) models in two aspects. First, the adjustment of class prototypes helps to rectify the mismatch between the diversity of various visual examples and the identity of unique semantic features, e.g., attributes or word vectors, which makes the visual-semantic mapping more robust and accurate. Second, the adjustment of global distribution helps to decrease the intra-class variance and to enlarge the inter-class diversity in the semantic feature space, which can further mitigate the domain shift problem. However, there may have a possible weakness of this method on directly adjusting the previous semantic features, which we could call it as a hard adjustment. 

In this chapter, we propose a novel model called AMS-SFE, to adjust the semantic feature space. It considers to align the manifold structures between the visual and semantic feature spaces, by semantic feature expansion. 
Specifically, we build upon an autoencoder-based model to expand the semantic features from the visual inputs. Additionally, the expansion is jointly guided by an embedded manifold extracted from the visual feature space of the data. 
Compared to the previous work which conducts a hard adjustment, the alignment process is more conservative for not directly adjusting the previous semantic features. Thus, we could call it as a soft adjustment. 
Extensive experiments show significant performance improvement, which verifies the effectiveness of our model.

\section{Introduction}
As a common practice in ZSL, an unseen class example is first mapped from the original input feature space, i.e., the visual feature space, to the semantic feature space by a mapping function trained on seen classes. Then, with such obtained semantic features, we search the most closely related prototype whose corresponding class is set to this example. Specifically, this relatedness can be measured by metrics such as the similarity or distance between the semantic features and prototypes. Thus, some simple algorithms, such as nearest-neighbor (NN), can be applied to search the class prototypes. However, due to the absence of unseen classes when training the mapping function, the domain shift problem \cite{fu2015transductive} easily occurs. This is mainly because the visual and semantic feature spaces are mutually independent. More specifically, visual features represented by high-dimensional vectors are usually not semantically meaningful, and the semantic features are not visually meaningful as well. Therefore, it is challenging to obtain a well-matched mapping between the visual and semantic feature spaces.

\begin{figure}[t]
  \centerline{\includegraphics[width=0.99\textwidth]{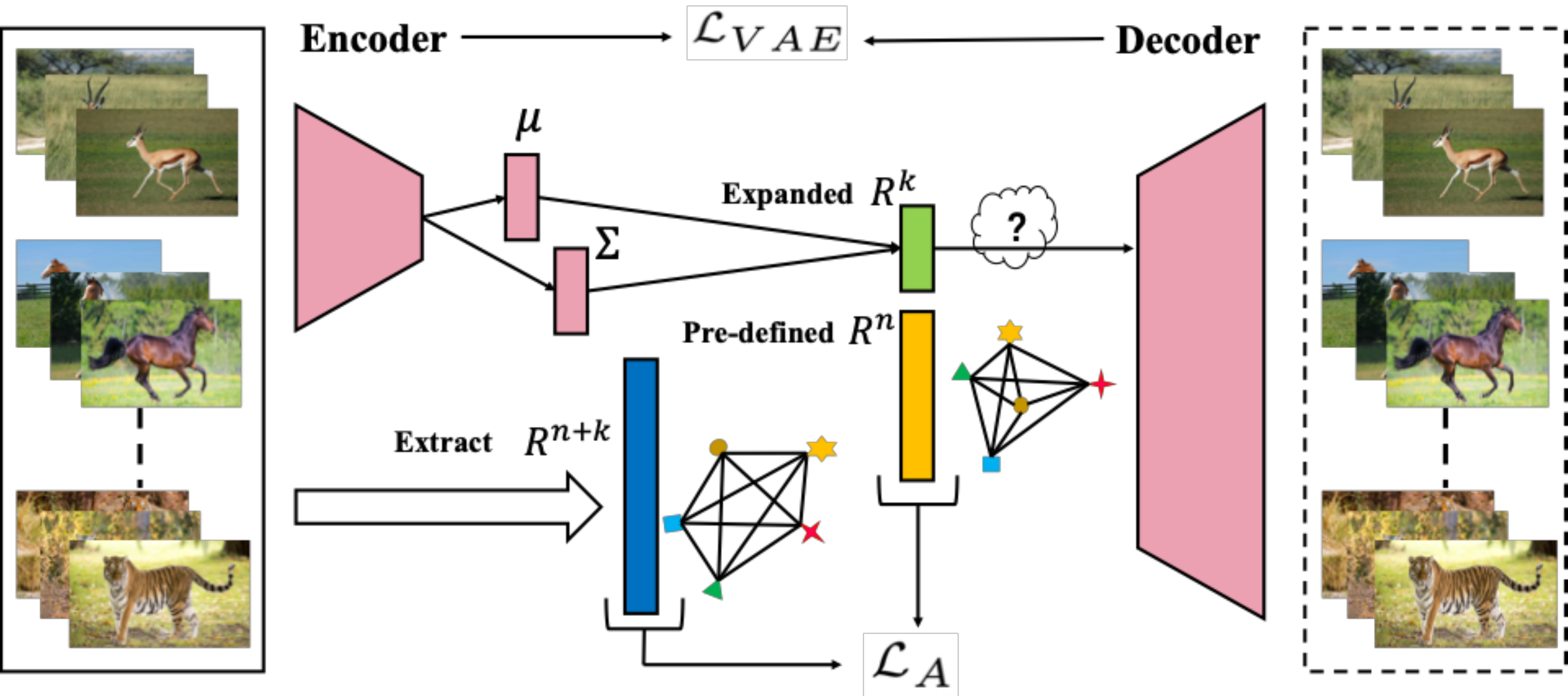}}
    \caption{The framework of AMS-SFE. The expansion contains three steps: (1) expand auxiliary semantic features (green) except for the predefined ones (yellow); (2) extract a low-dimensional embedded manifold of the visual feature space, which retains the geometrical and structural information (blue); and (3) combine two kinds of features and force their manifold structure to approximate with the structure of the extracted embedded manifold. The expansion and alignment are jointly achieved within the training of the VAE.}
    \label{ams-sfe}
\end{figure}

To address the above issues, we propose a novel model to align the manifold structures between the visual and semantic feature spaces, as shown in Figure \ref{ams-sfe}. Specifically, we train an autoencoder-based model that takes the visual features as input to generate $k$-dimensional auxiliary features for each prototype in the semantic feature space, except for the pre-defined $n$-dimensional features. 
Additionally, we combine these auxiliary semantic features with the pre-defined features to discover better adaptability for the semantic feature space. This adaptability is mainly achieved by aligning the manifold structures between the combined semantic feature space ($\mathbb{R}^{n+k}$) to an embedded $(n+k)$-dimensional manifold extracted from the original visual feature space. The expansion and alignment phases are conducted simultaneously by joint supervision from both the reconstruction and alignment terms within the autoencoder-based model. 
Our model is the first attempt to align these two feature spaces by expanding semantic features and derives benefits in two aspects. First, we can enhance the representation capability of semantic feature space, so it can better adapt to unseen classes. Second and more importantly, we can implicitly align the manifold structures between the visual and semantic feature spaces, so the domain shift problem can be better mitigated. Our contributions are three-fold: 
\begin{itemize}
\item We are the first to consider the expansion of semantic feature space for zero-shot learning and align between the visual and semantic feature spaces.
\item Our model obtains a well-matched visual-semantic projection that can mitigate the domain shift problem.
\item Our model outperforms various existing representative methods with significant improvements and shows its effectiveness.
\end{itemize}

The rest of this paper is organized as follows. Section \ref{c4:Related_Work} introduces the related work. Then, in Section \ref{c4:Methodology}, we present our proposed method. Section \ref{c4:Experiments} discusses the experiment, and the remarks is addressed in Section \ref{Remarks}.

\section{Related Work}
\label{c4:Related_Work}

\subsection{Existing Works}
The domain shift problem was first identified and studied by \textit{Fu et al.} \cite{fu2015transductive}. It refers to the phenomenon that when projecting unseen class examples from the visual feature space to the semantic feature space, the obtained results may shift away from the real results (prototypes). The domain shift problem is essentially caused by the nature of ZSL that the training (seen) and testing (unseen) classes are mutually disjoint. Recently, several researchers have investigated how to mitigate the domain shift problem, including \textit{inductive learning}-based methods, which enforce additional constraints from the training data \cite{fu2015zero,changpinyo2016synthesized}, and \textit{transductive learning}-based methods, which assume that the unseen class examples (unlabeled) are also available during training \cite{fu2015transductive,li2017zero,song2018transductive}. It should be noted that the model performance of the transductive setting is generally better than that of the inductive setting because of the utilization of extra information from unseen classes during training, thus naturally avoiding the domain shift problem. However, transductive learning does not fully comply with the \textit{zero-shot} setting in which no examples from an unseen class are available. 
Recently, some other methods have also proposed to address the domain shift problem. 
DeViSE \cite{frome2013devise} trains a linear mapping function between visual and semantic feature spaces by an effective ranking loss formulation. 
ESZSL \cite{romera2015embarrassingly} applie the square loss to learn the bilinear compatibility and adds regularization to the objective with respect to Frobenius norm. 
SSE \cite{zhang2015zero} proposes to use the mixture of seen class parts as the intermediate feature space. 
AMP \cite{bucher2016improving} embeds the visual features into the attribute space. 
RRZSL \cite{shigeto2015ridge} argues that using semantic space as shared latent space may reduce the variance of features and causes the domain shift, and thus proposes to map the semantic features into visual feature space for zero-shot recognition. 
SynC$^{struct}$ \cite{changpinyo2016synthesized} and CLN+KRR \cite{long2017zero} propose to jointly embed several kinds of textual features and visual features to ground attributes. 
Similarly, JLSE \cite{zhang2016zero} proposes a a joint discriminative learning framework based on dictionary learning to jointly learn model parameters for both domains. SAE \cite{kodirov2017semantic} proposes to use a linear semantic autoencoder to regularize the zero-shot learning and makes the model generalize better to unseen classes. MFMR \cite{xu2017matrix} utilizes the sophisticated technique of matrix tri-factorization with manifold regularizers to enhance the mapping function between the visual and semantic feature spaces. Guo \emph{et al.} \cite{guo2019ee} proposed to adaptively adjust the semantic feature space to better train the visual-semantic mapping. Differently, RELATION NET \cite{sung2018learning} learns a metric network that takes paired visual and semantic features as inputs and calculates their similarities. \emph{Chen et al.} \cite{chen2019learning} proposed learn dictionaries through joint training with examples, attributes and labels to achieve the zero-shot recognition. With the popularity of generative adversarial networks (GANs), CAPD-ZSL \cite{verma2017simple} proposes a simple generative framework for learning to predict previously unseen classes, based on estimating class-attribute-gated class-conditional distributions. GANZrl \cite{tong2018adversarial} proposes to apply GANs to synthesize examples with specified semantics to cover a higher diversity of seen classes. Instead, GAZSL \cite{zhu2018generative} applies GANs to imagine unseen classes from text descriptions. Recently, SGAL \cite{yu2019zero} uses the variational autoencoder with class-specific multi-modal prior to learn the conditional distribution of seen and unseen classes.

Although several works have already achieved some progress, the domain shift problem is still an open issue. In our model, the expansion phase is also a generative task but focuses on the semantic feature level. We adopt an autoencoder-based model that is lighter and easier to implement yet effective. Moreover, we strictly comply with the \textit{zero-shot} setting and isolate all unseen class examples from the training process.

\subsection{Manifold Learning}
The manifold is a concept from mathematics that refers to a topological space that locally resembles Euclidean space near each point. Manifold learning is based on the idea that there exists a lower-dimensional manifold embedded in a high-dimensional space \cite{wang2017quantifying}. Recently, some manifold learning-based ZSL models have been proposed. \textit{Fu et al.} introduces the semantic manifold distance to redefine the distance metric in the semantic feature space using an absorbing Markov chain process. MFMR \cite{xu2017matrix} leverages the sophisticated technique of matrix trifactorization with manifold regularizers to enhance the projection between the visual and semantic spaces. In our model, we consider to obtain an embedded manifold in a lower-dimensional space of data in the original visual feature space. This embedded manifold is expected to retain the geometrical and distribution constraints of the visual feature space. Such manifold information is further used to guide the alignment of manifold structures between the visual and semantic feature spaces.

\section{Methodology}
\label{c4:Methodology}
In this section, we first introduce our proposed method and formulation in detail. More specifically, an autoencoder-based network is first applied to generate and expand some auxiliary semantic features except for the predefined ones. 
Then, we extract a lower-dimensional embedded manifold from the original visual feature space, which properly retains its geometrical and distribution constraints. By using the obtained embedded manifold, we then construct an additional regularization term to guide the alignment of manifold structures between the visual and semantic feature spaces. 
Finally, the prototype updating strategy and the recognition process of our model are addressed.

\begin{figure}[t]
  \centerline{\includegraphics[width=0.97\textwidth]{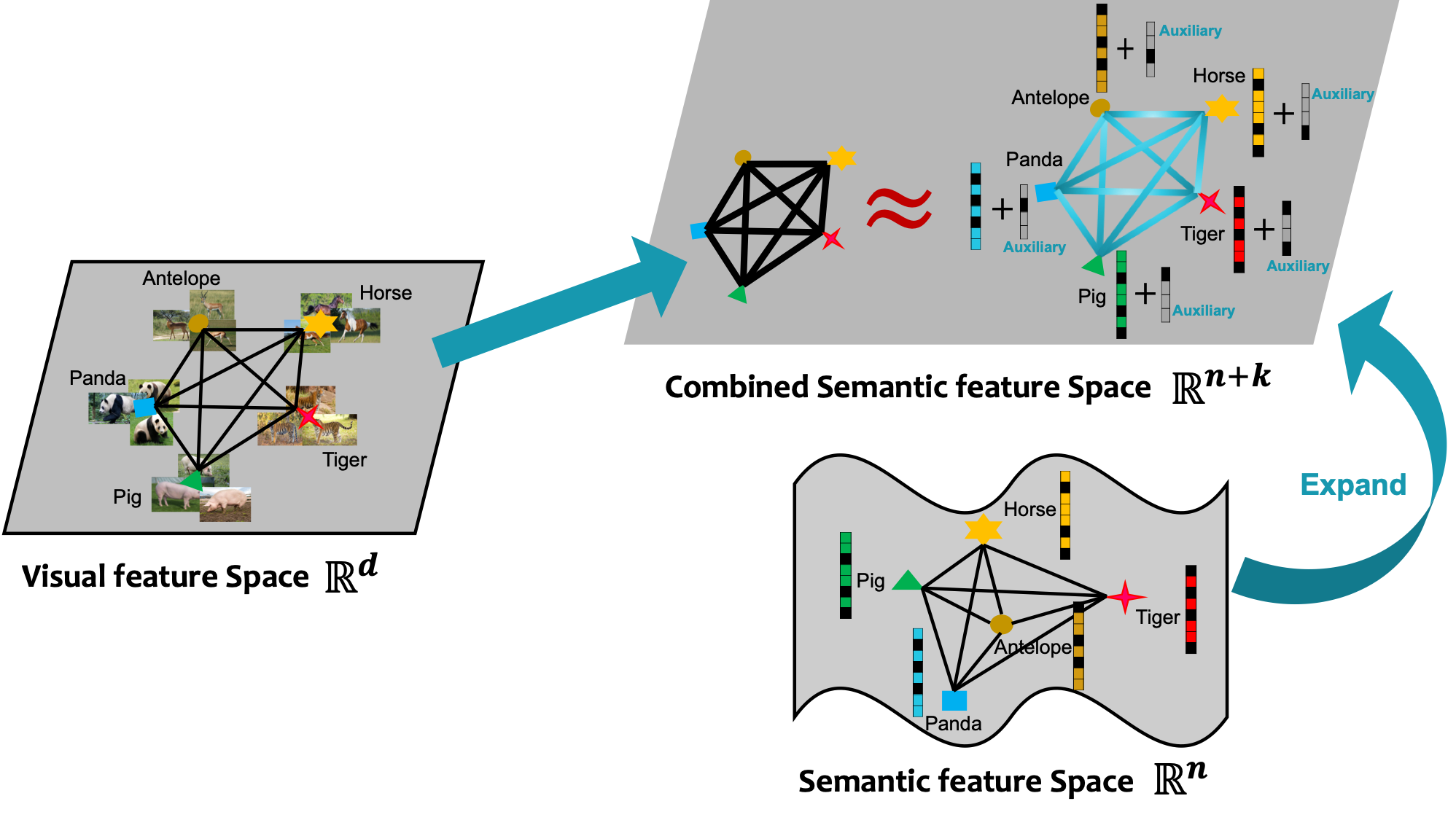}}
    \caption{We propose to align the manifold structures between the visual and semantic feature spaces by expanding the semantic features.}
    \label{ams_sfe_expand}
\end{figure}

\subsection{Semantic Feature Expansion}
To align the manifold structures between the visual and semantic feature spaces, the first step from the bottom up is to expand the semantic features. Specifically, we keep the pre-defined $n$-dimensional semantic features, i.e., $S^{p}=(a_{1}, a_{2}, \cdots , a_{n})\in \mathbb{R}^{n}$, fixed and expand extra $k$-dimensional auxiliary semantic features, i.e., $S^e = \left (a_{n+1}, a_{n+2}, \cdots , a_{n+k}\right )\in \mathbb{R}^{k}$, to enlarge the target semantic feature space as $\mathbb{R}^{n+k}$. Practically, there are several techniques such as the generative adversarial network (GAN) \cite{goodfellow2014generative}, and the autoencoder (AE) \cite{baldi2012autoencoders} can achieve this target on the basis that the expanded auxiliary semantic features are faithful to the original input features, i.e., visual features. Compared with the GAN, which is good at synthesizing data distribution, e.g., more realistic images in example-level \cite{zhang2018stackgan++}, the AE is much lighter and easier to train \cite{vincent2010stacked,yu2013embedding,chu2017stacked,guo2019ee}. Moreover, in our model, the expanded auxiliary features are expected to be more per-class semantically high-level in contrast to example-level features. Considering the generation target, the training cost and the model complexity, the AE is a better choice for our purpose. The standard AE consists of two components. The encoder maps the visual features $x \in \mathbb{R}^{d}$ to a latent feature space, in which the latent features $z \in \mathbb{R}^{k} (k\ll d)$ are normally a high-level and compact representation of the visual features. The decoder then maps the latent features back to the visual feature space and reconstructs the original visual features as $\hat{x} \in \mathbb{R}^{d}$. The AE loss measuring the reconstruction can be described as:
\begin{equation}
\mathcal{L}_{AE} = \sum_{x\sim \mathcal{D}}\left \| x - \hat{x} \right \|_{2}\,.
\label{l_ae}
\end{equation}
We minimize the objective of Eq. \ref{l_ae} to guarantee the learned latent features $z_{i}$ retain the most powerful information of the input $x_{i}$. It should be noted that the SAE \cite{kodirov2017semantic} first implemented the AE for the zero-shot learning task, which applies the semantic autoencoder to obtain the mapping function between the visual and semantic feature spaces. However, in our model, the AE is mainly implemented in the expansion process to obtain extra auxiliary semantic features.

The standard AE performs well in our model despite the latent feature space being point-wise sensitive, which means that the margins of each class within the latent feature space are discrete and the data in the latent feature space are unevenly distributed. In other words, some areas in the latent feature space do not represent any data. Although the prototypes of each class are also inherently discrete, a smooth and continuous latent feature space can intuitively better represent the margins among classes and makes the projection between the visual and latent feature spaces more robust. To this end, we further apply the variational autoencoder (VAE) \cite{DBLP:journals/corr/KingmaW13} to formulate the expansion process in our model. 

Different from the standard AE, the encoder of the VAE predicts the mean feature vector $\mu$ and the variance matrix $\Sigma$, such that the distribution of latent features $q \left ( z |x \right )$ can be approximated by $\mathcal{N} \left ( \mu, \Sigma \right )$, i.e., $q \left ( z |x \right ) = \mathcal{N} \left ( \mu, \Sigma \right )$, from which a latent feature $z$ is sampled and further decoded to reconstruct the original visual features as $\hat{x} \in \mathbb{R}^{d}$. The key difference between the AE and VAE is the embedding methods of the inputs in the latent feature space. The AE learns a compressed data representation that is normally more discrete, while the VAE attempts to learn the parameters of a probability distribution representing the data, which makes the learned latent features smoother and more continuous \cite{DBLP:journals/corr/KingmaW13,blei2017variational}. 
It should be noted that the sampling operation from $\mathcal{N} \left ( \mu, \Sigma \right )$ is usually non-differentiable which makes the backpropagation impossible. As suggested by the reparameterization trick \cite{DBLP:journals/corr/KingmaW13}, sampling from $z \sim \mathcal{N} \left ( \mu, \Sigma \right )$ is equivalent to sampling $\epsilon \sim \mathcal{N} \left ( 0, I \right )$ and setting $z = \mu + \Sigma ^{\frac{1}{2}} \epsilon$. Thus, $\epsilon$ can be regarded as an input of the network and makes the sampling operation differentiable. Moreover, we need an additional constraint other than the reconstruction loss, i.e., $\mathcal{L}_{AE}$, to guide the training of VAE. It should be noted that the additional constraint is expected to force the latent feature distribution to be similar to a prior, so the objective of the VAE can be further specified as:
\begin{equation}
\mathcal{L}_{VAE} = \sum_{x\sim \mathcal{D}}\left \| x - \hat{x} \right \|_{2}-D_{KL}  \big (q\left ( z | x \right )  \Vert p\left ( z \right ) \big ),
\label{eq2}
\end{equation}
where the first term is the conventional reconstruction loss, which forces the latent feature space to be faithful and restorable to the original visual feature space, and the second term is the unpacked Kullback-Leibler divergence between the latent feature and the chosen prior $p\left ( z \right )$, e.g., a multivariate standard Gaussian distribution, which further forces the margins of each class to be smooth and continuous and makes the visual-semantic mapping more robust.

\subsection{Manifold Extraction and Alignment}

\subsubsection{Extraction}
Before exploiting these auxiliary semantic features, we need to extract a lower-dimensional embedded manifold ($\mathbb{R}^{n+k}$) of the visual feature space ($\mathbb{R}^{d},n+k\ll d$) to utilize the structure information. To this end, we first find and define the center of each seen class in the visual feature space as $x^{c} = \left \{x^{c_{i}} \right \}_{i=1}^{m}$, where $\left \{c_{i} \right \}_{i=1}^{m}$ are $m$ class labels and $x^{c_{i}}$ is the center, e.g., the mean value, of all examples belonging to class $c_{i}$. We compose a matrix $\mathbf{D} \in \mathbb{R}^{m\times m}$ to record the distance of each center pair from $x^{c}$ in the original visual feature space as:
\begin{equation}
\mathbf{D} = \begin{bmatrix}
d_{1,1} & d_{1,2} & \cdots  & d_{1,m}\\ 
d_{2,1} & d_{2,2} & \cdots  & d_{2,m} \\  
\vdots  & \vdots  &   & \vdots \\ 
d_{m,1} & d_{m,2} & \cdots  & d_{m,m} 
\end{bmatrix},
\end{equation}
where $d_{i,j}$ is calculated as $\left \| x^{c_{i}} - x^{c_{j}} \right \|$. Then, our target is to search for a lower-dimensional embedded manifold ($\mathbb{R}^{n+k}$) that can be modeled by a $(n+k)$-dimensional embedded feature representation, i.e., denoted as $\mathbf{O}=\left [ o_{i} \right ]\in \mathbb{R}^{(n+k) \times m}$, where each $o_{i}$ is the embedded representation of the class center $x^{c_{i}}$. The embedded representation is expected to retain the geometrical and structural information of the visual feature space of the data. To obtain $\mathbf{O}$, a natural and straightforward approach is that the distance matrix $\mathbf{D}$ can also restrain the embedded representation $\mathbf{O}$, which means that the distance of each point pair of $\mathbf{O}$ also has the same distance matrix $\mathbf{D}$ in the corresponding $(n+k)$-dimensional feature space. To solve this problem, we denote the inner product of $\mathbf{O}$ as $\mathbf{B} = \mathbf{O}^\top \mathbf{O}\in \mathbb{R}^{m \times m}$, so that $b_{ij} = o_i^\top o_j$ and we obtain:
\begin{equation}
\begin{aligned}
d_{ij}^2 &= \left \| o_i \right \|^2 + \left \| o_j \right \|^2-2o_i^\top o_j = b_{ii} + b_{jj} - 2b_{ij}.
\end{aligned}
\label{eq4}
\end{equation}
We set $\sum_{i=1}^{m}o_{i}=0$ to simplify the problem so that the summation of each row/column of $\mathbf{O}$ equals zero. The zero-centered setting can reduce computations while retaining the data's geometrical and structural information. Then, we can easily obtain:
\begin{equation}
\sum_{i=1}^{m}d_{ij}^{2} = \mathrm{Tr}(\mathbf{B}) + mb_{jj},
\label{eq5} 
\end{equation}
\begin{equation}
\sum_{j=1}^{m}d_{ij}^{2} = \mathrm{Tr}(\mathbf{B}) + mb_{ii},
\label{eq6} 
\end{equation}
\begin{equation}
\sum_{i=1}^{m}\sum_{j=1}^{m}d_{ij}^{2} = 2m\mathrm{Tr}(\mathbf{B}),
\label{eq7} 
\end{equation}
where $\mathrm{Tr}(\cdot)$ is the trace of the matrix, i.e., $\mathrm{Tr}(B) = \sum_{i=1}^{m}\left \| o_i \right \|^2$. We denote:
\begin{equation}
d_{i\cdot }^{2} = \frac{1}{m}\sum_{j=1}^{m}d_{ij}^2,
\label{eq8}
\end{equation}
\begin{equation}
d_{\cdot j}^{2} = \frac{1}{m}\sum_{i=1}^{m}d_{ij}^2,
\label{eq9}
\end{equation}
\begin{equation}
d_{\cdot \cdot}^{2} = \frac{1}{m^2}\sum_{i=1}^{m}\sum_{j=1}^{m}d_{ij}^2.
\label{eq10}
\end{equation}
From Eq. \ref{eq4}, we can easily obtain:
\begin{equation}
b_{ij} = -\frac{1}{2}\left ( d_{ij}^{2} - b_{ii} - b_{jj} \right ).
\label{eq11}
\end{equation}
From Eq. \ref{eq7} and Eq. \ref{eq10}, we can obtain:
\begin{equation}
\mathrm{Tr}(\mathbf{B}) = \frac{1}{2m}\sum_{i=1}^{m}\sum_{j=1}^{m}d_{ij}^2 = \frac{1}{2}md_{\cdot \cdot}^{2},
\label{eq12}
\end{equation}
From Eq. \ref{eq6} and Eq. \ref{eq8}, and from Eq. \ref{eq5} and Eq. \ref{eq9}, respectively, we can obtain:
\begin{equation}
\left\{
\begin{array}{lr}
b_{ii} = \frac{1}{m}\sum_{j=1}^{m}d_{ij}^2 - \frac{1}{m}\mathrm{Tr}(\mathbf{B})=d_{i \cdot}^{2} - \frac{1}{2}d_{\cdot \cdot}^{2}  &  \\
b_{jj} = \frac{1}{m}\sum_{i=1}^{m}d_{ij}^2 - \frac{1}{m}\mathrm{Tr}(\mathbf{B})=d_{\cdot j}^{2} - \frac{1}{2}d_{\cdot \cdot}^{2} &  
\end{array}.
\right.
\label{eq13}
\end{equation}
Combine Eqs. \ref{eq11}$\sim$\ref{eq13}, we can obtain the inner product matrix $\mathbf{B}$ by the distance matrix $\mathbf{D}$ as:
\begin{equation}
b_{ij} = -\frac{1}{2}\left ( d_{ij}^2 - d_{i\cdot }^2 - d_{\cdot j}^2 + d_{\cdot \cdot }^2 \right ).
\end{equation}
By applying eigenvalue decomposition with $\mathbf{B}$, we can easily obtain the $(n+k)$-dimensional representation $\mathbf{O}$, which models the geometrical and structural information of the expected $(n+k)$-dimensional embedded manifold.

\subsubsection{Structure Alignment}
With the obtained $\mathbf{O}$, we now can align the manifold structures between the visual and semantic feature spaces. Specifically, we measure the similarity of the combined semantic feature representation $S^{p+e}$ (predefined $S^p$ combined with expanded $S^e$) and the embedded representation $\mathbf{O}$ by cosine distance, in which the output similarity between two vectors is bounded in $\left [ -1, 1 \right ]$ and is magnitude free as well. It should be noted that in our model, the alignment is jointly completed with the semantic feature expansion. To achieve this, we construct a regularization term to further guide the autoencoder-based network as:
\begin{equation}
\mathcal{L}_A = \sum_{i=1}^{l} \sum_{j=1}^{m} {\bf 1}\left [ y_{i} = c_{j} \right ] \cdot \left [ 1 - \frac{S_{i}^{p+e}\cdot o_{j}}{\left \| S_{i}^{p+e} \right \|\left \| o_{j} \right \|}  \right ],
\label{eq15}
\end{equation}
where $S_{i}^{p+e}$ is the combined semantic feature representation of the $i$-th seen class example $x_{i}$. $y_{i}$ is the class label and $c_{j}$ is the $j$-th class label among $m$ classes. ${\bf 1}\left [ y_{i} = c_{j} \right ]$ is an indicator function that takes a value of ``1'' if its argument is true, and ``0'' otherwise. Last, combined with Eq. \ref{eq2}, the unified objective can be described as:
\begin{equation}
\small
\begin{aligned}
\mathcal{L} = &\alpha \cdot \underset{\mathcal{L}_{VAE}}{\underbrace{\sum_{x\sim \mathcal{D}}\left \| x - \hat{x} \right \|_{2}-D_{KL}  \big (q\left ( z | x \right )  \Vert p\left ( z \right ) \big )
}} \\
&+ \beta \cdot \underset{\mathcal{L}_A}{\underbrace{\sum_{i=1}^{l} \sum_{j=1}^{m} {\bf 1}\left [ y_{i} = c_{j} \right ] \cdot  \left [ 1 - \frac{S_{i}^{p+e}\cdot o_{j}}{\left \| S_{i}^{p+e} \right \|\left \| o_{j} \right \|}  \right ]}}\,,
\end{aligned}
\label{eq16}
\end{equation}
where $\mathcal{L}_{VAE}$ acts as a base term that mainly guides the reconstruction of the input visual examples. $\mathcal{L}_{A}$ is an alignment term that provides additional guidance to learning latent vectors and forces the manifold structure of the combined semantic feature space to approximate the structure of the embedded manifold extracted from the visual feature space. $\alpha$ and $\beta$ are two hyper-parameters that control the balance between these two terms.

\subsection{Prototype Update}
After the expansion phase, we need to update the prototypes for each class. We have different strategies regarding the seen and unseen classes. First, for each seen class. Because we obtained the trained autoencoder and all the seen class examples are available, we can simply compute the center, i.e., the mean value, of all latent vectors $z_{i}$ belonging to the same class and combine the center with the predefined prototype for each seen class as:
\begin{equation}
P^{e} = \frac{1}{h}\sum_{i=1}^{h}z_{i}\,,
\end{equation}
\begin{equation}
P = P^{p} \uplus P^{e}\,,
\end{equation}
where $z_{i}$ is the expanded semantic features obtained by the encoder, $h$ is the number of examples belonging to this specific seen class, $P^{p}$ and $P^{e}$ are predefined and expanded prototypes, respectively, and $\uplus$ denotes the operation that concatenates two vectors. 
Second, for each unseen class, as no example is available during the whole training phase, we cannot apply the trained autoencoder to update the prototypes directly. Instead, we use another strategy by considering the local linearity among prototypes. Specifically, for each predefined unseen class prototype, we first obtain its $g$ nearest neighbors from predefined seen class prototypes. Then, we estimate each predefined unseen class prototype by a linear combination of its corresponding $g$ neighbors as:
\begin{equation}
\begin{aligned}
{P^{p}}' &= \theta _{1}P_{1}^{p} + \theta _{2}P_{2}^{p} + \cdots +\theta _{g}P_{g}^{p} \\
& = \theta P_{1\rightarrow g}^{p}\,,  
\end{aligned}
\label{eq19}
\end{equation}
where ${P^{p}}'$ is the predefined prototype of the unseen class, $\left \{ P_{i}^{p} \right \}_{i=1}^{g}$ are its $g$ nearest neighbors from predefined seen class prototypes, and $\left \{ \theta_{i} \right \}_{i=1}^{g}$ are the estimation parameters. Eq. \ref{eq19} is a simple linear programming, and we can easily obtain the estimation parameters by solving a minimization problem as:
\begin{equation}
\theta = \underset{\theta}{\arg \min} \left \| {P^{p}}'- \theta P_{1\rightarrow g}^{p} \right \|.
\end{equation}
With the obtained estimation parameter $\theta$, we can update the prototype for each unseen class as:
\begin{equation}
\begin{aligned}
{P^{e}}' &= \theta _{1}P_{1}^{e} + \theta _{2}P_{2}^{e} + \cdots +\theta _{g}P_{g}^{e} \\
&= \theta P_{1\rightarrow g}^{e}\,,
\end{aligned} 
\end{equation}
\begin{equation}
{P}' = {P^{p}}' \uplus  {P^{e}}'\,,
\end{equation}
where ${P}'$ is the updated prototype for the unseen class and $\left \{ P_{i}^{e} \right \}_{i=1}^{g}$ are the corresponding $g$ expanded prototypes of its $g$ seen class neighbors.

\subsection{Recognition}
\label{Recognition}

In our model, similar to some methods, we also adopt the simple autoencoder training framework \cite{kodirov2017semantic} to learn the mapping function between the visual and semantic feature spaces. The encoder $f_{e}(\cdot)$ first maps an example from the visual feature space to the semantic feature space to reach its prototype. Then, the decoder $f_{d}(\cdot)$ reversely maps it back to the visual feature space and reconstructs the example. The latent vectors of the autoencoder are forced to be the prototypes of each class. These two steps guarantee the robustness of our learned projection. Our model mainly focuses on the alignment of manifold structures by semantic feature expansion, so we do not apply any additional techniques to the mapping function training phase. A simple linear autoencoder with just one hidden layer is trained to obtain the visual-semantic mapping. Specifically, the encoder $f_{e}(\cdot)$ can be regarded as the forward mapping, and the decoder can be regarded as the reverse mapping. In the testing phase, taking the forward mapping as an example, we can map an unseen class example ${x_{i}}'$ to the semantic feature space and obtains its semantic feature representation as $f_{e}({x_{i}}')$. As to the recognition, we simply search its most closely related prototype and set the class corresponding with it to the testing example as:
\begin{equation}
c  ({x_{i}}') = \mathop{\arg\max}_{j} \ Sim(f_{e}({x_{i}}'), {p_{j}}')
\end{equation}
where ${p_{j}}'$ is the prototype of the $j$-th unseen class, $Sim(\cdot, \cdot)$ is a similarity metrics, e.g., cosine similarity, and $c(\cdot)$ returns the class label of ${x_{i}}'$.

\section{Experiments}
\label{c4:Experiments}

\subsection{Datasets and Metrics}
\label{Datasets_and_Metrics}
In our experiments, five widely used benchmark datasets are selected for the evaluation, including Animals with Attributes\footnote{http://cvml.ist.ac.at/AwA/} (AWA) \cite{lampert2014attribute}, CUB-200-2011 Birds\footnote{http://www.vision.caltech.edu/visipedia/CUB-200-2011.html} (CUB) \cite{wah2011caltech}, aPascal\&Yahoo\footnote{http://vision.cs.uiuc.edu/attributes/} (aPa\&Y) \cite{farhadi2009describing}, SUN Attribute\footnote{http://cs.brown.edu/~gmpatter/sunattributes.html} (SUN) \cite{patterson2014sun}, and ILSVRC2012\footnote{http://image-net.org/challenges/LSVRC/2012/index} / ILSVRC2010\footnote{http://image-net.org/challenges/LSVRC/2010/index} (ImageNet) \cite{russakovsky2015imagenet}. The first four are small and medium scale datasets, and ImageNet is a large-scale dataset. 
The AWA \cite{lampert2014attribute} consists of 30,475 images of 50 animal classes. Among them, 40 classes covering 24,295 images are used as seen classes for training, and 10 classes covering 6,180 images are used as unseen classes for testing. Each class is represented by an 85-dimensional numeric attribute vector as the prototype. 
The CUB \cite{wah2011caltech} consists of 11,788 images of 200 bird species. Among them, 150 species covering 8,855 images are used as seen classes for training, and 50 classes covering 2,933 images are used as unseen classes for testing. Each of their prototypes is represented by a 312-dimensional semantic attribute vector. 
The aPa\&Y \cite{farhadi2009describing} consists of 15,339 images from the Pascal VOC 2008 and Yahoo. Among them, 20 classes covering 12,695 images are used as seen classes for training, and 12 classes covering 2,644 images are used as unseen classes for testing. Each of the prototypes is represented by a 64-dimensional semantic attribute vector. 
The SUN \cite{patterson2014sun} consists of 14,340 scene images with two splits 707/10 and 645/72. In our model, we only consider the latter split, which contains more unseen classes. A total of 645 seen classes are used for training, and the remaining 72 unseen classes are used for testing. Each of the prototypes is represented by a 102-dimensional semantic attribute vector. 
The ImageNet \cite{russakovsky2015imagenet} consists of total 1,360 classes. Among them, 1,000 classes from ILSVRC2012 covering $2.0\times 10^5$ images are used as seen classes for training, and 360 classes from ILSVRC2010 covering $5.4\times 10^4$ images are used as unseen classes for testing. Each of their prototypes is represented by a 1,000-dimensional word vector.

In our experiments, the hit@k accuracy \cite{frome2013devise, norouzi2013zero} is used to evaluate the model performance. The hit@k accuracy is the standard evaluation metrics in ZSL which can be defined as:
\begin{equation}
	ACC_{hit@k} = \frac{\sum_{i=1}^{n} \mathbf{1}\left [ y_{i} \in \left \{ \tau^{j}\left ( x_{i} \right )  \right \}_{j=1}^{k} \right ]}{n},
\end{equation}
where $\mathbf{1}\left [ \cdot \right ]$ is an indicator function takes the value ``1'' if the argument is true, and ``0'' otherwise. $\tau(\cdot)$ is the operation which determines the class label of example. Similar with most methods, we choose hit@1 for AWA, CUB, aPa\&Y, and SUN, which is the normal accuracy, and choose hit@5 for ImageNet to fit a larger scale. 

\subsection{Experimental Setups}
\label{Experimental_Setups}
In our experiments, to consider all competitors, we also use GoogleNet \cite{szegedy2015going} to extract the visual features for image examples, from which each image is presented by a 1024-dimensional visual vector. Regarding the semantic features, we use the semantic attributes for AWA, CUB, aPa\&Y, and SUN, and use the semantic word vectors for ImageNet. In our model, the autoencoder-based network for expansion and alignment has five layers. Specifically, the encoder and decoder parts contain two layers with (1024,256) and (256,1024) neurons, respectively. The central hidden layer represents the expected latent feature space, which can be adjusted to the dimension of semantic features we expand. 

\begin{figure}[h]
    \centerline{\includegraphics[width=0.87\textwidth]{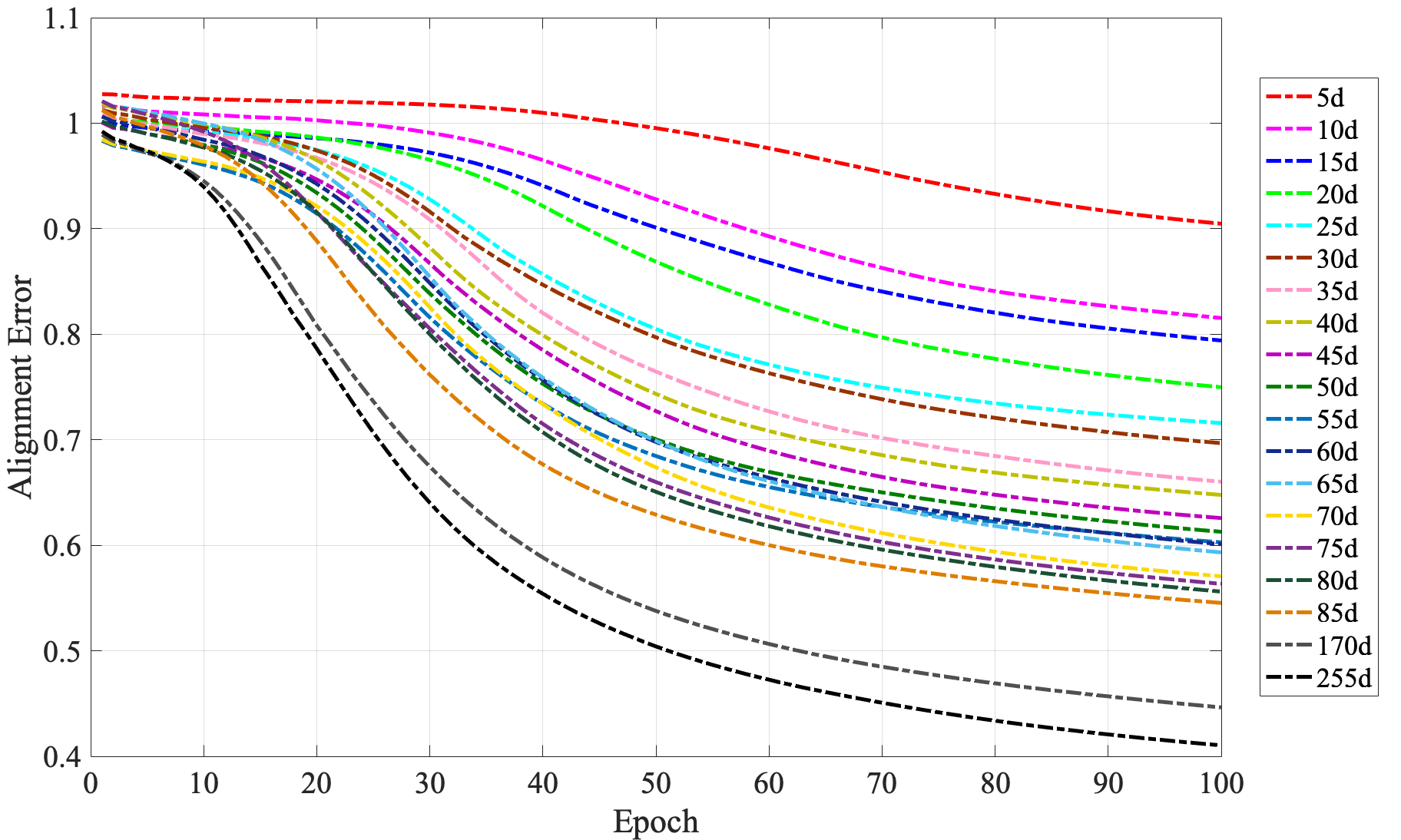}}
    \caption{Effects of expanded dimensions on alignment error. Curves in different colors represent the alignment error of different expanded dimensions ranging from 5 dimensions to 85 dimensions (expansion rate: 100\%) in an interval of 5 dimensions. In addition, we also show the results of 170 dimensions (expansion rate: 200\%) and 255 dimensions (expansion rate: 300\%) to better demonstrate the trends (better viewed in color).}
    \label{fig3}
\end{figure}

\begin{figure}[h]
    \centerline{\includegraphics[width=0.87\textwidth]{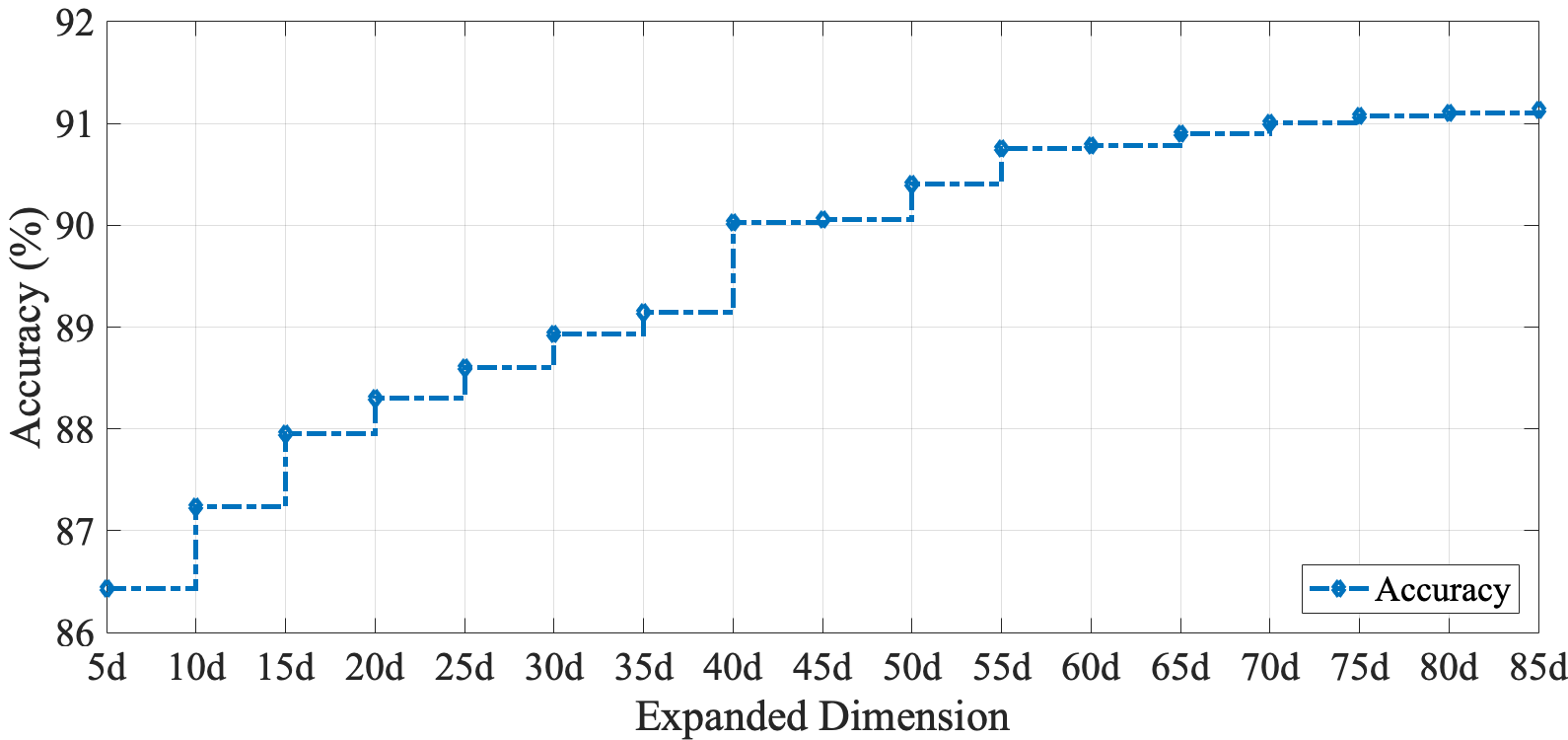}}
    \caption{Effects of expanded dimensions on prediction accuracy. The stepped curve represents the prediction accuracy under different expanded dimensions ranging from 5 dimensions to 85 dimensions (expansion rate: 100\%) in an interval of 5 dimensions.}
    \label{fig4}
\end{figure}

Theoretically, within a reasonable expansion range, we can expect that the more auxiliary semantic features we expand, the better the model performance will be. This is because with more auxiliary semantic features, more information can be utilized when training and inferring for the recognition. It will also be easier to align the manifold structures between the visual and semantic feature spaces; thus, the projection between these two feature spaces can be better trained and further mitigate the domain shift problem. Taking the AWA as an example, we conduct a comparison experiment to evaluate the effects of expanded dimensions of the auxiliary semantic features. The comparison results are shown in Figs. \ref{fig3} and \ref{fig4}. From Fig. \ref{fig3}, we can observe that within a reasonable expansion range, e.g., within an expansion rate of 100\% ($d_{expanded}/d_{pre-defined}=1$), as the expanded dimensions increase, the alignment error gradually decreases. In addition, we can also observe that even when the expansion rate reaches 200\% and 300\%, the trend of alignment error reduction remains. As to the prediction accuracy, it can be seen from Fig. \ref{fig4} that as the expanded dimensions increase, the accuracy also gradually increases and eventually becomes comparatively stable with larger expansion rates. Based on the above analysis, in our experiments, without loss of generality, we empirically apply a relatively medium expansion rate, i.e., $60\%\pm 15\%$, for all datasets except ImageNet. We expand 65, 138, 26, and 58 auxiliary semantic features for AWA, CUB, aPa\&Y, and SUN to fit the total semantic features as nice round numbers for 150, 450, 90, and 160. Regarding the ImageNet, because the dimensions of the visual features and predefined semantic features are 1,024 and 1,000, respectively, which results in expandable auxiliary semantic features ranging from 0 to 24 for a better projection. Thus, we fairly expand 12 auxiliary semantic features for ImageNet. In our model, the dimensional comparison of predefined and expanded semantic features for these five datasets is shown in Table \ref{tab1}. As mentioned above, the neurons of the central hidden layer are adjusted to 65, 138, 26, 58, and 12 for AWA, CUB, aPa\&Y, SUN, and ImageNet, respectively.

\begin{table}[t]
    \begin{center}
        \caption{Dimension of predefined (P) / expanded (E) features}
        \setlength{\tabcolsep}{1.8mm}{ 
        \label{tab1} 
            \begin{tabular}{cccccc}        
                \hline                   
                           & AWA     & CUB     & aPa\&Y  & SUN     & ImageNet\\
                \hline
                P                & 85      & 312     & 64      & 102     & 1000 \\
                E                & 65      & 138     & 26      & 58      & 12 \\
                P+E              & 150     & 450     & 90      & 160     & 1012 \\ 
                \hline  
        \end{tabular}}
    \end{center}    
\end{table}

\begin{figure}[h]
    \centerline{\includegraphics[width=0.67\textwidth]{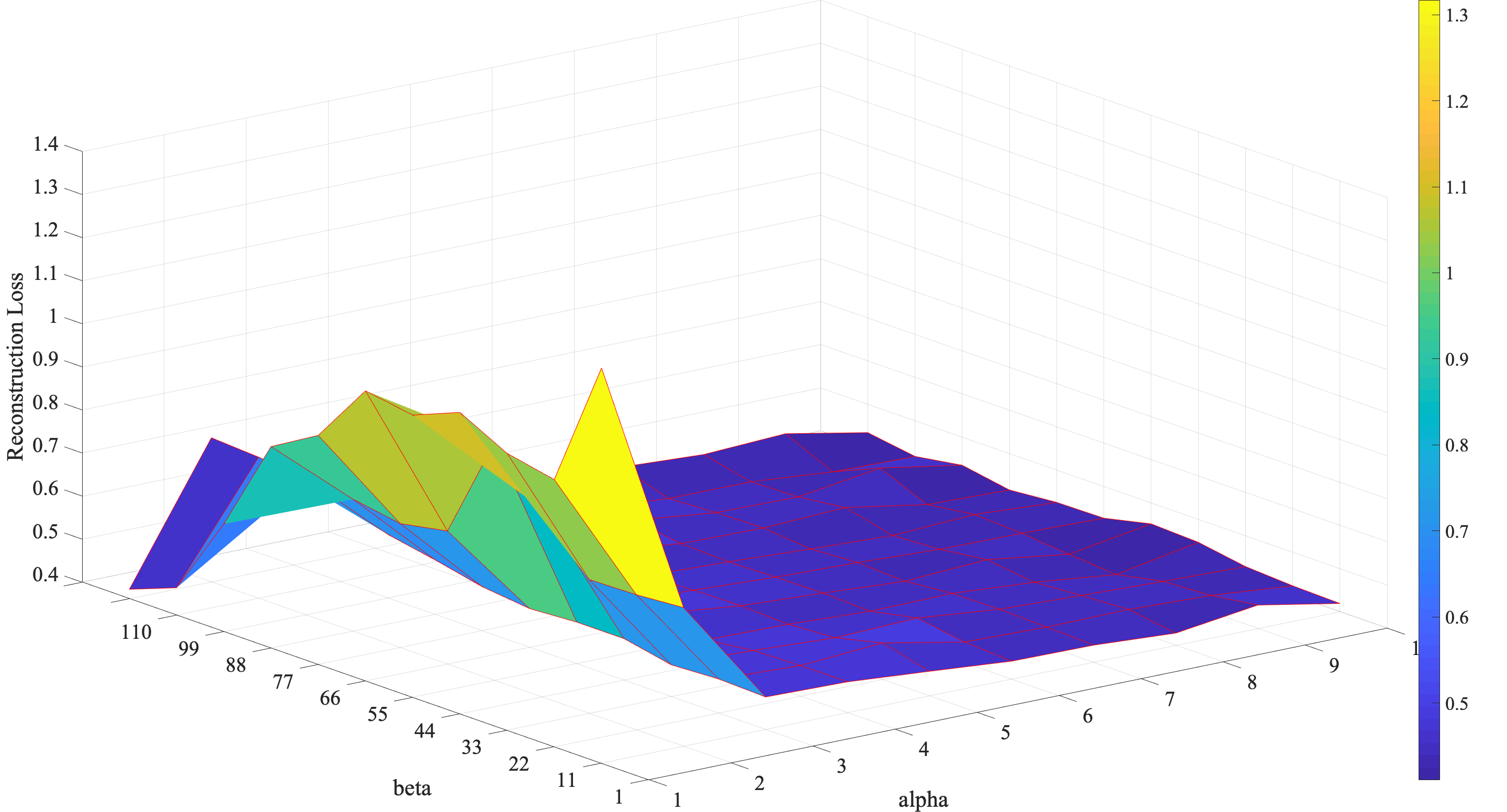}}
    \caption{The sensibilities of hyper-parameters $\alpha$ and $\beta$ on reconstruction loss.}
    \label{l1}
\end{figure}

For the hyper-parameters $\alpha$ and $\beta$ in Eq. \ref{eq16}, we conduct a grid-search experiment \cite{hsu2003practical,guo2016improved} to evaluate their sensibilities. Based on the definition in Eq. \ref{eq15}, the alignment loss is normally considerably smaller than the reconstruction loss. Thus, a larger penalty, i.e., $\beta$, should be applied to the alignment term in Eq. \ref{eq16} to accelerate the convergence of alignment loss \cite{kendall2018multi}. In our grid-search experiment, we set the $\alpha$ and $\beta$ ranges within $[1, 10]$ and $[1, 110]$, respectively. The grid-search results are shown in Figure. \ref{l1}$\sim$\ref{l12}, from which we can observe that the optimal $\alpha / \beta$ selections are 9/77 for the reconstruction loss, 10/110 for the alignment loss, and 9/77 for the total loss. Therefore, we choose to set the hyper-parameters $\alpha$ and $\beta$ as 9 and 77, respectively. The hyperparameter $g$ is used to update the class prototypes of unseen classes. Based on our previous work \cite{guo2019adaptive} which adopted the same strategy by linearly associating with seen class prototypes, $g$ is not a critical parameter when ranged on a reasonable scale, e.g., several neighbors to tens of neighbors. Therefore, without loss of generality, we empirically choose to consider 8 nearest neighbors in our model. As to the recognition phase, a simple linear autoencoder with just one hidden layer is trained to obtain the visual-semantic projection, i.e., the neurons of input/output are equal to the visual features, and the neurons of the central hidden layer are equal to the semantic features. The cosine similarity is applied to search the related prototype of the testing examples.

\begin{figure}[h]
    \centerline{\includegraphics[width=0.67\textwidth]{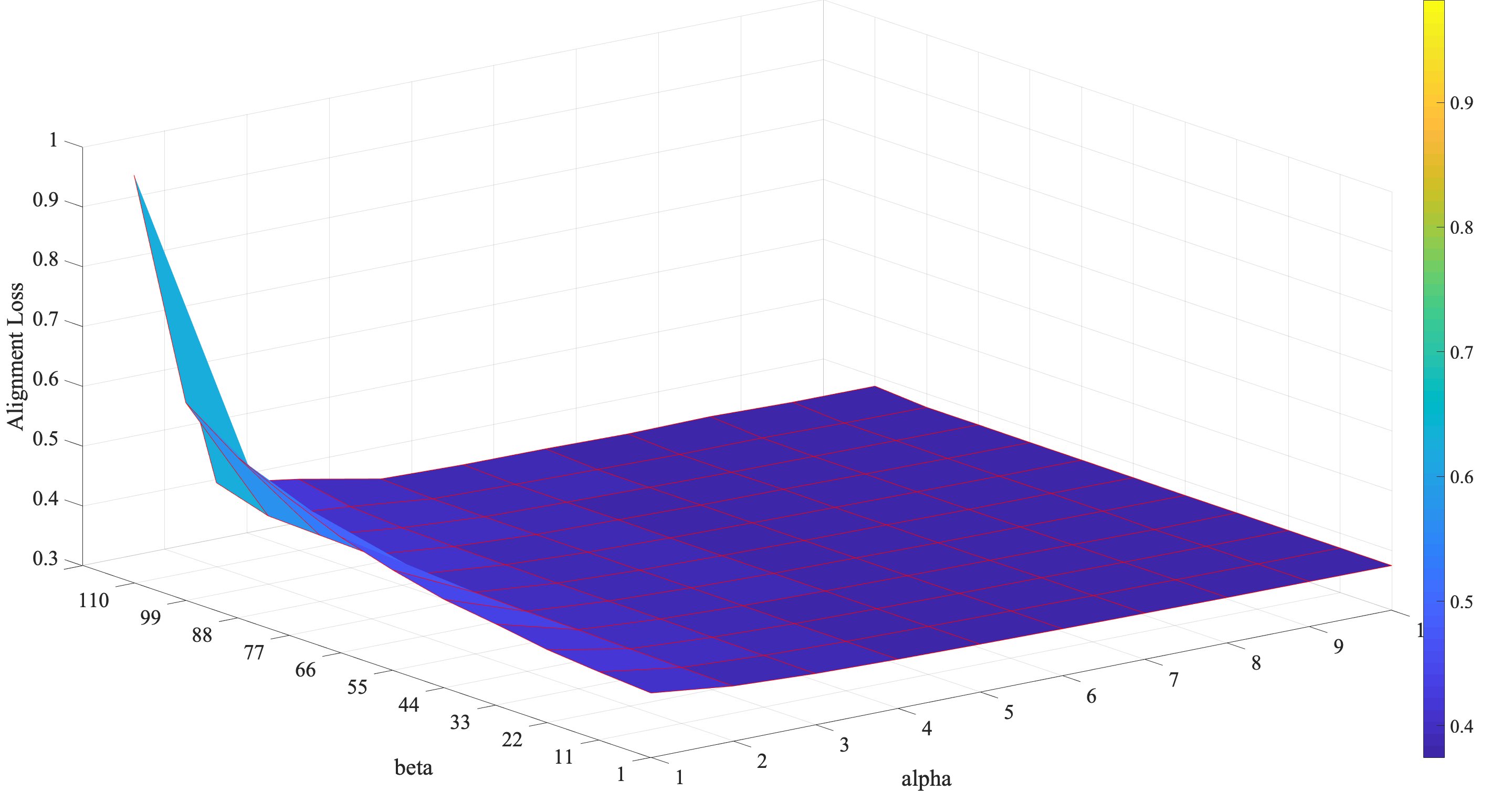}}
    \caption{The sensibilities of hyper-parameters $\alpha$ and $\beta$ on alignment loss.}
    \label{l2}
\end{figure}

\begin{figure}[h ]
    \centerline{\includegraphics[width=0.67\textwidth]{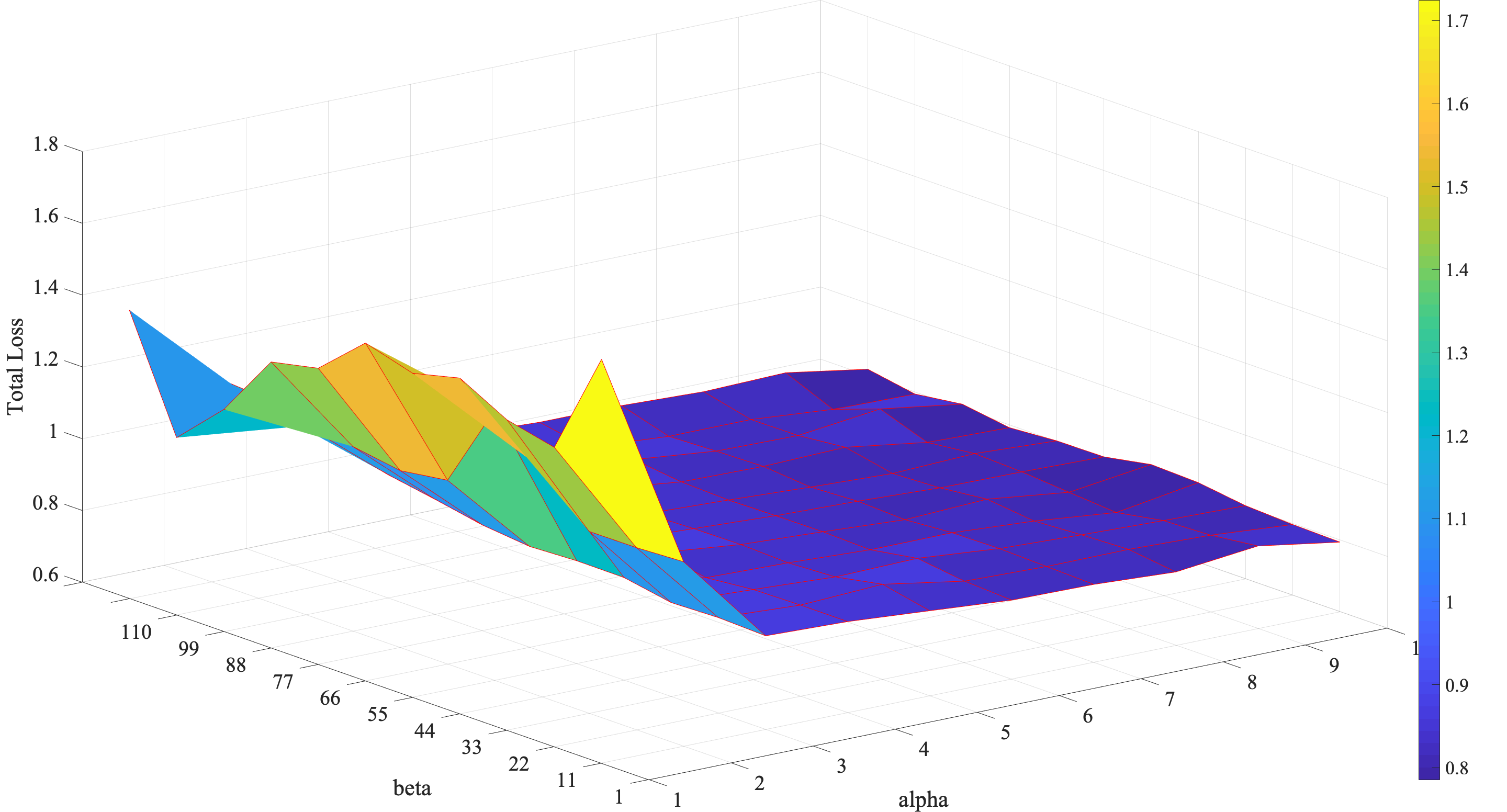}}
    \caption{The sensibilities of hyper-parameters $\alpha$ and $\beta$ on total loss.}
    \label{l12}
\end{figure}

\clearpage
\subsection{Results and Analysis}

\subsubsection{Competitors}
This section demonstrates the experimental results in detail. Our model is compared with several competitors including DeViSE \cite{frome2013devise}, DAP \cite{lampert2014attribute}, MTMDL \cite{yang2014unified}, ESZSL \cite{romera2015embarrassingly}, SSE \cite{zhang2015zero}, RRZSL \cite{shigeto2015ridge}, \textit{Ba et al.} \cite{ba2015predicting}, AMP \cite{bucher2016improving}, JLSE \cite{zhang2016zero}, SynC$^{struct}$ \cite{changpinyo2016synthesized}, MLZSC \cite{bucher2016improving}, SS-voc \cite{fu2016semi}, SAE \cite{kodirov2017semantic}, CLN+KRR \cite{long2017zero}, MFMR \cite{xu2017matrix}, RELATION NET \cite{sung2018learning}, CAPD-ZSL \cite{rahman2018unified}, LSE \cite{yu2018zero}, Chen \emph{et al.} \cite{chen2019learning}, and SGAL \cite{yu2019zero}. The selection standard for these competitors is based on following criteria: 1) all of these competitors are published in the most recent years; 2) they cover a wide range of models; 3) all of these competitors are under the same settings, i.e., datasets, evaluation criteria, etc.; and 4) they clearly represent the state-of-the-art. Moreover, as mentioned in previous, our model and all selected competitors strictly comply with the non-transductive zero-shot setting that the training only relies on seen class examples, while the unseen class examples are only available during the testing phase.

\subsubsection{Comparison Results}
The comparison results with the selected state-of-the-art competitors are shown in Table \ref{results}. It can be seen from the results that our model outperforms all competitors with great advantages in four datasets, including AWA, CUB, aPa\&Y, and SUN. The prediction accuracy of our model achieves 90.9\%, 70.1\%, 59.7\%, and 92.9\%, respectively. In ImageNet, our model obtains suboptimal performance among all competitors. Our prediction accuracy is 26.3\%, which is slightly weaker than LSE \cite{yu2018zero} and similar to SAE \cite{kodirov2017semantic}. It should be noted that these similar results in ImageNet of SAE and our model may be caused by the following two reasons. First, as mentioned in Sections \ref{Recognition} and \ref{Experimental_Setups}, our model also adopts a simple autoencoder training framework to obtain the projection between the visual and semantic feature spaces, which is similar to some existing methods such as SAE. Second, it can be seen in Table \ref{tab1} that there are only 12 expanded semantic features for ImageNet, which also makes the total semantic features similar to the predefined features. From the dataset description in Section \ref{Datasets_and_Metrics}, we can know that the dimension of the predefined semantic features of ImageNet is 1,000. However, the dimension of the visual features is 1,024. Based on the analysis in Section \ref{Experimental_Setups}, we only expand 12 auxiliary semantic features for ImageNet, which are far fewer than the predefined features. This limitation makes it difficult to perform the alignment between the visual and semantic feature spaces for ImageNet. Thus, the improvement regarding ImageNet is not that significant. In contrast, we have enough space to expand the auxiliary semantic features for AWA, CUB, aPa\&Y, and SUN, so the alignment can be better approximated for them. In our model, the dimensions of the expanded auxiliary semantic features for these , datasets are 65, 138, 26, 58, and 12, respectively.

From the comparison results, we can also observe that the performance of our VAE-based model is slightly better than the AE-based model for CUB, aPa\&Y, SUN, and ImageNet, from which the most significant improvement is from 67.8\% to 70.1\% for CUB. One possible reason is that CUB is also a good benchmark dataset for fine-grained image recognition tasks, except for ZSL, which consists of multiple bird species, and the transition between each class is relatively smoother and more continuous. Therefore, it is more likely to maintain these features in the semantic feature space, which helps the prediction by using the VAE.

\begin{sidewaystable*}[h]
    \centering
    \setlength{\tabcolsep}{3.9mm}{
    \begin{threeparttable}  
        \caption{Comparison with state-of-the-art competitors}  
        \label{results}    
        \begin{tabular}{lcccccccccc}  
            \toprule  
            \multirow{2}{*}{Method}&  
            \multicolumn{2}{c}{AWA}&\multicolumn{2}{c}{CUB}&\multicolumn{2}{c}{aPa\&Y}&\multicolumn{2}{c}{SUN}&\multicolumn{2}{c}{ImageNet}\cr  
            \cmidrule(lr){2-3} \cmidrule(lr){4-5} \cmidrule(lr){6-7} \cmidrule(lr){8-9} \cmidrule(lr){10-11} 
            &SS  &ACC               &SS   &ACC              &SS   &ACC                 &SS   &ACC              &SS   &ACC\cr  
            \midrule  
            DeViSE \cite{frome2013devise}            &A/W &56.7/50.4         &A/W  &33.5             &-    &-                   &-    &-                &A/W  &12.8\cr
            DAP \cite{lampert2014attribute}               &A   &60.1              &A    &-                &A    &38.2                &A    &72.0             &-    &-\cr
            MTMDL \cite{yang2014unified}             &A/W &63.7/55.3         &A/W  &32.3             &-    &-                   &-    &-                &-    &-\cr
            ESZSL \cite{romera2015embarrassingly}             &A   &75.3              &A    &48.7             &A    &24.3                &A    &82.1             &-    &-\cr
            SSE \cite{zhang2015zero}               &A   &76.3              &A    &30.4             &A    &46.2                &A    &82.5             &-    &-\cr
            RRZSL \cite{shigeto2015ridge}             &A   &80.4              &A    &52.4             &A    &48.8                &A    &84.5             &W    &-\cr
            \textit{Ba et al.}  \cite{ba2015predicting}         &A/W &69.3/58.7         &A/W  &34.0             &-    &-                   &-    &-                &-    &-\cr    
            AMP \cite{bucher2016improving}               &A+W &66.0              &A+W  &-                &-    &-                   &-    &-                &A+W  &13.1\cr
            JLSE \cite{zhang2016zero}              &A   &80.5              &A    &41.8             &A    &50.4                &A    &83.8             &-    &-\cr
            SynC$^{struct}$ \cite{changpinyo2016synthesized}   &A   &72.9              &A    &54.4             &-    &-                   &-    &-                &-    &-\cr
            MLZSC \cite{bucher2016improving}             &A   &77.3              &A    &43.3             &-    &53.2                &A    &84.4             &-    &-\cr
            SS-voc \cite{fu2016semi}            &A/W &78.3/68.9         &A/W  &-                &-    &-                   &-    &-                &A/W  &16.8\cr
            SAE \cite{kodirov2017semantic}              &A   &84.7              &A    &61.2             &A    &55.1                &A    &91.0             &W    &26.3\cr     
            CLN+KRR \cite{long2017zero}           &A   &81.0              &A    &58.6             &-    &-                   &-    &-                &-    &-\cr
            MFMR \cite{xu2017matrix}              &A   &76.6              &A    &46.2             &A    &46.4                &A    &81.5             &-    &-\cr
            RELATION NET \cite{sung2018learning}     &A   &84.5              &A    &62.0             &-    &-                   &-    &-                &-    &-\cr
            CAPD-ZSL \cite{rahman2018unified}          &A   &80.8              &A    &45.3             &A    &55.0                &A    &87.0             &W    &23.6\cr
            LSE \cite{yu2018zero}               &A   &81.6              &A    &53.2             &A    &53.9                &-    &-                &W    &\underline{27.4}\cr   
            Chen \emph{et al.} \cite{chen2019learning}       &A   &82.7             &A    &58.5             &-    &-                &A    &88.5                &-    &- \cr
            SGAL \cite{yu2019zero}       &A   &84.1             &A    &62.5             &-    &-                &-    &-                &-    &- \cr
            AMS-SFE (AE, Ours)           &A   &\underline{\bf 90.9}        &A    &{\bf 67.8}       &A    &{\bf 59.4}          &A    &{\bf 92.7}       &W    &{\bf 26.1}\cr  
            AMS-SFE (VAE, Ours)           &A   &{\bf 90.2}        &A    &\underline{\bf 70.1}       &A    &\underline{\bf 59.7}          &A    &\underline{\bf 92.9}       &W    &{\bf 26.3}\cr  
            \bottomrule  
        \end{tabular}
        \label{tab3}  
        \footnotesize{SS, A, and W represents semantic space, attribute and word vectors, respectively; '/' represents 'or' and '+' represents 'and'; '-' represents that there is no reported result. ACC stands for accuracy (\%), where Hit@1 is used for AWA, CUB, aPa\&Y, and SUN, and Hit@5 is used for ImageNet. The results of our model are shown in bold, and the underline denotes the best result among all competitors.}
    \end{threeparttable}}
\end{sidewaystable*}

\clearpage
\subsection{Further Analysis}

\subsubsection{Mapping Robustness}
\begin{figure}[h]
    \centerline{\includegraphics[width=0.99\textwidth]{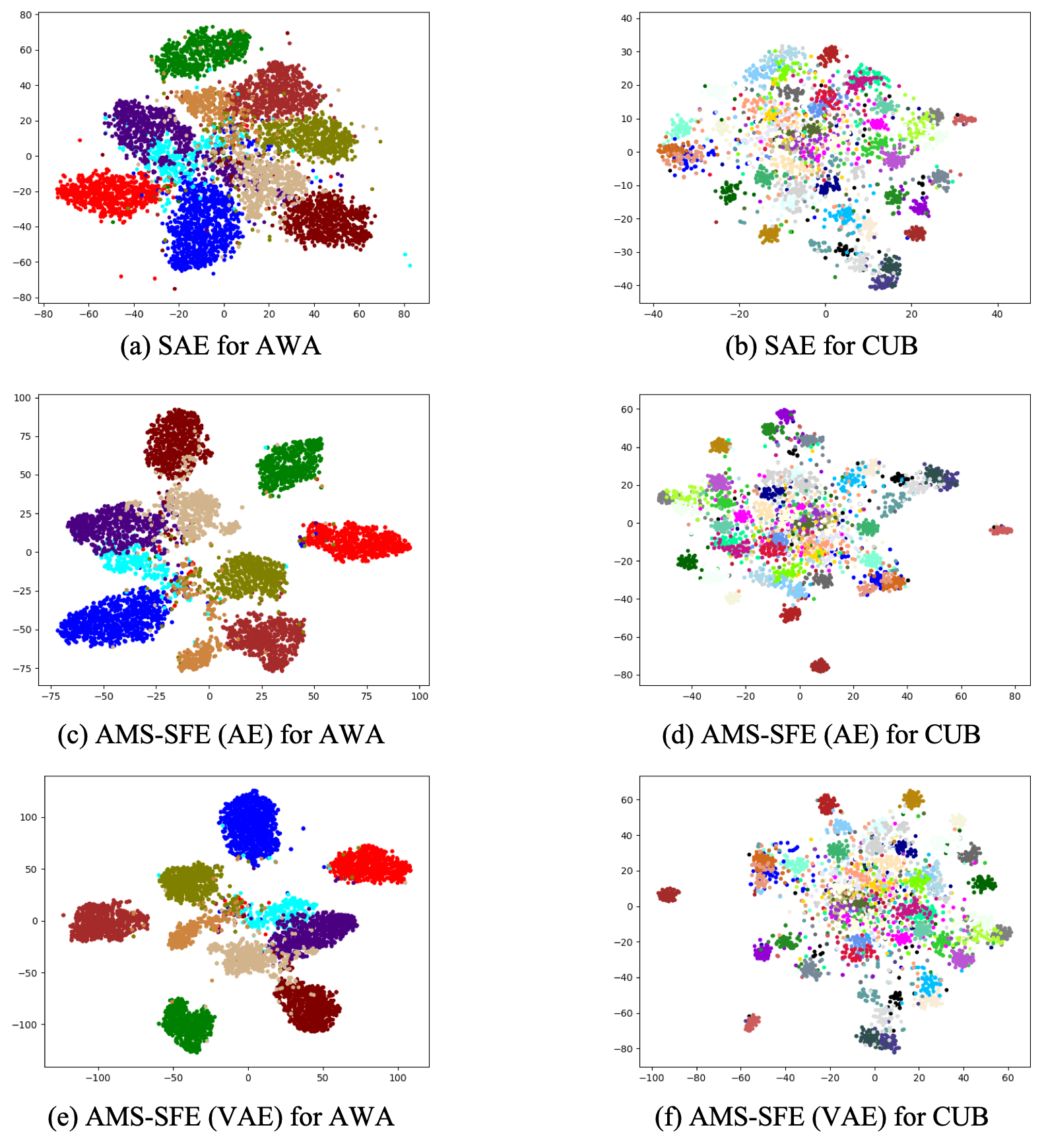}}
    \caption{Visualization results of our model and the strongest competitor SAE. The left column is the results for AWA, and the right column is the results for CUB. (a) and (b) are the visualization of SAE, (c) and (b) are the visualization of our AE-based model, and (e) and (f) are the visualization of our VAE-based model (better viewed in color).}
    \label{c4_mapping}
\end{figure}

\begin{figure}[h]
    \centerline{\includegraphics[width=0.67\textwidth]{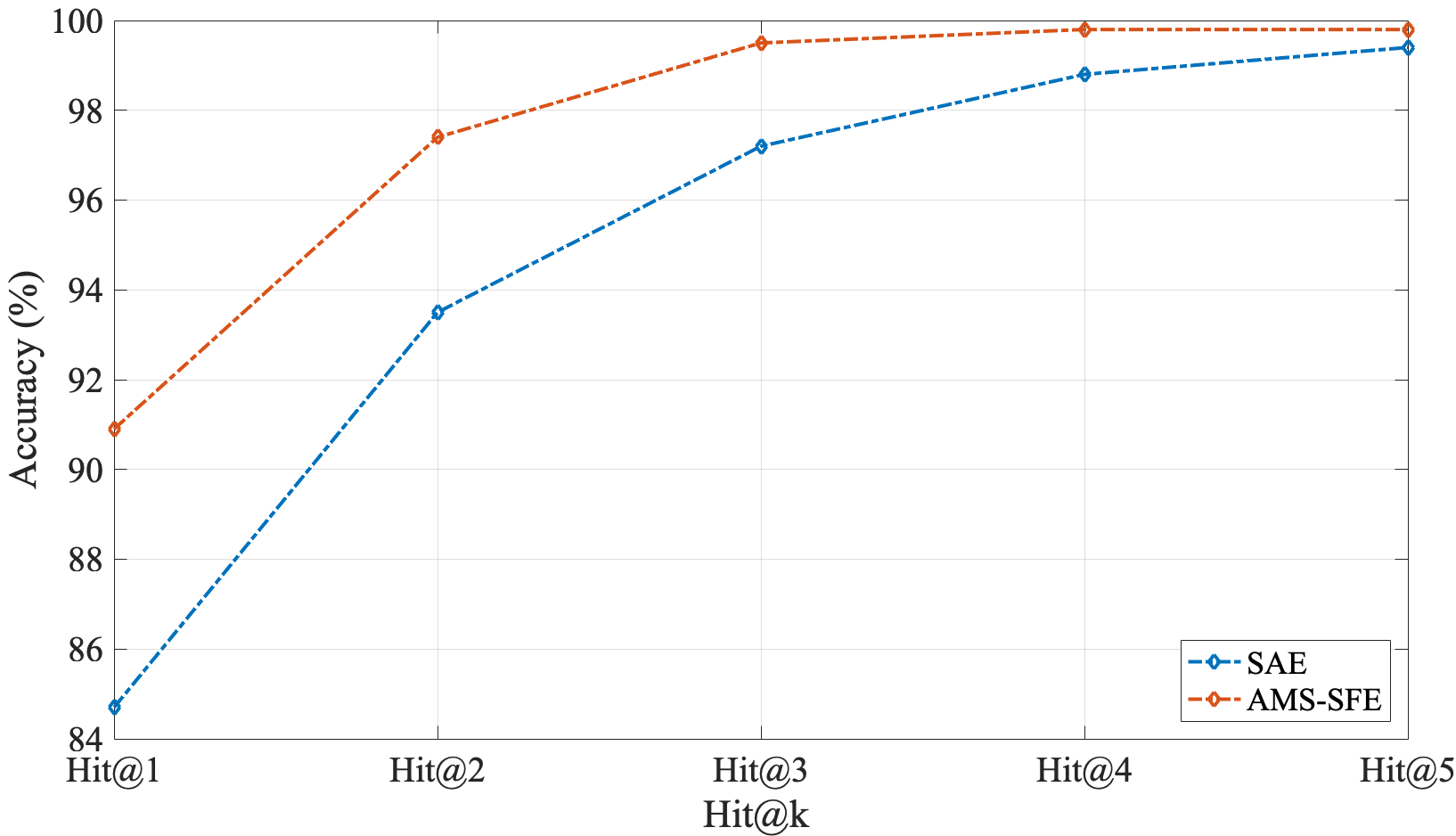}}
    \caption{Comparison of Hit@k accuracy for AWA. The blue curve denotes the SAE and the red curve denotes our model.}
    \label{hitatk}
\end{figure}

We conduct the evaluation of projection robustness on AWA and CUB. AWA consists of 10 unseen classes, and CUB consists of 50 unseen classes. Our model is compared with the overall strongest competitor SAE \cite{kodirov2017semantic} to verify the mapping robustness. A mapping function that maps from the visual to the semantic feature spaces is trained on seen class examples with our model. Then, we apply the mapping function to all unseen class examples to obtain their semantic features and visualize them in a 2D map by t-SNE \cite{maaten2008visualizing}. The visualization results are shown in Figure. \ref{c4_mapping}. The left column is the results for AWA, and the right column is the results for CUB. In AWA, we observe that by using our model, only a small portion of these unseen class examples are mis-mapped in the semantic feature space. Moreover, due to the alignment of manifold structures between the visual and semantic feature spaces, these mis-mapped examples are less shifted from their class prototypes, which means that our model can also obtain better results for Hit@k accuracy when k varies and converges faster to the best performance. The comparison is shown in Table \ref{tab3} and Figure. \ref{hitatk}. In CUB, the class number is much larger than that of AWA, so the visualization results seem more complicated and intertwined in the 2D map, but we still observe that our model can obtain more continuous and smoother semantic features in a wider visual field.

\begin{table}[h]
    \begin{center}
        \caption{Hit@k accuracy (\%) for AWA, $k\in \left [ 1,5 \right ]$}
        \label{tab3}
            \begin{tabular}{cccccc}        
                \hline                   
                Method           & Hit@1   & Hit@2   & Hit@3   & Hit@4   & Hit@5\\
                \hline
                SAE              & 84.7    & 93.5    & 97.2    & 98.8    & 99.4 \\
                AMS-SFE (ours)   & 90.9    & 97.4    & 99.5    & 99.8    & 99.8 \\ 
                \hline  
        \end{tabular}
    \end{center}   
\end{table}

\subsubsection{Fine-Grained Accuracy}
\begin{figure}[t]
    \centerline{\includegraphics[width=0.99\textwidth]{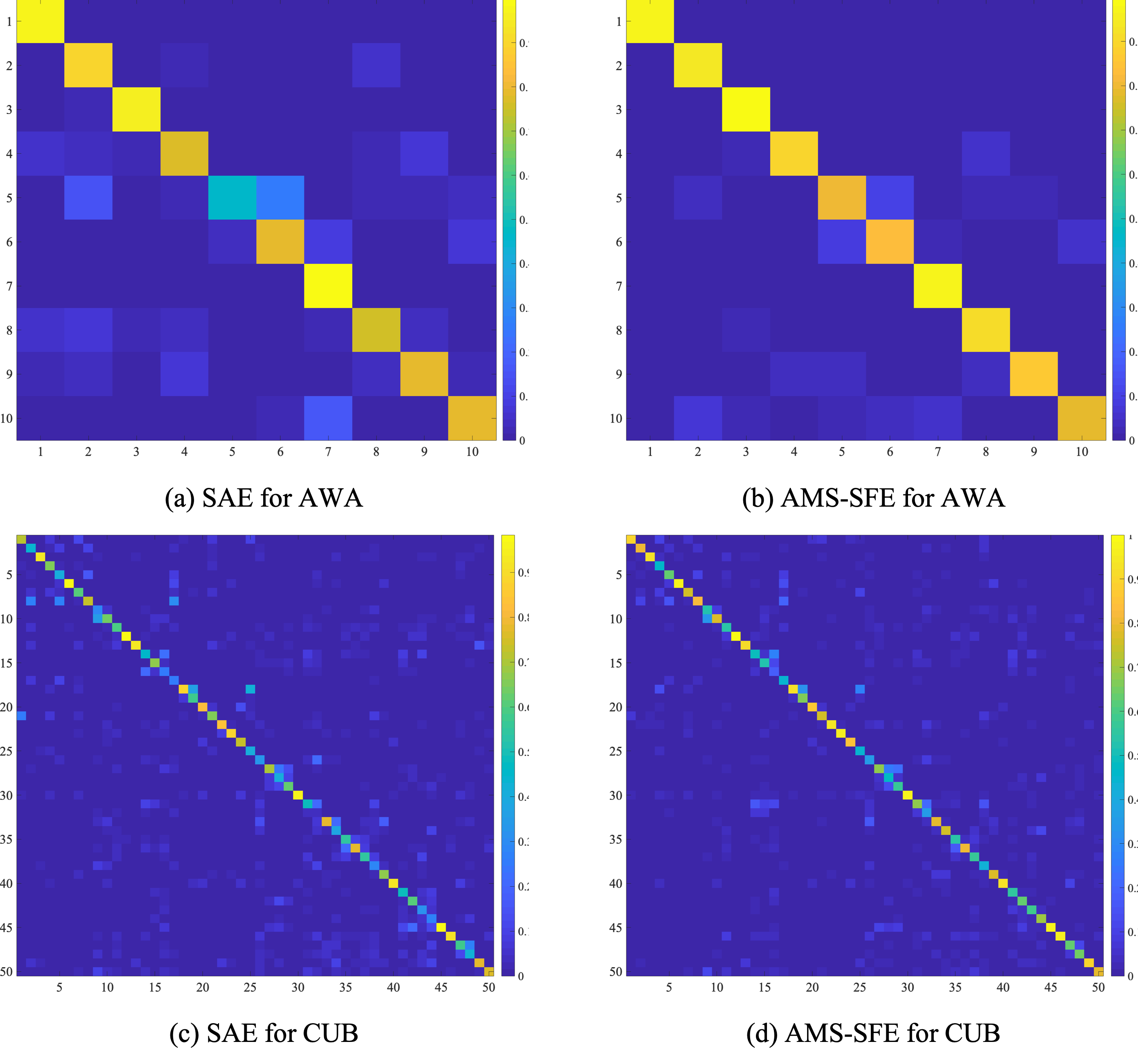}}
    \caption{The confusion matrix on AWA and CUB. The upper part is the results for AWA, and the lower part is the results for CUB. (a) and (c) are the confusion matrixes of SAE; (b) and (d) are the confusion matrixes of our model. The diagonal position indicates the classification accuracy for each class, the column position means the ground truth, and the row position denotes the predicted results (better viewed in color).}
    \label{c4_cm}
\end{figure}

To further evaluate the predictive power of our model, we record and count the prediction results for each unseen class example and analyze the per-class performance. This evaluation is conducted on AWA and CUB, and we also compare our model with the overall strongest competitor SAE \cite{kodirov2017semantic}. The results are presented by the confusion matrixes in Figure. \ref{c4_cm}. 
The upper part is the results for AWA, and the lower part is the results for CUB. In each confusion matrix, the diagonal position indicates the classification accuracy for each class. The column position indicates the ground truth, and the row position denotes the predicted results. It can be seen from the results that our model can obtain higher accuracy, along with a more balanced and robust prediction for each unseen class for both AWA and CUB.

\subsubsection{Ablation Comparison}
\begin{figure}[t] 
    \centerline{\includegraphics[width=0.87\textwidth]{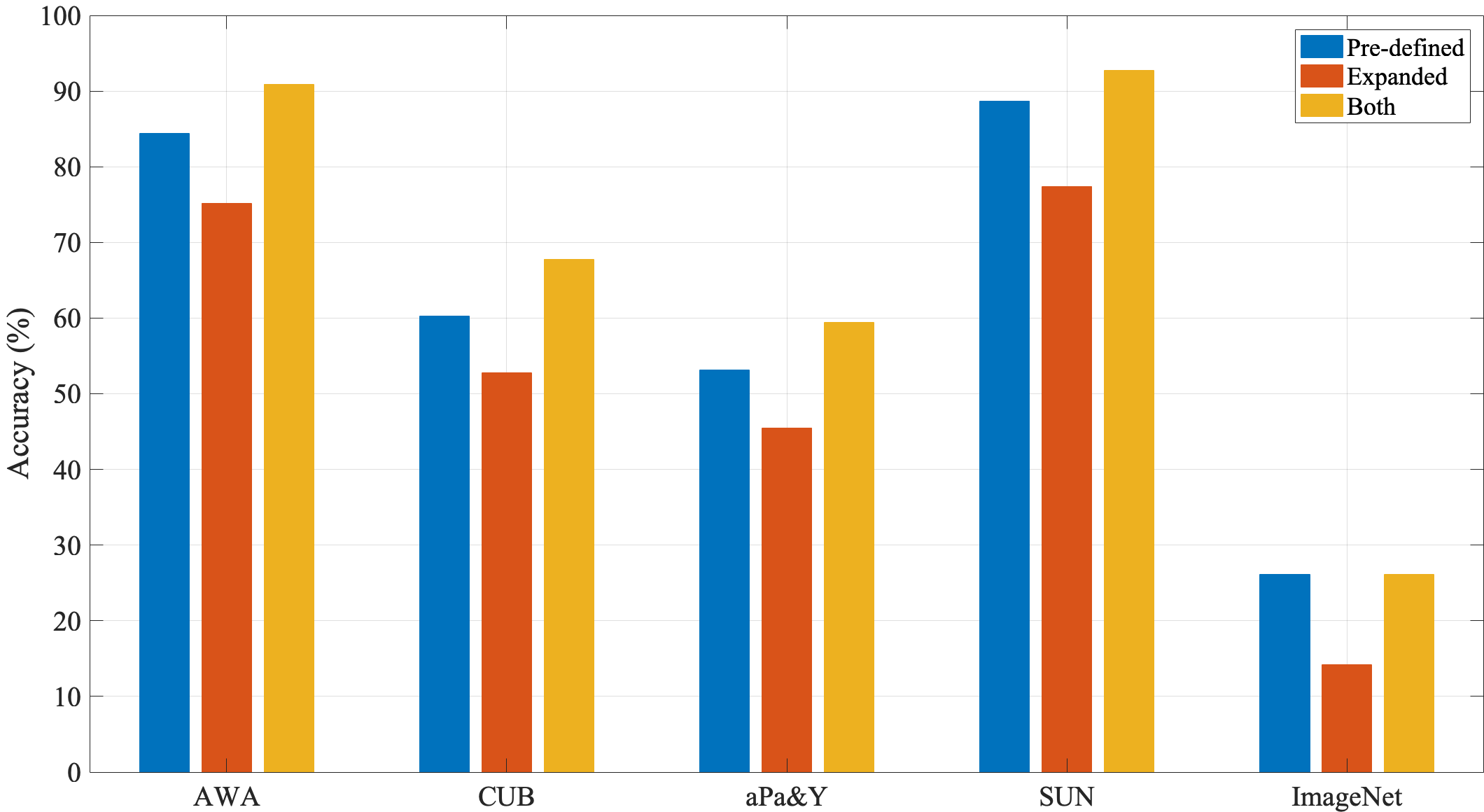}}
    \caption{Ablation comparison on five benchmark datasets. The blue bar denotes the predefined semantic feature, the yellow bar denotes the expanded semantic features, and the red bar denotes both.}
    \label{ab}
\end{figure}

We conduct an ablation experiment to further evaluate the effectiveness of our model. The performance is compared on all five benchmark datasets, including AWA, CUB, aPa\&Y, SUN, and ImageNet, with our AE-based model on three scenarios: (1) only predefined semantic features are used; (2) only expanded auxiliary semantic features are used; and (3) both predefined and expanded auxiliary semantic features are used. The comparison results are shown in Table \ref{tab_ab} and Figure. \ref{ab}. It can be seen from the results that our model greatly improves the performance of zero-shot learning by performing the alignment with the expanded auxiliary semantic features. Nevertheless, it should also be noted that, due to the very few (i.e., 12) expanded auxiliary semantic features for ImageNet as analyzed in Section \ref{Experimental_Setups}, the model obtains similar performance as 26.1\% for both scenarios (1) and (3), where only the predefined semantic features and both semantic features are used, respectively. Thus, the impact of expanded semantic features on ImageNet is limited in our model, and the visual and semantic feature spaces cannot be better aligned. We may further investigate this problem in our future work.

\begin{table}[t]
    \begin{center}
        \caption{Ablation comparison (accuracy\%) on predefined (P) / expanded (E) semantic features and both (P+E)}
          \label{tab_ab}  
            \begin{tabular}{cccccc}        
                \hline                   
                           & AWA     & CUB     & aPa\&Y  & SUN     & ImageNet\\
                \hline
                P                & 84.4    & 60.3    & 53.1    & 88.7  & 26.1 \\
                E                & 75.2    & 52.8    & 45.5    & 77.4    & 14.2 \\
                P+E              & 90.9    & 67.8    & 59.4    & 92.7    & 26.1 \\ 
                \hline  
        \end{tabular}
    \end{center} 
\end{table}

\begin{figure}[t] 
    \centerline{\includegraphics[width=0.99\textwidth]{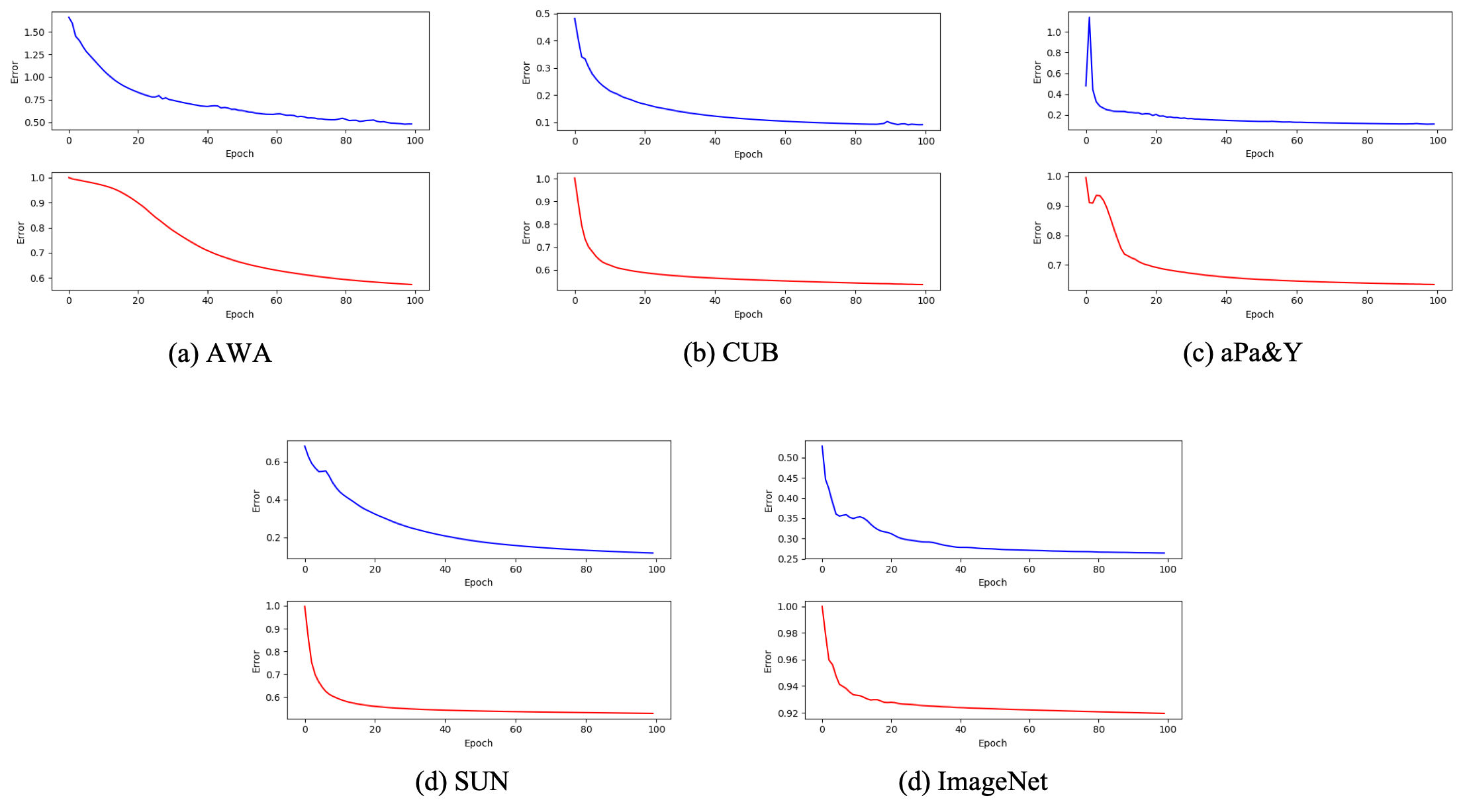}}
    \caption{The convergence curves of AWA, CUB, aPa\&Y, SUN, and ImageNet. The blue curve denotes the reconstruction error, and the red curve denotes the alignment error.}
    \label{c4_convergence}
\end{figure}

\clearpage
\section{Remarks}
\label{Remarks}
In this chapter, we propose a novel model called AMS-SFE for zero-shot learning. Our model aligns the manifold structures between the visual and semantic feature spaces by jointly conducting semantic feature expansion. Our model can better mitigate the domain shift problem and obtain a more robust and generalized visual-semantic projection. 
In the future, we have two research routes to further improve zero-shot learning. The first route investigates a more efficient and generalized method to further empower the semantic feature space. The second route goes from the coarse-grained model to the fine-grained model to better model the subtle differences among different classes.

\chapter{Zero-shot Learning as a Graph Recognition}\label{ch:5} 

To mitigate the domain shift problem, our previous works focus on two forms of adjustments on the semantic feature space to make the trained mapping function between the visual and semantic feature spaces more robust and accurate. \Cref{ch:3} proposes to adaptively adjust to rectify the semantic feature space regarding the class prototypes and global distribution, which can be considered as a hard adjustment. While \cref{ch:4} proposes to align the manifold structures between the visual and semantic feature spaces by semantic feature expansion, which is more conservative for not directly adjusting previous semantic features and can be regarded as a soft adjustment. Like most existing works, these methods consider constructing mapping functions in a coarse-grained manner, i.e., features are extracted from the whole examples.

In this chapter, unlike our previous and most existing works, our interest is to focus on the fine-grained perspective based on example-level graph. Specifically, we decompose an image example into several parts and use a graph-based model to measure and utilize certain relations between these parts. Taking advantage of recently developed graph neural networks, we formulate the zero shot learning (ZSL) problem to a \emph{graph-to-semantic mapping} problem, which can better exploit the \emph{part-semantic correlation} and local substructure information in examples. Experimental results demonstrate that the proposed method can mitigate the domain shift problem and achieve competitive performance against other representative methods. 

\section{Introduction}
In recent years, deep learning as one of the state-of-the-art techniques in current advanced machine learning, has attracted tremendous research interests in the domains of computer vision, natural language processing, speech analysis, and so on. The success of deep learning relies heavily on the emergence of large-scale annotated datasets such as ImageNet in the domain of computer vision. However, conventional deep learning techniques are still not capable of handling the issue of scalability that collecting and labeling examples for rare or fine-grained classes are difficult. Thus, the conventional deep learning techniques perform poorly in scenarios where only limited or even no training data are available. In the real-world application, these limitations naturally exist because novel classes of examples arise frequently, yet to obtain their annotations is expensive and time-consuming. Zero-shot learning (ZSL), which aims to imitate human ability in recognizing unseen classes under such restrictions, has received increasing attention in the most recent years \cite{lampert2009learning,frome2013devise,shigeto2015ridge,kodirov2017semantic,guo2020novel}. ZSL takes utilization of seen classes with labeled examples and auxiliary knowledge that can be shared and transferred between seen and unseen classes to solve the problem. This knowledge, e.g., semantic descriptions that exist in a high dimensional semantic feature space, can represent meaningful high-level information about examples of different classes. Intuitively, the cat is semantically closer to the tiger than to the snake. This intuition also holds and becomes the fundamental basis of ZSL in the the semantic feature space, in which each class is embedded and endowed with a prototype. In ZSL, the common practice is first to map an unseen class example from its original feature space, e.g., visual feature space, to semantic feature space by a mapping function trained on seen classes. With such semantic features, we then search the most closely related (measured by similarity or any other metrics) prototype whose corresponding class is set to this example. Specifically, the ZLS can be further restricted to the conventional ZSL (CZSL) and generalized ZSL (GZSL), where the former only considers the unseen classes during inference while the latter can also generalize well to novel examples from seen classes.

In ZSL, one non-negligible fundamental issue is that the visual and semantic feature spaces are generally independent and may exist in entirely different manifolds. This issue brings about the ubiquitous domain shift problem when generalizing the obtained knowledge, e.g., the trained mapping function, from seen to unseen classes. Recently, several methods have been proposed to mitigate the domain shift problem. For example, some methods consider directly doing the alignment between these two feature spaces when constructing the mapping function \cite{zhang2015zero,schonfeld2019generalized,guo2020novel}. Differently, some methods try to synthesize unseen class examples/features based on generative adversarial networks (GANs) or variational autoencoders (VAEs) and involve them in the training process \cite{kumar2018generalized,xian2018feature,huang2019generative}. Some other methods apply the encoder-decoder structure to maintain the robustness of mapping function \cite{kodirov2017semantic,yu2018zero}. However, most existing methods are trained on whole feature representations while ignored the fine-grained and structural information of examples from different classes. This limitation still hinders the domain shift problem being solved and makes it difficult to fully utilize the information of examples.

\begin{figure}[h]
  \centering
  \centerline{\includegraphics[width=0.99\textwidth]{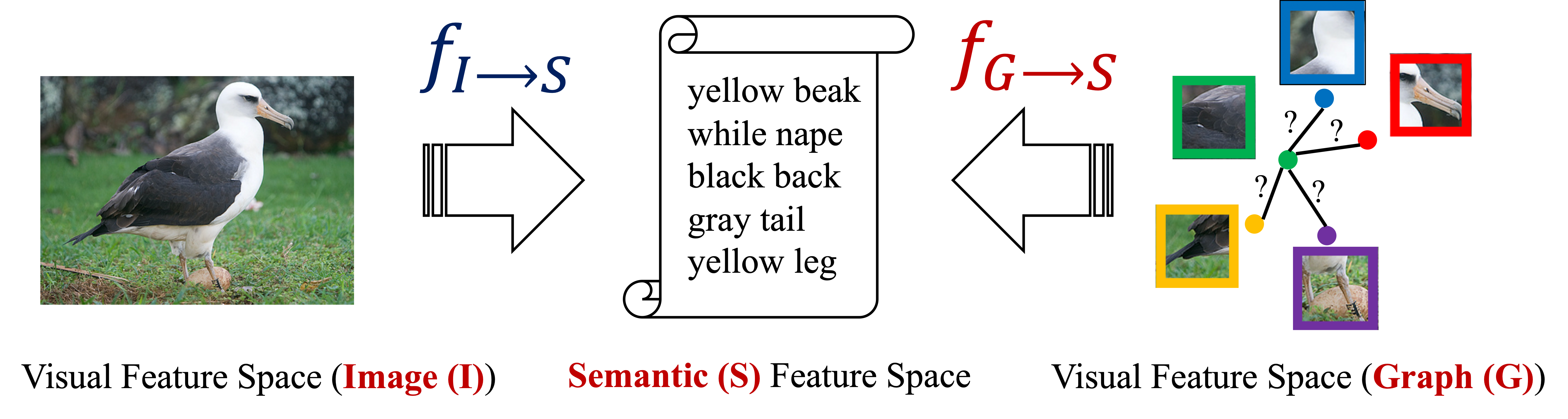}}
  \caption{Conventional whole feature representations and our method. Left part: conventional methods usually map from the whole example (image) visual features to semantic descriptions. Right part: our method maps from the decomposed example graph to semantic descriptions.}
  \label{fig1}
\end{figure}

To address these problems, we propose a fine-grained ZSL framework based on the example-level graph (Figure \ref{fig1}). Our method first decomposes an example to several parts, e.g., by key-point localization and cropping, and then constructs an example graph to present it. Specifically, we use the nodes to present each part of the example and use the linking edges to present whether a relation exists between two parts. Next, we build upon the currently popular graph neural networks (GNNs) to extract and fuse the example graph, and further formulate the original ZSL to a \textbf{graph-to-semantic mapping} problem which indeed converts the ZSL to a graph recognition task and soundly becomes our main contribution of this paper. Our method can benefit the ZSL in at least two folds. On the one hand, because the semantic descriptions are usually connected more to local features of examples rather than the global ones. Thus, the fine-grained decomposition can make the mapping more robust and conserves the part-semantic correlation. On the other hand, by using the GNNs we can fully extract the local substructure information of example graphs and fuse them to better feature representations. The above benefits ultimately resulting in a better mapping between the visual and semantic feature spaces, which can greatly mitigate the domain shift problem. 

The rest of this paper is organized as follows. Section \ref{c5:Related_Work} introduces the related work. Then, in Section \ref{c5:Methodology}, we present our proposed method. Section \ref{c5:Experiments} discusses the experiments and the remarks are addressed in Section \ref{Remarks}.

\section{Related Work}
\label{c5:Related_Work}

\subsection{Existing Works}
ZSL can be roughly divided into three categories according to the mapping direction that adopted as visual to semantic mapping, semantic to visual mapping, and intermedia mapping (metric learning). Most existing methods map the visual features to semantic feature space spanned by class descriptions and then perform nearest neighbor search \cite{lampert2009learning,frome2013devise,zhang2015zero,romera2015embarrassingly,akata2015evaluation,kodirov2017semantic,long2017zero,yu2018zero,schonfeld2019generalized,guo2019ams,guo2020novel,zhu2019semantic}. For example, DeViSE \cite{frome2013devise} trains a linear mapping function between visual and semantic feature spaces by an effective ranking loss formulation. ESZSL \cite{romera2015embarrassingly} applie the square loss to learn the bilinear compatibility and adds regularization to the objective with respect to Frobenius norm. SSE \cite{zhang2015zero} proposes to use the mixture of seen class parts as the intermediate feature space. SynC$^{struct}$ \cite{changpinyo2016synthesized} and CLN+KRR \cite{long2017zero} propose to jointly embed several kinds of textual features and visual features to ground attributes. SAE \cite{kodirov2017semantic} proposes to use a linear semantic autoencoder to regularize the zero-shot learning and makes the model generalize better to unseen classes. In contrast, some works propose to map the semantic features into visual feature space and point out that using semantic space as shared latent space may reduce the variance of features which makes results more clustered as hubs \cite{shigeto2015ridge,zhang2017learning}. Different from the directly mappings, some works propose to learn a metric network or compatibility measurement that takes paired visual and semantic features/classifiers as inputs and calculates their similarities \cite{sung2018learning,wang2018zero,kampffmeyer2019rethinking}. For example, RELATION NET \cite{sung2018learning} learns a metric network that takes paired visual and semantic features as inputs and calculates their similarities. Recently, some generative models have emerged. For example, f-CLSWGAN \cite{xian2018feature} proposes to synthesize several example features for unseen classes by using a Wasserstein GAN with a classification loss. SE-GZSL \cite{kumar2018generalized} uses a variational autoencoder based architecture to generate novel examples from seen/unseen classes conditional on respective class attributes. DASCN \cite{ni2019dual} proposes two GAN networks, namely the primal GAN and dual GAN in a unified framework, for generalized zero-shot recognition. Schonfeld \emph{et al.} \cite{schonfeld2019generalized} applied the variational autoencoder to do the cross-alignment of the visual and semantic feature spaces. SGAL \cite{yu2019zero} uses the variational autoencoder with class-specific multi-modal prior to learn the conditional distribution of seen and unseen classes. AMS-SFE \cite{guo2019ams,guo2020novel} proposes to expand the semantic feature space and implicitly to align the visual and semantic feature spaces.

Compared with above existing methods, our work is more likely related to some fine-grained ZSL methods such as ${\rm S^2GA}$ \cite{ji2018stacked}, \emph{Zhu et al.} \cite{zhu2019semantic} and \emph{Chen et al.} \cite{chen2019learning}, which make use of attention mechanism on local regions or learn dictionaries through joint training with examples, attributes and labels, and some graph-oriented ZSL methods such as \emph{Wang et al.} \cite{wang2018zero} and DGP \cite{kampffmeyer2019rethinking} which perform graph techniques to explore the class-level dependencies. Our method differs from above fine-grained and graph-oriented ZSL methods in that it decomposes an example to several parts and present it as a graph structure, which considers the example-level graph rather than class-level graph and converts the ZSL to graph-to-semantic mapping problem. It should be noted that our method is not in the same category of graph-oriented ZSL methods such as \emph{Wang et al.} \cite{wang2018zero} and DGP \cite{kampffmeyer2019rethinking}, and cannot be directly compared. Our method focuses on fine-grained stream.

\subsection{Graph Recognition with GNNs}
The GNNs are currently popular graph techniques in deep learning and attracts increasing attention from machine learning and data mining communities. We can train the GNNs in either a supervised or unsupervised manner to handle multiple tasks such as node classification, edge prediction, graph embedding, and graph classification. Specifically, the graph-level classification aims at classifying an entire graph \cite{zhang2018end,ying2018hierarchical,zhao2019learning}. Our method makes use of the graph-level classification to converts the ZSL to a graph-to-semantic mapping problem, where we replace the class labels to class semantic descriptions and indeed form a graph-level regression task.

\section{Methodology}
\label{c5:Methodology}

\subsection{Problem Definition}
We start by formalizing the ZSL task and then introduce our proposed method based on this formalization. Given a set of labeled seen class examples $\mathcal{D}=\left \{ x_{i}, y_{i} \right \}_{i=1}^{N}$, where $x_{i}$ is seen class example, i.e., image, with class label $y_{i}$ belonging to $m$ seen classes $C=\left \{ c_{1}, c_{2}, \cdots , c_{m}\right \}$. 
The goal is to construct a model for a set of unseen classes ${C}' = \left \{ {c}'_{1}, {c}'_{2}, \cdots , {c}'_{v}\right \}$ ($C \bigcap {C}' = \phi$) from which no example is available during training.
In the inference phase, given a testing unseen class example ${x}'$, the model is expected to predict its class label $c({x}')\in{C}'$. To this end, some auxiliary knowledge, e.g., the semantic descriptions, denoted as $s = \left ( a_{1}, a_{2}, \cdots , a_{n} \right )\in \mathbb{R}^{n}$, are needed to bridge the gaps between the seen and unseen classes. Therefore, the labeled seen class examples $\mathcal{D}$ can be further specified as $\mathcal{D}=\left \{ x_{i}, y_{i}, s_{i} \right \}_{i=1}^{N}$. Each seen class $c_{i}$ is endowed with a semantic prototype $p_{c_{i}}\in \mathbb{R}^{n}$, and similarly, each unseen class ${c_{i}}'$ is also endowed with a semantic prototype $p_{{c_{i}}'}\in \mathbb{R}^{n}$. Thus, for each seen class example we have its semantic features $s_{i} \in P=\left \{p_{c_{1}}, p_{c_{2}}, \cdots , p_{c_{m}}  \right \}$, while for testing unseen class example ${x}'$, we need to predict its semantic features ${s}' \in \mathbb{R}^{n}$ and set the class label by searching the most closely related semantic prototype within ${P}'=\left \{p_{{c_{1}}'}, p_{{c_{2}}'}, \cdots , p_{{c_{v}}'}  \right \}$. In summary, given $\mathcal{D}$, the training can be described as:
\begin{equation}
	\frac{1}{N}\cdot \sum_{i=1}^{N} \mathcal{L} \left ( f\left ( \phi \left ( x_{i} \right ); W \right ), s_{i} \right ) + \varphi \left ( W \right ),
\end{equation}
where $\mathcal{L}\left ( \cdot  \right )$ being the loss function and $\varphi \left ( \cdot \right )$ being the regularization term. The $f\left ( \cdot; W  \right )$ is a mapping function with parameter $W$ maps from the visual feature space to semantic feature space. The $\phi \left ( \cdot  \right )$ is a feature extractor, e.g., a pre-trained CNNs, to obtain the visual features of $x_{i}$. For the inference, given a testing example, e.g., $x_{test}$, the recognition can be described as:
\begin{align}
&c (x_{test}) = \mathop{\arg\max}_{j} \ Sim \left ( f\left ( \phi \left ( x_{test} \right ); W \right ), {{P}'}^{(j)} \right ),
\label{eq2} \\ 
&c (x_{test}) = \mathop{\arg\max}_{j} \ Sim \left ( f\left ( \phi \left ( x_{test} \right ); W \right ), \left \{ {{P}' \cup P} \right \}^{(j)} \right ),
\label{eq3} 
\end{align}
where $Sim \left (\cdot \right )$ is a similarity metric and $c (x_{test})$ searches the most closely related prototype and set the class corresponding with this prototype to $x_{test}$. Specifically, Eq. \ref{eq2} is used for conventional ZSL which the similarity search is only on unseen classes, and Eq. \ref{eq3} is used for generalized ZSL which the search can also generalize to novel examples from seen classes.

\subsection{Graph Neural Networks}
In our method, the GNNs are utilized to receive the example graph and outputs the corresponding class semantic descriptions of the example. Our GNNs framework mainly contains three consecutive parts as graph convolutional layers, graph pooling layers and regression layers. The graph convolution layers are responsible for exacting high-level part representations and local substructure information of examples. The graph pooling layers downsample and fuse the high-level features which coarsens them to more representative graphical features. Last, the regression layers are convolutional and dense layers outputs and links to the semantic descriptions of the example. We introduce the graph convolution layers here and then address our graph pooling and regression layers in Section \ref{Zero-shot-Learning-as-a-Graph-Recognition}. 

Given an example graph as $G = \left ( A, F \right )$, where $A \in \left \{ 0, 1 \right \}^{n \times n}$ is the adjacency matrix of example graph in which the element $1$ denotes there exists an edge between two nodes and $0$ otherwise. $F \in \mathbb{R}^{n \times d}$ is the node feature matrix represents the example graph has $n$ nodes and each node is constituted as $d$-dimensional features. Our graph convolution layers can be described as:
\begin{equation}
	H = \delta \left ( \tilde{D}^{-\frac{1}{2}} \tilde{A} \tilde{D}^{-\frac{1}{2}}F \theta \right ),
\end{equation}
where $\tilde{A} = A + I$ in which $I$ is the identity matrix denotes a self-loop in each node of an example graph. $\tilde{D} = {\sum}_{j} \tilde{A}_{ij}$ is the diagonal degree matrix of nodes. $\theta \in \mathbb{R}^{d \times m}$ is a trainable parameter matrix of graph convolution which receives $d$-dimensional features and outputs $m$-dimensional ones. The $\delta(\cdot)$ is a nonlinear activation function, e.g., $ReLU\left ( \cdot \right ) = max\left ( 0, \cdot \right )$. 

This graph convolution can be decoupled into four operations. A linear feature transformation is first performed by $F \theta$ which maps the node feature representation from $d$ to $m$ channels in the next graph convolution layers. Secondly, $\tilde{A} F \theta$ propagates node information to neighboring nodes as well as the node itself. The $\tilde{D}^{-\frac{1}{2}} \tilde{A} \tilde{D}^{-\frac{1}{2}}$ normalizes each row in obtained node feature matrix $H$ after the graph convolution. The last nonlinear activation function $\delta(\cdot)$ performs point-wise activation and outputs the graph convolution results. The graph convolution can aggregate node features in local neighborhoods to extract local substructure information. To further extract the deep high-level multi-scale features, we stack multiple above graph convolution layers as:
\begin{equation}
	H^{(k+1)} = \delta \left ( \tilde{D}^{-\frac{1}{2}} \tilde{A} \tilde{D}^{-\frac{1}{2}} H^{(k)} \theta^{(k)} \right ),
\end{equation}
where the $k$-th graph convolution layer's output $H^{(k)}$ is mapped by the next layer's parameter matrix $\theta^{(k)}$ and outputs the new $H^{(k+1)}$. By stacking multiple layers allows us to form a deep and hierarchical graph neural network that is powerful and capable of achieving high-level part features interaction and fusion. Assuming that we have stacked $k$ graph convolution layers, then the obtain $\left \{ H^{(i)} \right \}_{i=1}^{k}$ can be further received by subsequential graph pooling and regression layers to output semantic descriptions, which the process can be supervised by training seen class semantic prototypes.

\subsection{Zero-shot Learning as a Graph Recognition}
\label{Zero-shot-Learning-as-a-Graph-Recognition}
Given labeled seen class examples $\mathcal{D}=\left \{ x_{i}, y_{i}, s_{i} \right \}_{i=1}^{N}$, we have three steps to convert the ZSL to a graph-to-semantic mapping problem: 1) parts decomposition, 2) graph construction, and 3) example graph recognition. The parts decomposition obtains several parts of an example. These parts are then presented by a graph structure in which each node is an example part and each edge stands for a certain relation between two corresponding parts. With these presented graphs of labeled seen class examples, we feed them to our GNNs and sequentially pass through the graph convolutional layers, graph pooling layers and regression layers to output semantic descriptions.

\begin{figure}[t]
  \centering
  \centerline{\includegraphics[width=0.89\textwidth]{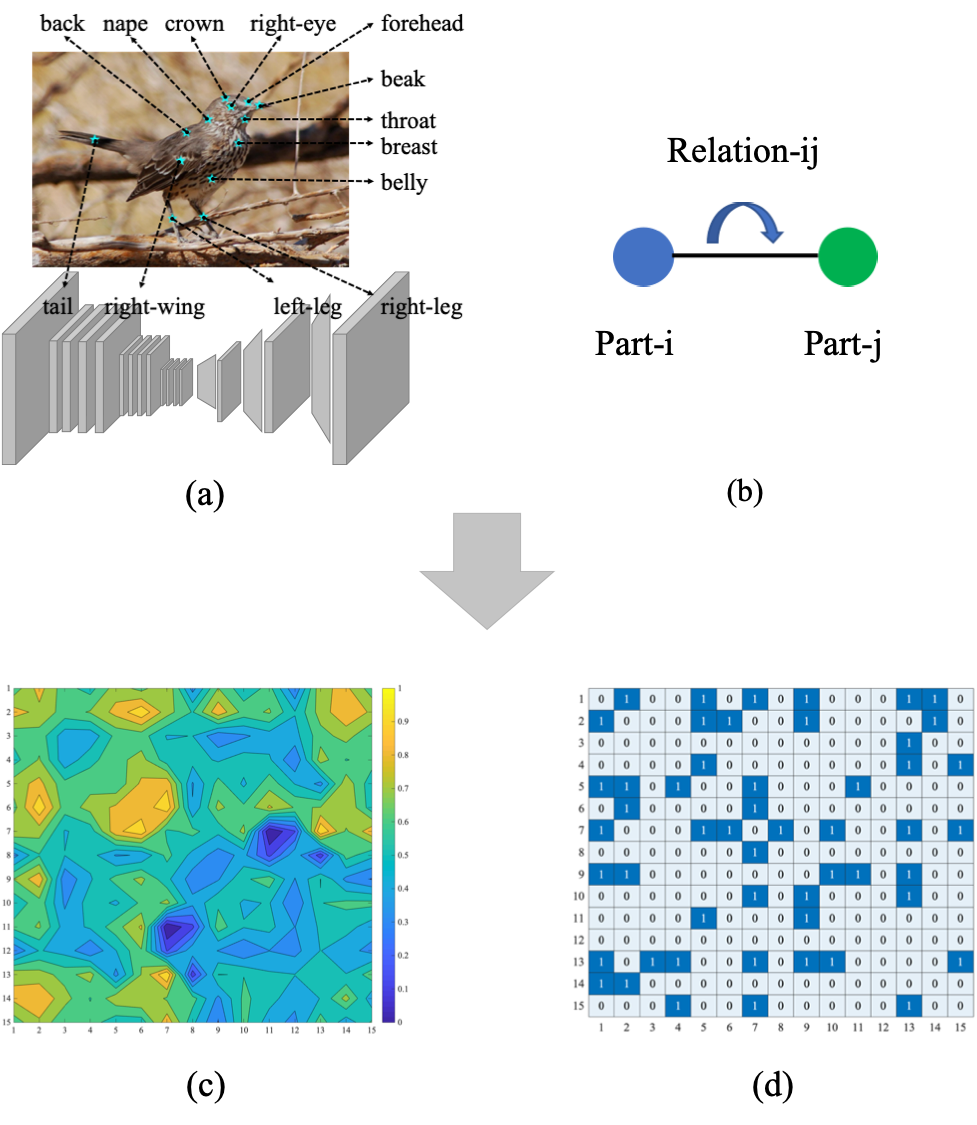}}
  \caption{Example graph construction based on part relations}
  \label{establish}
\end{figure}

\subsubsection{Parts Decomposition}
Key-point localization is usually applied to predict a set of semantic key-points for objects \cite{huang2017coarse,guo2019aligned}. For example, a bird can have several standard key-points reflecting its appearance and subtle characteristics. Taking the CUB-200-2011 Birds dataset \cite{WahCUB_200_2011} as an example, a bird image can be detected with 15 key-points from which each key-point is located at a specific part, e.g., forehead, beak, leg, tail, etc. These key-points can be used to align birds and reveal their subtle differences which helps to recognize different bird species. This recognition problem usually handles the scenario where objects have small inter-class variations such as fine-grained categorization. The deep learning based key-point localization networks usually follow two paradigms for prediction. The first is to directly regress discrete key-point coordinates, and the second is to uses a $2$-dimensional probability distribution heat map of the object image to localize the key-points. In our method, we follow PAIRS \cite{guo2019aligned} which falls into the second paradigm, to construct the parts decomposition module. A pre-trained network, e.g., ResNet-34 \cite{he2016deep} with the classification layer removed, is used as the encoder. Then, by stacking three blocks consisting of one upsampling (bilinear interpolation) layer, one convolution layer, one batch normalization layer and one ReLU layer each, and a final convolution layer and upsampling layer to output the key-points location tensor. 

The parts decomposition module can localize several key-points of an example (Figure \ref{establish}.a), and we can then use a cropping operation to decompose the example (e.g., with width $W$ and height $H$) to several parts based on these key-points as:
\begin{equation}
	\left\{\begin{matrix}
x_{p} = x_{k} - w/2  \\ 
y_{p} = y_{k} - h/2
\end{matrix}\right.,
\label{eq6}
\end{equation}
where $\left ( x_{k}, y_{k} \right )$ is the 2-d coordinate of one key-point. $w$ and $h$ are width and height of the cropped part. $\left ( x_{p}, y_{k} \right )$ is the 2-d coordinate of the top left corner of the cropped part where $x_{p}$ and $y_{p}$ are potentially set to $(W-w)$ and $(H-h)$ respectively, or 0, to retain the cropped part within the example size range. These parts are then can be regarded as nodes of an example graph, and each of them is further be extracted by a CNNs to obtain its visual features as the node features.

\subsubsection{Example Graph Construction}
Given the nodes, we need to determine their neighboring nodes and set an edge if two nodes can be regarded as neighbors. It should be noted that the neighboring nodes are not necessarily spatially adjacent, and the spatial neighbors are also not necessarily neighboring nodes of an example graph. To determine the edges, we assume that every 2-nodes pair is first imposed and established with a relation, and then to verify whether this relation can be satisfied under a certain measurement. In our method, we propose to first establish a relation between two nodes (parts) which is related to the nodes propagation of a graph. In Figure \ref{establish}.b, we assume two nodes Part-i and Part-j can propagate with each other. $f_{i}$ and $f_{j}$ are their node features and the propagation is achieved by $f_{i} = f_{i}w_{i} + f_{j}w_{j}$, $f_{j} = f_{j}w_{j} + f_{i}w_{i}$. Regardless of the orders and weights, we can simply use the mean features of these two nodes, i.e., $\frac{f_{i}+f_{j}}{2}$, to present whether the propagation is positive or not. This is similar to two nodes performing graph convolution with a same weight of $1/2$ and containing self-loop. To verify the positivity, we train a classifier based on whole feature representations of our training examples, and to test on each nodes (parts) pair to compute the classification confidence score on nodes $f_{i}$, $f_{j}$ and their propagated $\frac{f_{i}+f_{j}}{2}$, respectively. If both $score_{(f_{i}+f_{j})/2} > score_{f_{i}}+\varepsilon$ and $score_{(f_{i}+f_{j})/2} > score_{f_{j}}+\varepsilon$ hold, we can say the relation is satisfied. Here the $\varepsilon$ is a small constant that controls the threshold. Taking the CUB-200-2011 Birds\footnote{http://www.vision.caltech.edu/visipedia/CUB-200-2011.html} as an example, we obtain 15 parts of each training example and compute the classification confidence score on each 2-parts pair. The results are shown in Figure \ref{establish}.c which the filled contour plot reflects the confidence of each 2-parts pair under such a relation. The results on the diagonal represent the confidence of the 15 parts themselves and each intersection of a point on the horizontal axis, e.g., $i$, and a point on the vertical axis, e.g., $j$, represents the confidence of pair Part-i and Part-j. By choosing a threshold $\varepsilon$, we can further construct the adjacency matrix of example graph where a smaller threshold results more edges and a larger one controls the number of edges in a reasonable scale. The adjacency matrix is illustrated in Figure \ref{establish}.d. It should be noted that in our method, we focus on the fine-grained ZSL which considers the example-level graph. For simplicity, we assume each example, e.g., birds, has the same part relations and thus the example graph also has the same adjacency matrix.

\subsubsection{Example Graph Recognition}
Given labeled seen class examples $\mathcal{D}=\left \{ x_{i}, y_{i}, s_{i} \right \}_{i=1}^{N}$, we can present them by example graphs as $\mathcal{D}=\left \{ \left ( A_{i}, F_{i} \right ), y_{i}, s_{i} \right \}_{i=1}^{N}$, where $A_{i}$ and $F_{i}$ are the adjacency matrix and node feature matrix, respectively,  
in which we stack $k$ graph convolution layers to obtain $k$ outputs as:
\begin{align}
	H^{(1)} &= \delta \left ( \tilde{D}^{-\frac{1}{2}} \tilde{A} \tilde{D}^{-\frac{1}{2}} H^{(1)} \theta^{(1)} \right ),\\
	H^{(2)} &= \delta \left ( \tilde{D}^{-\frac{1}{2}} \tilde{A} \tilde{D}^{-\frac{1}{2}} H^{(2)} \theta^{(2)} \right ),\\
	&\cdots \\
	H^{(k)} &= \delta \left ( \tilde{D}^{-\frac{1}{2}} \tilde{A} \tilde{D}^{-\frac{1}{2}} H^{(k)} \theta^{(k)} \right ), 
\end{align}
where each obtained $H^{(i)} \in \mathbb{R}^{n \times c_{i}}$ is the propagated feature matrix of $i$-th layer. Each row represents a node and each column represents a feature channel. Specifically, $H^{(i)}$ allows every single node (part) to propagate information to its neighboring nodes (parts) and aggregate their information along with self-ones. 
Next, we add a layer to horizontally concatenate these outputs as:
\begin{equation}
	H^{(1:k)} = \left [ H^{(1)}, H^{(2)}, \cdots ,H^{(k)} \right ],
\end{equation}
where $H^{(1:k)} \in \mathbb{R}^{n \times \sum_{i=1}^{k}c_{i}}$ is the concatenated feature matrix in which each row can be regarded as the global feature representation of a node encoding the high-level and multi-scale local substructure information. For simplicity, we denote $H^{(1:k)} = GNNs\left ( A, F \right )$ and apply the column-wise average pooling to $H^{(1:k)}$ and obtain a $(1 \times \sum_{i=1}^{k}c_{i})$-dimensional feature vector as $V = AvPool\left ( H^{(1:k)} \right )$,
which downsamples and fuses high-level features. As to the regression layers, we simply apply the fully-connected dense multi-layers to $V$ and links to its class semantic prototypes. Thus, given $\mathcal{D}=\left \{ \left ( A_{i}, F_{i} \right ), y_{i}, s_{i} \right \}_{i=1}^{N}$, the training of ZSL can be formalized as:
\begin{equation}
	\frac{1}{N}\cdot \sum_{i=1}^{N} \mathcal{L} \left ( MLP\left ( AvPool\left ( GNNs\left ( A_{i}, F_{i}; W_{g} \right ) \right ); W_{m} \right ), s_{i} \right ) + \varphi \left ( W_{g} \right ) + \gamma \left ( W_{g} \right ),
\end{equation}
where $\mathcal{L}\left ( \cdot  \right )$ is the loss function, $W_{g}$ and $W_{m}$ are trainable network parameters for GNNs and fully-connected dense regression layers, respectively. $\varphi \left ( \cdot \right )$ and $\gamma \left ( \cdot \right )$ being the regularization terms. For the inference, given a testing example, e.g., $\left ( A_{test}, F_{test} \right )$, the recognition can be described as:
\begin{align}
&c \left ( A_{test}, F_{test} \right ) = \mathop{\arg\max}_{j} \ Sim \left ( MLP\left ( AvPool\left ( GNNs\left ( A_{test}, F_{test}; W_{g} \right ) \right ); W_{m} \right ), {{P}'}^{(j)} \right ), 
\label{czsl}   \\ 
&c \left ( A_{test}, F_{test} \right ) = \mathop{\arg\max}_{j} \ Sim \left ( MLP\left ( AvPool\left ( GNNs\left ( A_{test}, F_{test}; W_{g} \right ) \right ); W_{m} \right ), \left \{ {{P}' \cup P} \right \}^{(j)} \right ),
\label{gzsl}
\end{align}
where $Sim \left (\cdot \right )$ is a similarity metric, i.e., we select cosine similarity, and $c \left ( A_{test}, F_{test} \right )$ searches the most closely related prototype and set the class corresponding with this prototype to the test example. Eq. \ref{czsl} is used for the conventional ZSL and Eq. \ref{gzsl} is used for the generalized ZSL.

\section{Experiments}
\label{c5:Experiments}

\subsection{Datasets and Metrics}
To demonstrate the effectiveness of our model, the dataset needs to satisfy some criteria: 1) focuses on the fine-grained domain; 2) annotations on key-points or parts are required to train and obtain parts of examples; and 3) is the benchmark dataset of ZSL. Considering the above criteria on all five benchmark datasets of ZSL, i.e., CUB-200-2011 Birds, Animals with Attributes, aPascal\&Yahoo, SUN Attribute, and ILSVRC2012/ILSVRC2010, only the CUB-200-2011 Birds \cite{WahCUB_200_2011} can satisfy the needs. The CUB-200-2011 Birds\footnote{http://www.vision.caltech.edu/visipedia/CUB-200-2011.html} (CUB) dataset consists of 11,788 images of 200 bird species. Each image is annotated with part location, bounding box, and attribute labels. For ZSL, 150 species containing 8,855 images act as seen classes for training, and the other 50 species containing 2,933 images are unseen classes for testing. Each of their prototypes is represented by a 312-dimensional semantic attribute descriptions present meaningful class-level information. 

In our experiments, we compare our method on two different ZSL settings including 1) \textbf{conventional ZSL} and 2) \textbf{generalized ZSL}. 
For the conventional ZSL setting, all test examples only belong to unseen classes, i.e., only search the class prototypes on seen classes ${P}'$ (Eq. \ref{czsl}). For the generalized ZSL setting \cite{xian2017zero}, the search can also generalize to novel examples from seen classes, i.e., search the class prototypes on both seen and unseen classes $\left \{ {P}' \cup P \right \}$ (Eq. \ref{gzsl}). The hit@k accuracy \cite{frome2013devise, norouzi2013zero} is used to evaluate the model performance. The hit@k accuracy is the standard evaluation metrics in ZSL which can be defined as:
\begin{equation}
	ACC_{hit@k} = \frac{\sum_{i=1}^{n} \mathbf{1}\left [ y_{i} \in \left \{ \tau^{j}\left ( x_{i} \right )  \right \}_{j=1}^{k} \right ]}{n},
\end{equation}
where $\mathbf{1}\left [ \cdot \right ]$ is an indicator function takes the value ``1'' if the argument is true, and ``0'' otherwise. $\tau(\cdot)$ is the operation which determines the class label of example. In our experiments, hit@1 accuracy is used for CUB.

\subsection{Experimental Setups}
Our method is implemented by Pytorch and trained with one Nvidia GTX 1080 Ti GPU. The GNNs consist of four graph convolution layers with hidden channel size 512. A concatenation layer is used to horizontally concatenate outputs of different layers, and a column-wise average pooling is further applied to downsample and fuse the high-level features. 
The regression layers are simple fully-connected dense multi-layers and output 312-dimensional features and supervised by class semantic prototypes. The mean absolute error (MAE) is selected as the loss function $\mathcal{L}\left ( \cdot  \right )$ in Eq. (9). We have the GNNs 5000 runs on the training set.
As to the parts decomposition, we follow all settings of PAIRS \cite{guo2019aligned} to construct the key-point localization network. A pre-trained ResNet-34 \cite{he2016deep} with the classification layer removed is acted as an encoder. Three blocks consisting of one upsampling (bilinear interpolation) layer, one convolution layer, one batch normalization layer, and one ReLU layer each, and a final convolution and upsampling layers are stacked to decode the key-points location. The cropped part size $w$ and $h$ are set to 56 in Eq. \ref{eq6}.
In the experiments, we also report the result when directly use the key-point annotations to construct the example graph, which the two results are denoted as detection based (Ours (D)) model and annotation based (Ours (A)) model, respectively. 
When constructing the example graph, we control the threshold to empirically retain around 50 edges (52 in CUB-200-2011 Birds verified on training examples). 
We use GoogleNet \cite{szegedy2015going} to extract the visual features (1024-d) for nodes (parts).

\subsection{Results and Analysis}

\subsection{Conventional ZSL Results}
To demonstrate the effectiveness of our proposed method, we first compare it with existing representative methods in the conventional ZSL setting. We selected 16 competitors based on the following criteria: 1) published in the most recent years; 2) cover a wide range of models; 3) they clearly represent the state-of-the-art; and 4) all of them are under the standard splits \cite{xian2017zero}. We compute and report the multi-way classification accuracy as in previous works.
The comparison results with the selected representative competitors are shown in Table \ref{results_czsl}. It can be seen from the results that our method outperforms all competitors with great advantages. The prediction accuracy of our method achieves 76.1\% and 78.4\% for detection and annotation based models, respectively. Moreover, we can also observe that the performance of fine-grained based methods is generally better than the average result of all competitors. Specifically, comparing with ${\rm S^2GA}$ \cite{ji2018stacked}, \emph{Chen et al.} \cite{chen2019learning} and \emph{Zhu et al.} \cite{zhu2019semantic} which also fall into the fine-grained ZSL models, our method improves the prediction accuracy by a large margin as 3.1\%, 20.1\% and 7.9\%, respectively, which fully demonstrate the effectiveness of our method.

\begin{table}[h]
    \begin{center}
        \caption{Results of conventional ZSL. F denotes visual features: GoogleNet (G), VGGNet (V). SS denotes semantic features: Attribute (A), Word Vector (W). Reported by accuracy (\%).}
        \setlength{\tabcolsep}{5mm}{ 
        \label{results_czsl} 
            \begin{tabular}{|lccc|c|}        
                \hline                   
                                  Method                                      & F               & SS                         & Fing-grained     & \textbf{50-way 0-shot}\\
                \hline
                                 DeViSE \cite{frome2013devise}                &G    &A/W    &$\times$         &33.5 \\
                                 MTMDL \cite{yang2014unified}                 &G    &A/W    &$\times$         &32.3 \\
                                 ESZSL \cite{romera2015embarrassingly}        &G    &A                &$\times$         &48.7 \\
                                 SSE \cite{zhang2015zero}                     &V    &A                &$\times$         &30.4 \\
                                 SJE \cite{akata2015evaluation}               &G    &A                &$\times$         &50.1 \\
                                 \emph{Ba et al.} \cite{ba2015predicting}     &G    &A/W    &$\times$         &34.0 \\
                                 JLSE \cite{zhang2016zero}                    &V    &A                &$\times$         &42.1 \\
                                 SYNC$^{Struct}$ \cite{changpinyo2016synthesized} &G&A                &$\times$         &54.4 \\
                                 SAE \cite{kodirov2017semantic}               &G    &A                &$\times$         &61.4 \\
                                 RELATION NET \cite{sung2018learning}          &G    &A                &$\times$         &62.0 \\
                                 LSE \cite{yu2018zero}                        &G    &A                &$\times$         &53.2 \\
                                 ${\rm S^2GA}$ \cite{ji2018stacked}           &V    &A                &$\checkmark$     &75.3 \\
                                 \emph{Chen et al.} \cite{chen2019learning}   &G    &A                &$\checkmark$     &58.3 \\
                                 SGAL \cite{yu2019zero}                       &G    &A                &$\times$         &62.5 \\
                                 \emph{Zhu et al.} \cite{zhu2019semantic}     &G    &A                &$\checkmark$     &70.5 \\
                                 AMS-SFE \cite{guo2019ams,guo2020novel}                  &G    &A                &$\times$         &70.1 \\
                                 \hline
                                 Ours (D)                                      &G    &A                &$\checkmark$     &76.1 \\
                                 Ours (A)                                      &G    &A                &$\checkmark$     &\textbf{78.4} \\
                \hline  
        \end{tabular}
        }
    \end{center}  
\end{table}

\subsection{Generalized ZSL Results}
In Table \ref{results_gzsl}, we compare our method with 15 competitors on generalized ZSL setting \cite{xian2017zero}. In addition to the selection criteria of 1), 2) and 3) of conventional ZSL, all selected competitors are also required to comply with the disjoint assumption stated by \cite{xian2017zero}. For the generalized ZSL, we compute and report the average per-class prediction accuracy on test images from unseen classes (U) and seen classes (S), respectively, and report the Harmonic Mean calculated by $H = \left ( 2 \times U \times S \right )/\left ( U + S \right )$, which can quantify the aggregate performance across both seen and unseen classes.
It can be seen from the results that, although most of these competitors cannot retain the same level performance on both seen and unseen classes, our method can achieve the best balanced prediction accuracy. For example, ESZSL \cite{romera2015embarrassingly}, SYNC \cite{changpinyo2016synthesized} and SAE \cite{kodirov2017semantic} have a very large margin, i.e., 51.2\%, 59.4\% and 50.1\%, between the accuracy of seen and unseen classes. In contrast, our method can obtain both comparative results on unseen classes and seen classes as 42.8\% and 69.7\%, and thus resulting the best result of Harmonic Mean as 53.0\%. Our method outperforms all competitors for the most balanced prediction accuracy which makes it better fits a more realistic application scenario.

\begin{table}[h]
    \begin{center}
        \caption{Results of generalized ZSL. F denotes visual features: GoogleNet (G), VGGNet (V), ResNet (R). SS denotes semantic features: Attribute (A), Word Vector (W). Reported by accuracy (\%).}
        \setlength{\tabcolsep}{2.1mm}{ 
        \label{results_gzsl} 
            \begin{tabular}{|lccc|cc|c|}        
                \hline                  
                        Method                               & F               & SS                        & Fine-grained     & U     & S    & \textbf{Harmonic Mean} \\
                \hline
               DeViSE \cite{frome2013devise}         &G    &A/W  &$\times$          &23.8        &53.0      &32.8           \\
               ESZSL \cite{romera2015embarrassingly} &V    &A              &$\times$          &12.6        &63.8      &21.0           \\
               SJE \cite{akata2015evaluation}        &G    &A              &$\times$          &23.5        &59.2      &33.6           \\
               SynC$^{struct}$ \cite{changpinyo2016synthesized} &G    &A              &$\times$          &11.5        &70.9      &19.8           \\
               DEM \cite{zhang2017learning}          &G    &A/W  &$\times$          &19.6        &54.0      &13.6           \\                       
               SAE \cite{kodirov2017semantic}        &G    &A              &$\times$          &7.8         &57.9      &29.2           \\
               RELATION NET \cite{sung2018learning}   &G    &A              &$\times$          &38.1        &61.1      &47.0           \\
               f-CLSWGAN \cite{xian2018feature}      &R    &A              &$\times$          &43.7        &57.7      &49.7           \\
               SE-GZSL \cite{kumar2018generalized}   &R    &A              &$\times$          &41.5        &53.3      &46.7           \\
               SP-AEN \cite{chen2018zero}            &R    &A              &$\times$          &34.7        &70.6      &46.6           \\
               DCN \cite{liu2018generalized}         &G    &A              &$\times$          &28.4        &60.7      &38.7           \\
               SRZSL \cite{annadani2018preserving}     &R  &A              &$\times$          &24.6        &54.3      &33.9           \\
               \emph{Zhu et al.} \cite{zhu2019semantic} &V    &A           &$\checkmark$      &36.7        &71.3      &48.5           \\
               SGAL \cite{yu2019zero}                &G    &A              &$\times$          &40.9        &55.3      &47.0           \\
               DASCN \cite{ni2019dual}               &R    &A              &$\times$          &45.9        &59.0      &51.6           \\
               \hline
               Ours (D)                               &G    &A         &$\checkmark$           &39.1        &64.6      &48.7           \\
               Ours (A)                               &G    &A         &$\checkmark$           &42.8        &69.7      &\textbf{53.0}  \\

                \hline  
        \end{tabular}
        }
    \end{center}  
\end{table}

\subsection{Mapping Robustness}
\begin{figure}[h]
  \centering
  \centerline{\includegraphics[width=0.99\textwidth]{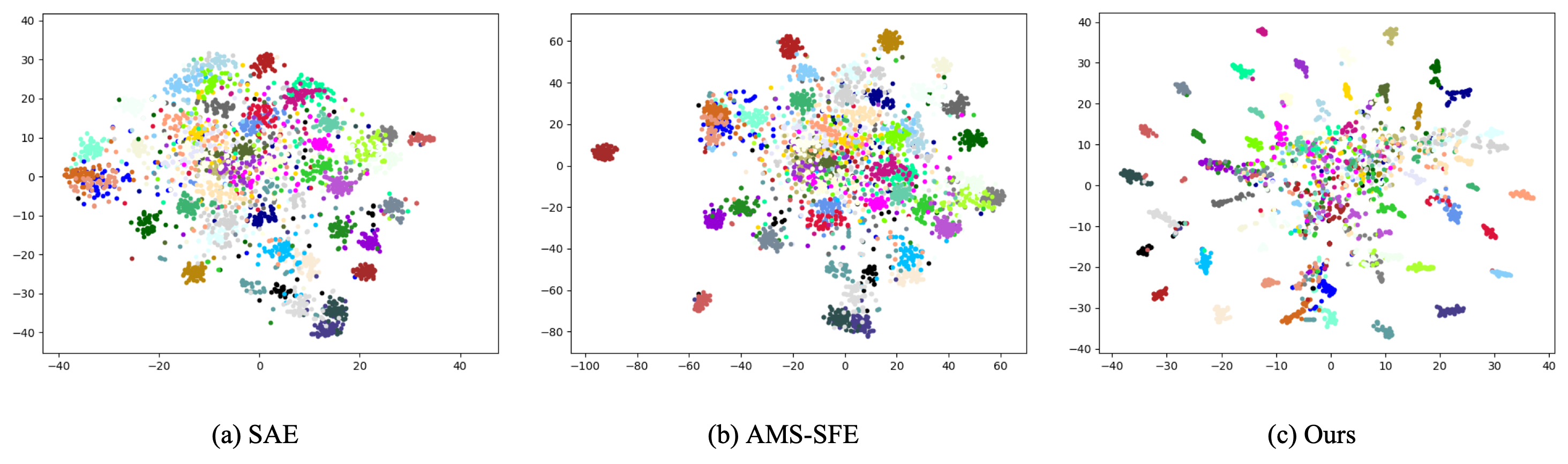}}
  \caption{Visualization results of mapping robustness (better viewed in color).}
  \label{c5_mapping}
\end{figure}

We further conduct the evaluation of mapping robustness on our method. Given the trained model, we map the unseen class examples from the visual to semantic feature space. With these obtained semantic features of examples, we apply t-SNE to visualize them in a 2D map. We show the visualization results on SAE \cite{kodirov2017semantic}, AMS-SFE \cite{guo2020novel} and our method under the conventional ZSL setting in Figure \ref{c5_mapping}.a, Figure \ref{c5_mapping}.b and Figure \ref{c5_mapping}.c, respectively. It can be seen from our method that only a small portion of unseen class examples are shifted in the semantic feature space. Moreover, the obtained semantic features are also more continuous and aggregated. These merits demonstrate that our method can significantly mitigate the domain shift problem.

\section{Remarks}
\label{Remarks}
This chapter proposes a novel fine-grained ZSL framework based on example-level graph, to address the challenging domain shift problem. Our method decomposes examples into several parts to be presented as a graph, in which nodes and edges are different parts and certain relations between parts. Taking advantages from recently developed GNNs, we formulate the ZSL to a graph-to-semantic mapping problem which can better exploit part-semantic correlation and local substructure information in examples. Experimental results verified the effectiveness of our method.

\chapter{Conclusion and Future Work}
\label{ch:6}

\section{Conclusion}
This thesis explores three efficient ways to mitigate the domain shift problem and hence obtain a more robust visual-semantic mapping function for zero-shot learning (ZSL). Our solutions focus on fully empowering the semantic feature space, which is the key building block of ZSL. 
First, we propose to adaptively adjust to rectify the semantic feature space regarding the class prototypes and global distribution. The adjustment improves the ZSL models in two aspects. First, the adjustment of class prototypes helps to rectify the mismatch between the diversity of various visual examples and the identity of unique semantic features, e.g., attributes or word vectors, which makes the visual-semantic mapping more robust and accurate. Second, the adjustment of global distribution helps to decrease the intra-class variance and to enlarge the inter-class diversity in the semantic feature space, which can further mitigate the domain shift problem. 
Second, we propose to align the manifold structures between the visual and semantic feature spaces, by semantic feature expansion. We build upon an autoencoder-based model to expand the semantic features from the visual inputs. Additionally, the expansion is jointly guided by an embedded manifold extracted from the visual feature space. Compared to the previous work which conducts a hard adjustment, the alignment process is more conservative for not directly adjusting the previous semantic features. Thus, we could call it as a soft adjustment. This alignment can derive two benefits: 1) we expand some auxiliary features that can enhance the semantic feature space; 2) more importantly, we implicitly align the manifold structures between the visual and semantic feature spaces. Thus, the mapping function can be better trained and mitigate the domain shift problem. 
Last, we propose to further explore the correlation between the visual and semantic feature spaces in a more fine-grained perspective. Unlike several existing works, we decompose an image example into several parts and use an example-level graph-based model to measure and utilize certain relations between these parts. Taking advantage of recently developed graph neural networks, we further formulate the ZSL to a graph-to-semantic mapping problem, which can better exploit the visual and semantic correlation and the local substructure information in examples. By converting the ZSL to the graph recognition task, we can better utilize the correlation between two feature spaces, thus can mitigate the domain shift problem.

\section{Future Work}
In the future, we mainly have three research directions. First, we will continue investigating the empowerment of semantic feature space for ZSL. This direction aims to enhance the semantic feature space, thus to mitigate the domain shift problem in natural and obtain the more robust visual-semantic mapping. Second, we will go from the coarse-grained method to the fine-grained method to better model the subtle differences among classes in ZSL. Last, we will investigate and conduct more work on the graph-based method to solve the ZSL problem. This direction makes use of the powerful representation ability of the graph techniques, e.g., graph neural networks, to model complex correlation in real world application.



\addcontentsline{toc}{chapter}{Bibliography}

\bibliography{my}

\begin{thebibliography}{135}
\providecommand{\natexlab}[1]{#1}
\providecommand{\url}[1]{\texttt{#1}}
\expandafter\ifx\csname urlstyle\endcsname\relax
  \providecommand{\doi}[1]{doi: #1}\else
  \providecommand{\doi}{doi: \begingroup \urlstyle{rm}\Url}\fi

\bibitem[Krizhevsky et~al.(2012)Krizhevsky, Sutskever, and
  Hinton]{krizhevsky2012imagenet}
Alex Krizhevsky, Ilya Sutskever, and Geoffrey~E Hinton.
\newblock Imagenet classification with deep convolutional neural networks.
\newblock In \emph{Advances in neural information processing systems}, pages
  1097--1105, 2012.

\bibitem[Ren et~al.(2016)Ren, He, Girshick, and Sun]{ren2016faster}
Shaoqing Ren, Kaiming He, Ross Girshick, and Jian Sun.
\newblock Faster r-cnn: Towards real-time object detection with region proposal
  networks.
\newblock \emph{IEEE transactions on pattern analysis and machine
  intelligence}, 39\penalty0 (6):\penalty0 1137--1149, 2016.

\bibitem[Zhao et~al.(2017)Zhao, Wu, Feng, Peng, and Yan]{zhao2017diversified}
Bo~Zhao, Xiao Wu, Jiashi Feng, Qiang Peng, and Shuicheng Yan.
\newblock Diversified visual attention networks for fine-grained object
  classification.
\newblock \emph{IEEE Transactions on Multimedia}, 19\penalty0 (6):\penalty0
  1245--1256, 2017.

\bibitem[Zhang et~al.(2018{\natexlab{a}})Zhang, Wu, Shen, Zhang, and
  Lu]{zhang2018multilabel}
Junjie Zhang, Qi~Wu, Chunhua Shen, Jian Zhang, and Jianfeng Lu.
\newblock Multilabel image classification with regional latent semantic
  dependencies.
\newblock \emph{IEEE Transactions on Multimedia}, 20\penalty0 (10):\penalty0
  2801--2813, 2018{\natexlab{a}}.

\bibitem[Guo et~al.(2020)Guo, Ma, Zhang, Zhou, and Guo]{guo2020dual}
Jingcai Guo, Shiheng Ma, Jie Zhang, Qihua Zhou, and Song Guo.
\newblock Dual-view attention networks for single image super-resolution.
\newblock In \emph{Proceedings of the 28th ACM International Conference on
  Multimedia}, pages 2728--2736, 2020.

\bibitem[Hu et~al.(2014)Hu, Lu, and Tan]{hu2014discriminative}
Junlin Hu, Jiwen Lu, and Yap-Peng Tan.
\newblock Discriminative deep metric learning for face verification in the
  wild.
\newblock In \emph{Proceedings of the IEEE conference on computer vision and
  pattern recognition}, pages 1875--1882, 2014.

\bibitem[Ding and Tao(2015)]{ding2015robust}
Changxing Ding and Dacheng Tao.
\newblock Robust face recognition via multimodal deep face representation.
\newblock \emph{IEEE Transactions on Multimedia}, 17\penalty0 (11):\penalty0
  2049--2058, 2015.

\bibitem[Taigman et~al.(2014)Taigman, Yang, Ranzato, and
  Wolf]{taigman2014deepface}
Yaniv Taigman, Ming Yang, Marc'Aurelio Ranzato, and Lior Wolf.
\newblock Deepface: Closing the gap to human-level performance in face
  verification.
\newblock In \emph{Proceedings of the IEEE conference on computer vision and
  pattern recognition}, pages 1701--1708, 2014.

\bibitem[Chopra et~al.(2005)Chopra, Hadsell, and LeCun]{chopra2005learning}
Sumit Chopra, Raia Hadsell, and Yann LeCun.
\newblock Learning a similarity metric discriminatively, with application to
  face verification.
\newblock In \emph{2005 IEEE Computer Society Conference on Computer Vision and
  Pattern Recognition (CVPR'05)}, volume~1, pages 539--546. IEEE, 2005.

\bibitem[Kumar et~al.(2011)Kumar, Berg, Belhumeur, and
  Nayar]{kumar2011describable}
Neeraj Kumar, Alexander Berg, Peter~N Belhumeur, and Shree Nayar.
\newblock Describable visual attributes for face verification and image search.
\newblock \emph{IEEE Transactions on Pattern Analysis and Machine
  Intelligence}, 33\penalty0 (10):\penalty0 1962--1977, 2011.

\bibitem[Rasiwasia et~al.(2010)Rasiwasia, Costa~Pereira, Coviello, Doyle,
  Lanckriet, Levy, and Vasconcelos]{rasiwasia2010new}
Nikhil Rasiwasia, Jose Costa~Pereira, Emanuele Coviello, Gabriel Doyle, Gert~RG
  Lanckriet, Roger Levy, and Nuno Vasconcelos.
\newblock A new approach to cross-modal multimedia retrieval.
\newblock In \emph{Proceedings of the 18th ACM international conference on
  Multimedia}, pages 251--260. ACM, 2010.

\bibitem[Kang et~al.(2015)Kang, Xiang, Liao, Xu, and Pan]{kang2015learning}
Cuicui Kang, Shiming Xiang, Shengcai Liao, Changsheng Xu, and Chunhong Pan.
\newblock Learning consistent feature representation for cross-modal multimedia
  retrieval.
\newblock \emph{IEEE Transactions on Multimedia}, 17\penalty0 (3):\penalty0
  370--381, 2015.

\bibitem[Shen et~al.(2010)Shen, Li, and Huang]{shen2010active}
Tian Shen, Hongsheng Li, and Xiaolei Huang.
\newblock Active volume models for medical image segmentation.
\newblock \emph{IEEE transactions on medical imaging}, 30\penalty0
  (3):\penalty0 774--791, 2010.

\bibitem[Wernick et~al.(2010)Wernick, Yang, Brankov, Yourganov, and
  Strother]{wernick2010machine}
Miles~N Wernick, Yongyi Yang, Jovan~G Brankov, Grigori Yourganov, and Stephen~C
  Strother.
\newblock Machine learning in medical imaging.
\newblock \emph{IEEE signal processing magazine}, 27\penalty0 (4):\penalty0
  25--38, 2010.

\bibitem[Greenspan et~al.(2016)Greenspan, Van~Ginneken, and
  Summers]{greenspan2016guest}
Hayit Greenspan, Bram Van~Ginneken, and Ronald~M Summers.
\newblock Guest editorial deep learning in medical imaging: Overview and future
  promise of an exciting new technique.
\newblock \emph{IEEE Transactions on Medical Imaging}, 35\penalty0
  (5):\penalty0 1153--1159, 2016.

\bibitem[Oakden-Rayner et~al.(2020)Oakden-Rayner, Dunnmon, Carneiro, and
  R{\'e}]{oakden2020hidden}
Luke Oakden-Rayner, Jared Dunnmon, Gustavo Carneiro, and Christopher R{\'e}.
\newblock Hidden stratification causes clinically meaningful failures in
  machine learning for medical imaging.
\newblock In \emph{Proceedings of the ACM Conference on Health, Inference, and
  Learning}, pages 151--159, 2020.

\bibitem[Ding et~al.(2015)Ding, Zhang, Liu, and Duan]{ding2015deep}
Xiao Ding, Yue Zhang, Ting Liu, and Junwen Duan.
\newblock Deep learning for event-driven stock prediction.
\newblock In \emph{Twenty-fourth international joint conference on artificial
  intelligence}, 2015.

\bibitem[Luo et~al.(2018)Luo, Ao, Pan, Wang, Zhao, Yu, and He]{luo2018beyond}
Ling Luo, Xiang Ao, Feiyang Pan, Jin Wang, Tong Zhao, Ningzi Yu, and Qing He.
\newblock Beyond polarity: Interpretable financial sentiment analysis with
  hierarchical query-driven attention.
\newblock In \emph{IJCAI}, pages 4244--4250, 2018.

\bibitem[Ma et~al.(2019)Ma, Guo, Guo, and Guo]{ma2019position}
Shiheng Ma, Jingcai Guo, Song Guo, and Minyi Guo.
\newblock Position-aware convolutional networks for traffic prediction.
\newblock \emph{arXiv preprint arXiv:1904.06187}, 2019.

\bibitem[Deng et~al.(2009)Deng, Dong, Socher, Li, Li, and
  Fei-Fei]{deng2009imagenet}
Jia Deng, Wei Dong, Richard Socher, Li-Jia Li, Kai Li, and Li~Fei-Fei.
\newblock Imagenet: A large-scale hierarchical image database.
\newblock In \emph{2009 IEEE conference on computer vision and pattern
  recognition}, pages 248--255. Ieee, 2009.

\bibitem[He et~al.(2015)He, Zhang, Ren, and Sun]{he2015delving}
Kaiming He, Xiangyu Zhang, Shaoqing Ren, and Jian Sun.
\newblock Delving deep into rectifiers: Surpassing human-level performance on
  imagenet classification.
\newblock In \emph{Proceedings of the IEEE international conference on computer
  vision}, pages 1026--1034, 2015.

\bibitem[Pan and Yang(2009)]{pan2009survey}
Sinno~Jialin Pan and Qiang Yang.
\newblock A survey on transfer learning.
\newblock \emph{IEEE Transactions on knowledge and data engineering},
  22\penalty0 (10):\penalty0 1345--1359, 2009.

\bibitem[Torrey and Shavlik(2010)]{torrey2010transfer}
Lisa Torrey and Jude Shavlik.
\newblock Transfer learning.
\newblock In \emph{Handbook of research on machine learning applications and
  trends: algorithms, methods, and techniques}, pages 242--264. IGI Global,
  2010.

\bibitem[Long et~al.(2017{\natexlab{a}})Long, Zhu, Wang, and
  Jordan]{long2017deep}
Mingsheng Long, Han Zhu, Jianmin Wang, and Michael~I Jordan.
\newblock Deep transfer learning with joint adaptation networks.
\newblock In \emph{International conference on machine learning}, pages
  2208--2217. PMLR, 2017{\natexlab{a}}.

\bibitem[Jia et~al.(2018)Jia, Zhang, Weiss, Wang, Shen, Ren, Nguyen, Pang,
  Moreno, Wu, et~al.]{jia2018transfer}
Ye~Jia, Yu~Zhang, Ron Weiss, Quan Wang, Jonathan Shen, Fei Ren, Patrick Nguyen,
  Ruoming Pang, Ignacio~Lopez Moreno, Yonghui Wu, et~al.
\newblock Transfer learning from speaker verification to multispeaker
  text-to-speech synthesis.
\newblock In \emph{Advances in neural information processing systems}, pages
  4480--4490, 2018.

\bibitem[Biederman(1987)]{biederman1987recognition}
Irving Biederman.
\newblock Recognition-by-components: a theory of human image understanding.
\newblock \emph{Psychological review}, 94\penalty0 (2):\penalty0 115, 1987.

\bibitem[Lampert et~al.(2009)Lampert, Nickisch, and
  Harmeling]{lampert2009learning}
Christoph~H Lampert, Hannes Nickisch, and Stefan Harmeling.
\newblock Learning to detect unseen object classes by between-class attribute
  transfer.
\newblock In \emph{2009 IEEE Conference on Computer Vision and Pattern
  Recognition}, pages 951--958. IEEE, 2009.

\bibitem[Frome et~al.(2013)Frome, Corrado, Shlens, Bengio, Dean, Ranzato, and
  Mikolov]{frome2013devise}
Andrea Frome, Greg~S Corrado, Jon Shlens, Samy Bengio, Jeff Dean, Marc'Aurelio
  Ranzato, and Tomas Mikolov.
\newblock Devise: A deep visual-semantic embedding model.
\newblock In \emph{Advances in neural information processing systems}, pages
  2121--2129, 2013.

\bibitem[Lampert et~al.(2014)Lampert, Nickisch, and
  Harmeling]{lampert2014attribute}
Christoph~H Lampert, Hannes Nickisch, and Stefan Harmeling.
\newblock Attribute-based classification for zero-shot visual object
  categorization.
\newblock \emph{IEEE Transactions on Pattern Analysis and Machine
  Intelligence}, 36\penalty0 (3):\penalty0 453--465, 2014.

\bibitem[Shigeto et~al.(2015)Shigeto, Suzuki, Hara, Shimbo, and
  Matsumoto]{shigeto2015ridge}
Yutaro Shigeto, Ikumi Suzuki, Kazuo Hara, Masashi Shimbo, and Yuji Matsumoto.
\newblock Ridge regression, hubness, and zero-shot learning.
\newblock In \emph{Joint European Conference on Machine Learning and Knowledge
  Discovery in Databases}, pages 135--151. Springer, 2015.

\bibitem[Wang et~al.(2015)Wang, Hu, Liang, Yu, Jiang, Ye, Chen, and
  Leng]{wang2015zero}
Zheng Wang, Ruimin Hu, Chao Liang, Yi~Yu, Junjun Jiang, Mang Ye, Jun Chen, and
  Qingming Leng.
\newblock Zero-shot person re-identification via cross-view consistency.
\newblock \emph{IEEE Transactions on Multimedia}, 18\penalty0 (2):\penalty0
  260--272, 2015.

\bibitem[Bucher et~al.(2016)Bucher, Herbin, and Jurie]{bucher2016improving}
Maxime Bucher, St{\'e}phane Herbin, and Fr{\'e}d{\'e}ric Jurie.
\newblock Improving semantic embedding consistency by metric learning for
  zero-shot classiffication.
\newblock In \emph{European Conference on Computer Vision}, pages 730--746.
  Springer, 2016.

\bibitem[Zhang and Saligrama(2016)]{zhang2016zero}
Ziming Zhang and Venkatesh Saligrama.
\newblock Zero-shot learning via joint latent similarity embedding.
\newblock In \emph{Proceedings of the IEEE Conference on Computer Vision and
  Pattern Recognition}, pages 6034--6042, 2016.

\bibitem[Changpinyo et~al.(2016)Changpinyo, Chao, Gong, and
  Sha]{changpinyo2016synthesized}
Soravit Changpinyo, Wei-Lun Chao, Boqing Gong, and Fei Sha.
\newblock Synthesized classifiers for zero-shot learning.
\newblock In \emph{Proceedings of the IEEE Conference on Computer Vision and
  Pattern Recognition}, pages 5327--5336, 2016.

\bibitem[Kodirov et~al.(2017)Kodirov, Xiang, and Gong]{kodirov2017semantic}
Elyor Kodirov, Tao Xiang, and Shaogang Gong.
\newblock Semantic autoencoder for zero-shot learning.
\newblock In \emph{Proceedings of the IEEE Conference on Computer Vision and
  Pattern Recognition}, pages 3174--3183, 2017.

\bibitem[Zhu et~al.(2018)Zhu, Elhoseiny, Liu, Peng, and
  Elgammal]{zhu2018generative}
Yizhe Zhu, Mohamed Elhoseiny, Bingchen Liu, Xi~Peng, and Ahmed Elgammal.
\newblock A generative adversarial approach for zero-shot learning from noisy
  texts.
\newblock In \emph{Proceedings of the IEEE Conference on Computer Vision and
  Pattern Recognition (CVPR)}, 2018.

\bibitem[Guo and Guo(2019{\natexlab{a}})]{guo2019ams}
Jingcai Guo and Song Guo.
\newblock Ams-sfe: Towards an alignment of manifold structures via semantic
  feature expansion for zero-shot learning.
\newblock In \emph{2019 IEEE International Conference on Multimedia and Expo
  (ICME)}, pages 73--78. IEEE, 2019{\natexlab{a}}.

\bibitem[Fu et~al.(2015{\natexlab{a}})Fu, Hospedales, Xiang, and
  Gong]{fu2015transductive}
Yanwei Fu, Timothy~M Hospedales, Tao Xiang, and Shaogang Gong.
\newblock Transductive multi-view zero-shot learning.
\newblock \emph{IEEE transactions on pattern analysis and machine
  intelligence}, 37\penalty0 (11):\penalty0 2332--2345, 2015{\natexlab{a}}.

\bibitem[Farhadi et~al.(2009)Farhadi, Endres, Hoiem, and
  Forsyth]{farhadi2009describing}
Ali Farhadi, Ian Endres, Derek Hoiem, and David Forsyth.
\newblock Describing objects by their attributes.
\newblock In \emph{Computer Vision and Pattern Recognition, 2009. CVPR 2009.
  IEEE Conference on}, pages 1778--1785. IEEE, 2009.

\bibitem[Patterson et~al.(2014)Patterson, Xu, Su, and Hays]{patterson2014sun}
Genevieve Patterson, Chen Xu, Hang Su, and James Hays.
\newblock The sun attribute database: Beyond categories for deeper scene
  understanding.
\newblock \emph{International Journal of Computer Vision}, 108\penalty0
  (1-2):\penalty0 59--81, 2014.

\bibitem[Wah et~al.(2011{\natexlab{a}})Wah, Branson, Welinder, Perona, and
  Belongie]{wah2011caltech}
Catherine Wah, Steve Branson, Peter Welinder, Pietro Perona, and Serge
  Belongie.
\newblock The caltech-ucsd birds-200-2011 dataset.
\newblock 2011{\natexlab{a}}.

\bibitem[Mikolov et~al.(2013{\natexlab{a}})Mikolov, Chen, Corrado, and
  Dean]{mikolov2013efficient}
Tomas Mikolov, Kai Chen, Greg Corrado, and Jeffrey Dean.
\newblock Efficient estimation of word representations in vector space.
\newblock In \emph{Proceedings of the 1st International Conference on Learning
  Representations (ICLR)}, 2013{\natexlab{a}}.

\bibitem[Mikolov et~al.(2013{\natexlab{b}})Mikolov, Sutskever, Chen, Corrado,
  and Dean]{mikolov2013distributed}
Tomas Mikolov, Ilya Sutskever, Kai Chen, Greg~S Corrado, and Jeff Dean.
\newblock Distributed representations of words and phrases and their
  compositionality.
\newblock In \emph{Advances in neural information processing systems}, pages
  3111--3119, 2013{\natexlab{b}}.

\bibitem[Joulin et~al.(2017)Joulin, Grave, Bojanowski, and
  Mikolov]{joulin2017bag}
Armand Joulin, {\'E}douard Grave, Piotr Bojanowski, and Tom{\'a}{\v{s}}
  Mikolov.
\newblock Bag of tricks for efficient text classification.
\newblock In \emph{Proceedings of the 15th Conference of the European Chapter
  of the Association for Computational Linguistics: Volume 2, Short Papers},
  pages 427--431, 2017.

\bibitem[Pennington et~al.(2014)Pennington, Socher, and
  Manning]{pennington2014glove}
Jeffrey Pennington, Richard Socher, and Christopher~D Manning.
\newblock Glove: Global vectors for word representation.
\newblock In \emph{Proceedings of the 2014 conference on empirical methods in
  natural language processing (EMNLP)}, pages 1532--1543, 2014.

\bibitem[Xian et~al.(2016)Xian, Akata, Sharma, Nguyen, Hein, and
  Schiele]{xian2016latent}
Yongqin Xian, Zeynep Akata, Gaurav Sharma, Quynh Nguyen, Matthias Hein, and
  Bernt Schiele.
\newblock Latent embeddings for zero-shot classification.
\newblock In \emph{Proceedings of the IEEE Conference on Computer Vision and
  Pattern Recognition}, pages 69--77, 2016.

\bibitem[Zhang et~al.(2016)Zhang, Gong, and Shah]{zhang2016fast}
Yang Zhang, Boqing Gong, and Mubarak Shah.
\newblock Fast zero-shot image tagging.
\newblock In \emph{2016 IEEE Conference on Computer Vision and Pattern
  Recognition (CVPR)}, pages 5985--5994. IEEE, 2016.

\bibitem[Bansal et~al.(2018)Bansal, Sikka, Sharma, Chellappa, and
  Divakaran]{bansal2018zero}
Ankan Bansal, Karan Sikka, Gaurav Sharma, Rama Chellappa, and Ajay Divakaran.
\newblock Zero-shot object detection.
\newblock In \emph{Proceedings of the European Conference on Computer Vision
  (ECCV)}, pages 384--400, 2018.

\bibitem[Guo and Guo(2020)]{guo2020novel}
Jingcai Guo and Song Guo.
\newblock A novel perspective to zero-shot learning: Towards an alignment of
  manifold structures via semantic feature expansion.
\newblock \emph{IEEE Transactions on Multimedia}, 2020.

\bibitem[Xian et~al.(2018{\natexlab{a}})Xian, Lampert, Schiele, and
  Akata]{xian2018zero}
Yongqin Xian, Christoph~H Lampert, Bernt Schiele, and Zeynep Akata.
\newblock Zero-shot learning—a comprehensive evaluation of the good, the bad
  and the ugly.
\newblock \emph{IEEE transactions on pattern analysis and machine
  intelligence}, 41\penalty0 (9):\penalty0 2251--2265, 2018{\natexlab{a}}.

\bibitem[Miller(1995)]{miller1995wordnet}
George~A Miller.
\newblock Wordnet: a lexical database for english.
\newblock \emph{Communications of the ACM}, 38\penalty0 (11):\penalty0 39--41,
  1995.

\bibitem[Palatucci et~al.(2009)Palatucci, Pomerleau, Hinton, and
  Mitchell]{palatucci2009zero}
Mark Palatucci, Dean Pomerleau, Geoffrey~E Hinton, and Tom~M Mitchell.
\newblock Zero-shot learning with semantic output codes.
\newblock In \emph{Advances in neural information processing systems}, pages
  1410--1418, 2009.

\bibitem[Akata et~al.(2013)Akata, Perronnin, Harchaoui, and
  Schmid]{akata2013label}
Zeynep Akata, Florent Perronnin, Zaid Harchaoui, and Cordelia Schmid.
\newblock Label-embedding for attribute-based classification.
\newblock In \emph{Proceedings of the IEEE Conference on Computer Vision and
  Pattern Recognition}, pages 819--826, 2013.

\bibitem[Akata et~al.(2015)Akata, Reed, Walter, Lee, and
  Schiele]{akata2015evaluation}
Zeynep Akata, Scott Reed, Daniel Walter, Honglak Lee, and Bernt Schiele.
\newblock Evaluation of output embeddings for fine-grained image
  classification.
\newblock In \emph{Proceedings of the IEEE Conference on Computer Vision and
  Pattern Recognition}, pages 2927--2936, 2015.

\bibitem[Romera-Paredes and Torr(2015)]{romera2015embarrassingly}
Bernardino Romera-Paredes and Philip Torr.
\newblock An embarrassingly simple approach to zero-shot learning.
\newblock In \emph{International Conference on Machine Learning}, pages
  2152--2161, 2015.

\bibitem[Akata et~al.(2016{\natexlab{a}})Akata, Perronnin, Harchaoui, and
  Schmid]{akata2016label}
Zeynep Akata, Florent Perronnin, Zaid Harchaoui, and Cordelia Schmid.
\newblock Label-embedding for image classification.
\newblock \emph{IEEE transactions on pattern analysis and machine
  intelligence}, 38\penalty0 (7):\penalty0 1425--1438, 2016{\natexlab{a}}.

\bibitem[Socher et~al.(2013)Socher, Ganjoo, Manning, and Ng]{socher2013zero}
Richard Socher, Milind Ganjoo, Christopher~D Manning, and Andrew Ng.
\newblock Zero-shot learning through cross-modal transfer.
\newblock In \emph{Advances in neural information processing systems}, pages
  935--943, 2013.

\bibitem[Lei~Ba et~al.(2015)Lei~Ba, Swersky, Fidler, et~al.]{lei2015predicting}
Jimmy Lei~Ba, Kevin Swersky, Sanja Fidler, et~al.
\newblock Predicting deep zero-shot convolutional neural networks using textual
  descriptions.
\newblock In \emph{Proceedings of the IEEE International Conference on Computer
  Vision}, pages 4247--4255, 2015.

\bibitem[Chen et~al.(2018)Chen, Zhang, Xiao, Liu, and Chang]{chen2018zero}
Long Chen, Hanwang Zhang, Jun Xiao, Wei Liu, and Shih-Fu Chang.
\newblock Zero-shot visual recognition using semantics-preserving adversarial
  embedding networks.
\newblock In \emph{Proceedings of the IEEE Conference on Computer Vision and
  Pattern Recognition}, pages 1043--1052, 2018.

\bibitem[Dinu et~al.(2015)Dinu, Lazaridou, and Baroni]{dinu2014improving}
Georgiana Dinu, Angeliki Lazaridou, and Marco Baroni.
\newblock Improving zero-shot learning by mitigating the hubness problem.
\newblock In \emph{Proceedings of the 3rd International Conference on Learning
  Representations (ICLR)}, 2015.

\bibitem[Zhang et~al.(2017)Zhang, Xiang, and Gong]{zhang2017learning}
Li~Zhang, Tao Xiang, and Shaogang Gong.
\newblock Learning a deep embedding model for zero-shot learning.
\newblock In \emph{Proceedings of the IEEE Conference on Computer Vision and
  Pattern Recognition}, pages 2021--2030, 2017.

\bibitem[Ba et~al.(2015)Ba, Swersky, Fidler, and
  Salakhutdinov]{ba2015predicting}
Lei~Jimmy Ba, Kevin Swersky, Sanja Fidler, and Ruslan Salakhutdinov.
\newblock Predicting deep zero-shot convolutional neural networks using textual
  descriptions.
\newblock In \emph{ICCV}, pages 4247--4255, 2015.

\bibitem[Changpinyo et~al.(2017)Changpinyo, Chao, and
  Sha]{changpinyo2016predicting}
Soravit Changpinyo, Wei-Lun Chao, and Fei Sha.
\newblock Predicting visual exemplars of unseen classes for zero-shot learning.
\newblock In \emph{Proceedings of the IEEE international conference on computer
  vision}, pages 3476--3485, 2017.

\bibitem[Mishra et~al.(2018{\natexlab{a}})Mishra, Krishna~Reddy, Mittal, and
  Murthy]{mishra2018generative}
Ashish Mishra, Shiva Krishna~Reddy, Anurag Mittal, and Hema~A Murthy.
\newblock A generative model for zero shot learning using conditional
  variational autoencoders.
\newblock In \emph{Proceedings of the IEEE Conference on Computer Vision and
  Pattern Recognition Workshops}, pages 2188--2196, 2018{\natexlab{a}}.

\bibitem[Verma and Rai(2017)]{verma2017simple}
Vinay~Kumar Verma and Piyush Rai.
\newblock A simple exponential family framework for zero-shot learning.
\newblock In \emph{Joint European Conference on Machine Learning and Knowledge
  Discovery in Databases}, pages 792--808. Springer, 2017.

\bibitem[Kumar~Verma et~al.(2018)Kumar~Verma, Arora, Mishra, and
  Rai]{kumar2018generalized}
Vinay Kumar~Verma, Gundeep Arora, Ashish Mishra, and Piyush Rai.
\newblock Generalized zero-shot learning via synthesized examples.
\newblock In \emph{Proceedings of the IEEE conference on computer vision and
  pattern recognition}, pages 4281--4289, 2018.

\bibitem[Xian et~al.(2018{\natexlab{b}})Xian, Lorenz, Schiele, and
  Akata]{xian2018feature}
Yongqin Xian, Tobias Lorenz, Bernt Schiele, and Zeynep Akata.
\newblock Feature generating networks for zero-shot learning.
\newblock In \emph{Proceedings of the IEEE conference on computer vision and
  pattern recognition}, pages 5542--5551, 2018{\natexlab{b}}.

\bibitem[Lu(2016)]{lu2015unsupervised}
Yao Lu.
\newblock Unsupervised learning on neural network outputs: with application in
  zero-shot learning.
\newblock In \emph{Proceedings of the Twenty-Fifth International Joint
  Conference on Artificial Intelligence}, pages 3432--3438, 2016.

\bibitem[Zhang and Saligrama(2015)]{zhang2015zero}
Ziming Zhang and Venkatesh Saligrama.
\newblock Zero-shot learning via semantic similarity embedding.
\newblock In \emph{Proceedings of the IEEE international conference on computer
  vision}, pages 4166--4174, 2015.

\bibitem[Fu et~al.(2015{\natexlab{b}})Fu, Xiang, Kodirov, and Gong]{fu2015zero}
Zhenyong Fu, Tao Xiang, Elyor Kodirov, and Shaogang Gong.
\newblock Zero-shot object recognition by semantic manifold distance.
\newblock In \emph{Proceedings of the IEEE Conference on Computer Vision and
  Pattern Recognition}, pages 2635--2644, 2015{\natexlab{b}}.

\bibitem[Akata et~al.(2016{\natexlab{b}})Akata, Malinowski, Fritz, and
  Schiele]{akata2016multi}
Zeynep Akata, Mateusz Malinowski, Mario Fritz, and Bernt Schiele.
\newblock Multi-cue zero-shot learning with strong supervision.
\newblock In \emph{Proceedings of the IEEE Conference on Computer Vision and
  Pattern Recognition}, pages 59--68, 2016{\natexlab{b}}.

\bibitem[Long et~al.(2017{\natexlab{b}})Long, Liu, Shao, Shen, Ding, and
  Han]{long2017zero}
Yang Long, Li~Liu, Ling Shao, Fumin Shen, Guiguang Ding, and Jungong Han.
\newblock From zero-shot learning to conventional supervised classification:
  Unseen visual data synthesis.
\newblock 2017{\natexlab{b}}.

\bibitem[Sung et~al.(2018)Sung, Yang, Zhang, Xiang, Torr, and
  Hospedales]{sung2018learning}
Flood Sung, Yongxin Yang, Li~Zhang, Tao Xiang, Philip~HS Torr, and Timothy~M
  Hospedales.
\newblock Learning to compare: Relation network for few-shot learning.
\newblock In \emph{Proceedings of the IEEE Conference on Computer Vision and
  Pattern Recognition}, pages 1199--1208, 2018.

\bibitem[Liu et~al.(2018)Liu, Long, Wang, and Jordan]{liu2018generalized}
Shichen Liu, Mingsheng Long, Jianmin Wang, and Michael~I Jordan.
\newblock Generalized zero-shot learning with deep calibration network.
\newblock In \emph{Advances in Neural Information Processing Systems}, pages
  2005--2015, 2018.

\bibitem[Kankuekul et~al.(2012)Kankuekul, Kawewong, Tangruamsub, and
  Hasegawa]{kankuekul2012online}
Pichai Kankuekul, Aram Kawewong, Sirinart Tangruamsub, and Osamu Hasegawa.
\newblock Online incremental attribute-based zero-shot learning.
\newblock In \emph{Computer Vision and Pattern Recognition (CVPR), 2012 IEEE
  Conference on}, pages 3657--3664. IEEE, 2012.

\bibitem[Norouzi et~al.(2014)Norouzi, Mikolov, Bengio, Singer, Shlens, Frome,
  Corrado, and Dean]{norouzi2013zero}
Mohammad Norouzi, Tomas Mikolov, Samy Bengio, Yoram Singer, Jonathon Shlens,
  Andrea Frome, Greg~S Corrado, and Jeffrey Dean.
\newblock Zero-shot learning by convex combination of semantic embeddings.
\newblock In \emph{Proceedings of the 2nd International Conference on Learning
  Representations (ICLR)}, 2014.

\bibitem[Jayaraman and Grauman(2014)]{jayaraman2014zero}
Dinesh Jayaraman and Kristen Grauman.
\newblock Zero-shot recognition with unreliable attributes.
\newblock In \emph{Advances in neural information processing systems}, pages
  3464--3472, 2014.

\bibitem[Al-Halah et~al.(2016)Al-Halah, Tapaswi, and
  Stiefelhagen]{al2016recovering}
Ziad Al-Halah, Makarand Tapaswi, and Rainer Stiefelhagen.
\newblock Recovering the missing link: Predicting class-attribute associations
  for unsupervised zero-shot learning.
\newblock In \emph{Proceedings of the IEEE Conference on Computer Vision and
  Pattern Recognition}, pages 5975--5984, 2016.

\bibitem[Wang et~al.(2018)Wang, Ye, and Gupta]{wang2018zero}
Xiaolong Wang, Yufei Ye, and Abhinav Gupta.
\newblock Zero-shot recognition via semantic embeddings and knowledge graphs.
\newblock In \emph{Proceedings of the IEEE conference on computer vision and
  pattern recognition}, pages 6857--6866, 2018.

\bibitem[Kampffmeyer et~al.(2019)Kampffmeyer, Chen, Liang, Wang, Zhang, and
  Xing]{kampffmeyer2019rethinking}
Michael Kampffmeyer, Yinbo Chen, Xiaodan Liang, Hao Wang, Yujia Zhang, and
  Eric~P Xing.
\newblock Rethinking knowledge graph propagation for zero-shot learning.
\newblock In \emph{Proceedings of the IEEE Conference on Computer Vision and
  Pattern Recognition}, pages 11487--11496, 2019.

\bibitem[Ji et~al.(2018)Ji, Fu, Guo, Pang, Zhang, et~al.]{ji2018stacked}
Zhong Ji, Yanwei Fu, Jichang Guo, Yanwei Pang, Zhongfei~Mark Zhang, et~al.
\newblock Stacked semantics-guided attention model for fine-grained zero-shot
  learning.
\newblock In \emph{Advances in Neural Information Processing Systems}, pages
  5995--6004, 2018.

\bibitem[Zhu et~al.(2019)Zhu, Xie, Tang, Peng, and Elgammal]{zhu2019semantic}
Yizhe Zhu, Jianwen Xie, Zhiqiang Tang, Xi~Peng, and Ahmed Elgammal.
\newblock Semantic-guided multi-attention localization for zero-shot learning.
\newblock In \emph{Advances in Neural Information Processing Systems}, pages
  14917--14927, 2019.

\bibitem[Chen et~al.(2019)Chen, Cao, and Ji]{chen2019learning}
Hong Chen, Liujuan Cao, and Rongrong Ji.
\newblock Learning similarity-specific dictionary for zero-shot fine-grained
  recognition.
\newblock In \emph{ICASSP 2019-2019 IEEE International Conference on Acoustics,
  Speech and Signal Processing (ICASSP)}, pages 3697--3701. IEEE, 2019.

\bibitem[Russakovsky et~al.(2015)Russakovsky, Deng, Su, Krause, Satheesh, Ma,
  Huang, Karpathy, Khosla, Bernstein, et~al.]{russakovsky2015imagenet}
Olga Russakovsky, Jia Deng, Hao Su, Jonathan Krause, Sanjeev Satheesh, Sean Ma,
  Zhiheng Huang, Andrej Karpathy, Aditya Khosla, Michael Bernstein, et~al.
\newblock Imagenet large scale visual recognition challenge.
\newblock \emph{International Journal of Computer Vision}, 115\penalty0
  (3):\penalty0 211--252, 2015.

\bibitem[Welinder et~al.(2010)Welinder, Branson, Perona, and
  Belongie]{welinder2010multidimensional}
Peter Welinder, Steve Branson, Pietro Perona, and Serge~J Belongie.
\newblock The multidimensional wisdom of crowds.
\newblock In \emph{Advances in neural information processing systems}, pages
  2424--2432, 2010.

\bibitem[Li et~al.(2017)Li, Wang, Hu, Lin, and Zhuang]{li2017zero}
Yanan Li, Donghui Wang, Huanhang Hu, Yuetan Lin, and Yueting Zhuang.
\newblock Zero-shot recognition using dual visual-semantic mapping paths.
\newblock In \emph{IEEE Conference on Computer Vision and Pattern Recognition
  (CVPR)}, pages 5207--5215, 2017.

\bibitem[Song et~al.(2018)Song, Shen, Yang, Liu, and
  Song]{song2018transductive}
Jie Song, Chengchao Shen, Yezhou Yang, Yang Liu, and Mingli Song.
\newblock Transductive unbiased embedding for zero-shot learning.
\newblock In \emph{Proceedings of the IEEE Conference on Computer Vision and
  Pattern Recognition}, pages 1024--1033, 2018.

\bibitem[Tong et~al.(2018)Tong, Klinkigt, Chen, Cui, Kong, Murakami, and
  Kobayashi]{tong2018adversarial}
Bin Tong, Martin Klinkigt, Junwen Chen, Xiankun Cui, Quan Kong, Tomokazu
  Murakami, and Yoshiyuki Kobayashi.
\newblock Adversarial zero-shot learning with semantic augmentation.
\newblock In \emph{AAAI}, 2018.

\bibitem[Rahman et~al.(2018)Rahman, Khan, and Porikli]{rahman2018unified}
Shafin Rahman, Salman Khan, and Fatih Porikli.
\newblock A unified approach for conventional zero-shot, generalized zero-shot
  and few-shot learning.
\newblock \emph{IEEE Transactions on Image Processing}, 2018.

\bibitem[Radovanovic et~al.(2010)Radovanovic, Nanopoulos, and
  Ivanovic]{radovanovic2010hubs}
Milos Radovanovic, Alexandros Nanopoulos, and Mirjana Ivanovic.
\newblock Hubs in space: Popular nearest neighbors in high-dimensional data.
\newblock \emph{Journal of Machine Learning Research}, 11\penalty0
  (sept):\penalty0 2487--2531, 2010.

\bibitem[Xu et~al.(2017)Xu, Shen, Yang, Zhang, Shen, and Song]{xu2017matrix}
Xing Xu, Fumin Shen, Yang Yang, Dongxiang Zhang, Heng~Tao Shen, and Jingkuan
  Song.
\newblock Matrix tri-factorization with manifold regularizations for zero-shot
  learning.
\newblock In \emph{Proceeding of the IEEE conference on computer vision and
  pattern recognition. CVPR}, 2017.

\bibitem[Yu and Lee(2019)]{yu2019zero}
Hyeonwoo Yu and Beomhee Lee.
\newblock Zero-shot learning via simultaneous generating and learning.
\newblock In \emph{Advances in Neural Information Processing Systems}, pages
  46--56, 2019.

\bibitem[Vincent et~al.(2010)Vincent, Larochelle, Lajoie, Bengio, and
  Manzagol]{vincent2010stacked}
Pascal Vincent, Hugo Larochelle, Isabelle Lajoie, Yoshua Bengio, and
  Pierre-Antoine Manzagol.
\newblock Stacked denoising autoencoders: Learning useful representations in a
  deep network with a local denoising criterion.
\newblock \emph{Journal of machine learning research}, 11\penalty0
  (Dec):\penalty0 3371--3408, 2010.

\bibitem[Xu et~al.(2016)Xu, Xiang, Liu, Gilmore, Wu, Tang, and
  Madabhushi]{xu2016stacked}
Jun Xu, Lei Xiang, Qingshan Liu, Hannah Gilmore, Jianzhong Wu, Jinghai Tang,
  and Anant Madabhushi.
\newblock Stacked sparse autoencoder (ssae) for nuclei detection on breast
  cancer histopathology images.
\newblock \emph{IEEE transactions on medical imaging}, 35\penalty0
  (1):\penalty0 119--130, 2016.

\bibitem[Yu et~al.(2013)Yu, Zeng, Luo, Zhuang, He, and Shi]{yu2013embedding}
Wenchao Yu, Guangxiang Zeng, Ping Luo, Fuzhen Zhuang, Qing He, and Zhongzhi
  Shi.
\newblock Embedding with autoencoder regularization.
\newblock In \emph{Joint European Conference on Machine Learning and Knowledge
  Discovery in Databases}, pages 208--223. Springer, 2013.

\bibitem[Makhzani and Frey(2014)]{makhzani2014winner}
Alireza Makhzani and Brendan Frey.
\newblock A winner-take-all method for training sparse convolutional
  autoencoders.
\newblock In \emph{NIPS Deep Learning Workshop}. Citeseer, 2014.

\bibitem[Chu and Cai(2017)]{chu2017stacked}
Wenqing Chu and Deng Cai.
\newblock Stacked similarity-aware autoencoders.
\newblock In \emph{Proceedings of the 26th International Joint Conference on
  Artificial Intelligence}, pages 1561--1567. AAAI Press, 2017.

\bibitem[Awange et~al.(2018)Awange, Pal{\'a}ncz, Lewis, and
  V{\"o}lgyesi]{awange2018particle}
Joseph~L Awange, B{\'e}la Pal{\'a}ncz, Robert~H Lewis, and Lajos V{\"o}lgyesi.
\newblock Particle swarm optimization.
\newblock In \emph{Mathematical Geosciences}, pages 167--184. Springer, 2018.

\bibitem[Eberhart and Kennedy(1995)]{eberhart1995new}
Russell Eberhart and James Kennedy.
\newblock A new optimizer using particle swarm theory.
\newblock In \emph{Micro Machine and Human Science, 1995. MHS'95., Proceedings
  of the Sixth International Symposium on}, pages 39--43. IEEE, 1995.

\bibitem[Boureau et~al.(2008)Boureau, Cun, et~al.]{boureau2008sparse}
Y-lan Boureau, Yann~L Cun, et~al.
\newblock Sparse feature learning for deep belief networks.
\newblock In \emph{Advances in neural information processing systems}, pages
  1185--1192, 2008.

\bibitem[Pomet and Praly(1992)]{pomet1992adaptive}
J-B Pomet and Laurent Praly.
\newblock Adaptive nonlinear regulation: Estimation from the lyapunov equation.
\newblock \emph{IEEE Transactions on automatic control}, 37\penalty0
  (6):\penalty0 729--740, 1992.

\bibitem[Bartels and Stewart(1972)]{bartels1972solution}
Richard~H. Bartels and George~W Stewart.
\newblock Solution of the matrix equation ax+ xb= c [f4].
\newblock \emph{Communications of the ACM}, 15\penalty0 (9):\penalty0 820--826,
  1972.

\bibitem[Osherson et~al.(1991)Osherson, Stern, Wilkie, Stob, and
  Smith]{osherson1991default}
Daniel~N Osherson, Joshua Stern, Ormond Wilkie, Michael Stob, and Edward~E
  Smith.
\newblock Default probability.
\newblock \emph{Cognitive Science}, 15\penalty0 (2):\penalty0 251--269, 1991.

\bibitem[Kemp et~al.(2006)Kemp, Tenenbaum, Griffiths, Yamada, and
  Ueda]{kemp2006learning}
Charles Kemp, Joshua~B Tenenbaum, Thomas~L Griffiths, Takeshi Yamada, and
  Naonori Ueda.
\newblock Learning systems of concepts with an infinite relational model.
\newblock In \emph{AAAI}, volume~3, page~5, 2006.

\bibitem[Hsu et~al.()Hsu, Chang, Lin, et~al.]{hsu2003practical}
Chih-Wei Hsu, Chih-Chung Chang, Chih-Jen Lin, et~al.
\newblock A practical guide to support vector classification.

\bibitem[Szegedy et~al.(2015)Szegedy, Liu, Jia, Sermanet, Reed, Anguelov,
  Erhan, Vanhoucke, and Rabinovich]{szegedy2015going}
Christian Szegedy, Wei Liu, Yangqing Jia, Pierre Sermanet, Scott Reed, Dragomir
  Anguelov, Dumitru Erhan, Vincent Vanhoucke, and Andrew Rabinovich.
\newblock Going deeper with convolutions.
\newblock In \emph{Proceedings of the IEEE conference on computer vision and
  pattern recognition}, pages 1--9, 2015.

\bibitem[Wang and Chen(2017)]{wang2017zero}
Qian Wang and Ke~Chen.
\newblock Zero-shot visual recognition via bidirectional latent embedding.
\newblock \emph{International Journal of Computer Vision}, 124\penalty0
  (3):\penalty0 356--383, 2017.

\bibitem[Yang and Hospedales(2015)]{yang2014unified}
Yongxin Yang and Timothy~M Hospedales.
\newblock A unified perspective on multi-domain and multi-task learning.
\newblock In \emph{Proceedings of the 3rd International Conference on Learning
  Representations (ICLR)}, 2015.

\bibitem[Fu and Sigal(2016)]{fu2016semi}
Yanwei Fu and Leonid Sigal.
\newblock Semi-supervised vocabulary-informed learning.
\newblock In \emph{Proceedings of the IEEE Conference on Computer Vision and
  Pattern Recognition}, pages 5337--5346, 2016.

\bibitem[Mishra et~al.(2018{\natexlab{b}})Mishra, Krishna~Reddy, Mittal, and
  Murthy]{mishra2017generative}
Ashish Mishra, Shiva Krishna~Reddy, Anurag Mittal, and Hema~A Murthy.
\newblock A generative model for zero shot learning using conditional
  variational autoencoders.
\newblock In \emph{Proceedings of the IEEE Conference on Computer Vision and
  Pattern Recognition Workshops}, pages 2188--2196, 2018{\natexlab{b}}.

\bibitem[Reed et~al.(2016)Reed, Akata, Lee, and Schiele]{reed2016learning}
Scott Reed, Zeynep Akata, Honglak Lee, and Bernt Schiele.
\newblock Learning deep representations of fine-grained visual descriptions.
\newblock In \emph{Proceedings of the IEEE Conference on Computer Vision and
  Pattern Recognition}, pages 49--58, 2016.

\bibitem[Guo and Guo(2019{\natexlab{b}})]{guo2019ee}
Jingcai Guo and Song Guo.
\newblock Ee-ae: An exclusivity enhanced unsupervised feature learning
  approach.
\newblock In \emph{ICASSP 2019-2019 IEEE International Conference on Acoustics,
  Speech and Signal Processing (ICASSP)}, pages 3517--3521. IEEE,
  2019{\natexlab{b}}.

\bibitem[Wang et~al.(2017)Wang, Chen, and Li]{wang2017quantifying}
Qi~Wang, Mulin Chen, and Xuelong Li.
\newblock Quantifying and detecting collective motion by manifold learning.
\newblock In \emph{AAAI}, pages 4292--4298, 2017.

\bibitem[Goodfellow et~al.(2014)Goodfellow, Pouget-Abadie, Mirza, Xu,
  Warde-Farley, Ozair, Courville, and Bengio]{goodfellow2014generative}
Ian Goodfellow, Jean Pouget-Abadie, Mehdi Mirza, Bing Xu, David Warde-Farley,
  Sherjil Ozair, Aaron Courville, and Yoshua Bengio.
\newblock Generative adversarial nets.
\newblock In \emph{Advances in neural information processing systems}, pages
  2672--2680, 2014.

\bibitem[Baldi(2012)]{baldi2012autoencoders}
Pierre Baldi.
\newblock Autoencoders, unsupervised learning, and deep architectures.
\newblock In \emph{Proceedings of ICML workshop on unsupervised and transfer
  learning}, pages 37--49, 2012.

\bibitem[Zhang et~al.(2018{\natexlab{b}})Zhang, Xu, Li, Zhang, Wang, Huang, and
  Metaxas]{zhang2018stackgan++}
Han Zhang, Tao Xu, Hongsheng Li, Shaoting Zhang, Xiaogang Wang, Xiaolei Huang,
  and Dimitris~N Metaxas.
\newblock Stackgan++: Realistic image synthesis with stacked generative
  adversarial networks.
\newblock \emph{IEEE transactions on pattern analysis and machine
  intelligence}, 41\penalty0 (8):\penalty0 1947--1962, 2018{\natexlab{b}}.

\bibitem[Kingma and Welling(2014)]{DBLP:journals/corr/KingmaW13}
Diederik~P. Kingma and Max Welling.
\newblock Auto-encoding variational bayes.
\newblock In \emph{2nd International Conference on Learning Representations,
  {ICLR} 2014, Banff, AB, Canada, April 14-16, 2014, Conference Track
  Proceedings}, 2014.

\bibitem[Blei et~al.(2017)Blei, Kucukelbir, and McAuliffe]{blei2017variational}
David~M Blei, Alp Kucukelbir, and Jon~D McAuliffe.
\newblock Variational inference: A review for statisticians.
\newblock \emph{Journal of the American statistical Association}, 112\penalty0
  (518):\penalty0 859--877, 2017.

\bibitem[Guo(2016)]{guo2016improved}
Jingcai Guo.
\newblock An improved incremental training approach for large scaled dataset
  based on support vector machine.
\newblock In \emph{Big Data Computing Applications and Technologies (BDCAT),
  2016 IEEE/ACM 3rd International Conference on}, pages 149--157. IEEE, 2016.

\bibitem[Kendall et~al.(2018)Kendall, Gal, and Cipolla]{kendall2018multi}
Alex Kendall, Yarin Gal, and Roberto Cipolla.
\newblock Multi-task learning using uncertainty to weigh losses for scene
  geometry and semantics.
\newblock In \emph{Proceedings of the IEEE Conference on Computer Vision and
  Pattern Recognition}, pages 7482--7491, 2018.

\bibitem[Guo and Guo(2019{\natexlab{c}})]{guo2019adaptive}
Jingcai Guo and Song Guo.
\newblock Adaptive adjustment with semantic feature space for zero-shot
  recognition.
\newblock In \emph{ICASSP 2019-2019 IEEE International Conference on Acoustics,
  Speech and Signal Processing (ICASSP)}, pages 3287--3291. IEEE,
  2019{\natexlab{c}}.

\bibitem[Yu et~al.(2018)Yu, Ji, Guo, and Zhang]{yu2018zero}
Yunlong Yu, Zhong Ji, Jichang Guo, and Zhongfei Zhang.
\newblock Zero-shot learning via latent space encoding.
\newblock \emph{IEEE transactions on cybernetics}, 49\penalty0 (10):\penalty0
  3755--3766, 2018.

\bibitem[Maaten and Hinton(2008)]{maaten2008visualizing}
Laurens van~der Maaten and Geoffrey Hinton.
\newblock Visualizing data using t-sne.
\newblock \emph{Journal of machine learning research}, 9\penalty0
  (Nov):\penalty0 2579--2605, 2008.

\bibitem[Schonfeld et~al.(2019)Schonfeld, Ebrahimi, Sinha, Darrell, and
  Akata]{schonfeld2019generalized}
Edgar Schonfeld, Sayna Ebrahimi, Samarth Sinha, Trevor Darrell, and Zeynep
  Akata.
\newblock Generalized zero-and few-shot learning via aligned variational
  autoencoders.
\newblock In \emph{Proceedings of the IEEE Conference on Computer Vision and
  Pattern Recognition}, pages 8247--8255, 2019.

\bibitem[Huang et~al.(2019)Huang, Wang, Yu, and Wang]{huang2019generative}
He~Huang, Changhu Wang, Philip~S Yu, and Chang-Dong Wang.
\newblock Generative dual adversarial network for generalized zero-shot
  learning.
\newblock In \emph{Proceedings of the IEEE conference on computer vision and
  pattern recognition}, pages 801--810, 2019.

\bibitem[Ni et~al.(2019)Ni, Zhang, and Xie]{ni2019dual}
Jian Ni, Shanghang Zhang, and Haiyong Xie.
\newblock Dual adversarial semantics-consistent network for generalized
  zero-shot learning.
\newblock In \emph{Advances in Neural Information Processing Systems}, pages
  6143--6154, 2019.

\bibitem[Zhang et~al.(2018{\natexlab{c}})Zhang, Cui, Neumann, and
  Chen]{zhang2018end}
Muhan Zhang, Zhicheng Cui, Marion Neumann, and Yixin Chen.
\newblock An end-to-end deep learning architecture for graph classification.
\newblock In \emph{Thirty-Second AAAI Conference on Artificial Intelligence},
  2018{\natexlab{c}}.

\bibitem[Ying et~al.(2018)Ying, You, Morris, Ren, Hamilton, and
  Leskovec]{ying2018hierarchical}
Zhitao Ying, Jiaxuan You, Christopher Morris, Xiang Ren, Will Hamilton, and
  Jure Leskovec.
\newblock Hierarchical graph representation learning with differentiable
  pooling.
\newblock In \emph{Advances in neural information processing systems}, pages
  4800--4810, 2018.

\bibitem[Zhao and Wang(2019)]{zhao2019learning}
Qi~Zhao and Yusu Wang.
\newblock Learning metrics for persistence-based summaries and applications for
  graph classification.
\newblock In \emph{Advances in Neural Information Processing Systems}, pages
  9855--9866, 2019.

\bibitem[Huang et~al.(2017)Huang, Gong, and Tao]{huang2017coarse}
Shaoli Huang, Mingming Gong, and Dacheng Tao.
\newblock A coarse-fine network for keypoint localization.
\newblock In \emph{Proceedings of the IEEE International Conference on Computer
  Vision}, pages 3028--3037, 2017.

\bibitem[Guo and Farrell(2019)]{guo2019aligned}
Pei Guo and Ryan Farrell.
\newblock Aligned to the object, not to the image: A unified pose-aligned
  representation for fine-grained recognition.
\newblock In \emph{2019 IEEE Winter Conference on Applications of Computer
  Vision (WACV)}, pages 1876--1885. IEEE, 2019.

\bibitem[Wah et~al.(2011{\natexlab{b}})Wah, Branson, Welinder, Perona, and
  Belongie]{WahCUB_200_2011}
C.~Wah, S.~Branson, P.~Welinder, P.~Perona, and S.~Belongie.
\newblock {The Caltech-UCSD Birds-200-2011 Dataset}.
\newblock Technical report, 2011{\natexlab{b}}.

\bibitem[He et~al.(2016)He, Zhang, Ren, and Sun]{he2016deep}
Kaiming He, Xiangyu Zhang, Shaoqing Ren, and Jian Sun.
\newblock Deep residual learning for image recognition.
\newblock In \emph{Proceedings of the IEEE conference on computer vision and
  pattern recognition}, pages 770--778, 2016.

\bibitem[Xian et~al.(2017)Xian, Schiele, and Akata]{xian2017zero}
Yongqin Xian, Bernt Schiele, and Zeynep Akata.
\newblock Zero-shot learning-the good, the bad and the ugly.
\newblock In \emph{Proceedings of the IEEE Conference on Computer Vision and
  Pattern Recognition}, pages 4582--4591, 2017.

\bibitem[Annadani and Biswas(2018)]{annadani2018preserving}
Yashas Annadani and Soma Biswas.
\newblock Preserving semantic relations for zero-shot learning.
\newblock In \emph{Proceedings of the IEEE Conference on Computer Vision and
  Pattern Recognition}, pages 7603--7612, 2018.

\end{thebibliography}

\end{document}